\pdfoutput=1
\documentclass{article}
\usepackage[utf8]{inputenc} %
\usepackage[T1]{fontenc}    %
\usepackage{xcolor}        %
\usepackage{algorithm}     %
\usepackage{hyperref}       %

\usepackage[accepted]{mlsys2025}
\usepackage{etoolbox}
\makeatletter
\renewcommand{\Notice@String}{}
\patchcmd{\printAffiliationsAndNotice}{\ \\ \Notice@String}{}{}%
  {\PackageError{main}{Could not remove the MLSys notice}{}}
\patchcmd{\printAffiliationsAndNotice}{\hspace*{-\footnotesep}}{\raggedright\hspace*{-\footnotesep}}{}%
  {\PackageError{main}{Could not set the affiliation footnote ragged right}{}}
\makeatother
\hypersetup{pdftitle={Switchcraft: AI Model Router for Agentic Tool Calling},
  pdfauthor={Sharad Agarwal, Pooria Namyar, Alec Wolman, Rahul Ambavat, Ankur Gupta, Qizheng Zhang},
  pdfsubject={}}
\usepackage{xurl}           %
\usepackage{graphicx}       %
\usepackage{booktabs}       %
\usepackage{tabularx}       %
\usepackage{multirow}       %
\usepackage{amsfonts}       %
\usepackage{amsmath}        %
\usepackage{amssymb}        %
\usepackage{nicefrac}       %
\usepackage{microtype}      %
\usepackage{enumitem}       %
\setlist{nosep}             %
\usepackage{tikz}           %
\usetikzlibrary{shapes, arrows.meta, positioning, calc}
\usepackage{pgfplots}       %
\pgfplotsset{compat=1.18}
\usepgfplotslibrary{groupplots}

\usepackage[disable, textsize=small]{todonotes}

\mlsyssetsymbol{intern}{$\dagger$}
\makeatletter
\newcommand{\printReviewNotice}{{\let\thefootnote\relax\footnotetext{\hspace*{-1.8em}\Notice@String}}}
\makeatother

\begin{document}

\twocolumn[
\mlsystitle{Switchcraft: AI Model Router for Agentic Tool Calling}

\begin{mlsysauthorlist}
\mlsysauthor{Sharad Agarwal}{msr}
\mlsysauthor{Pooria Namyar}{msr}
\mlsysauthor{Alec Wolman}{msr}
\mlsysauthor{Rahul Ambavat}{ms}
\mlsysauthor{Ankur Gupta}{ms}
\mlsysauthor{Qizheng Zhang}{intern,stanford}
\end{mlsysauthorlist}

\mlsysaffiliation{msr}{Microsoft Research}
\mlsysaffiliation{ms}{Microsoft}
\mlsysaffiliation{stanford}{Stanford University}
\mlsyscorrespondingauthor{Sharad~Agarwal}{\mbox{sagarwal@microsoft.com}}
\mlsyscorrespondingauthor{Pooria~Namyar}{\mbox{namyarpooria@microsoft.com}}
\mlsyscorrespondingauthor{Alec~Wolman}{\mbox{alecw@microsoft.com}}
\mlsyscorrespondingauthor{Rahul~Ambavat}{\mbox{raambava@microsoft.com}}
\mlsyscorrespondingauthor{Ankur~Gupta}{\mbox{angup@microsoft.com}}
\mlsyscorrespondingauthor{Qizheng~Zhang}{\mbox{qizhengz@stanford.edu}}
\mlsyskeywords{Model routing, agentic AI, tool calling, cost-aware inference}

\vskip 0.3in
\begin{abstract}

Agentic AI systems that invoke external tools are powerful but costly,
leading developers to default to large models and overspend inference
budgets. Existing model routers target chat
responses or end-task success, not the correctness of the individual tool
calls on which agent tasks depend; we present \textbf{Switchcraft}, the first
(to the best of our knowledge) model router optimized for strict tool-call
correctness in agentic workloads. We construct an evaluation framework on six
function-calling benchmarks, holding one out entirely, and fine-tune a
calibrated DistilBERT-based classifier, deployed inline under a latency
budget, to select the lowest-cost model subject to correctness or abstain
when none is likely to answer correctly.
Switchcraft reaches 78.3\% answer accuracy (86.4\% when correct abstentions
are credited) at 59\% lower cost than the best individual model (75.6\%),
saving over \$3{,}800 per million queries, and outperforms kNN, MIRT,
heuristic, and random routers. On the held-out
benchmark, it matches or exceeds every individual model at 5--7$\times$
lower cost than the strongest ones. We find two counter-intuitive results:
larger models do not consistently outperform smaller ones on tool-use tasks,
and nominally cheaper models can incur higher total cost due to
token-intensive output.

\end{abstract}

]

\printAffiliationsAndNotice{\textsuperscript{$\dagger$}Work performed while an intern at Microsoft Research. }

\section{Introduction}
\label{sec:introduction}

\textbf{Managing inference costs} for agentic AI systems -- where LLMs invoke
external tools and APIs to perform complex
tasks~\cite{llexus,agentic_ai_survey} -- is emerging as the critical issue
facing enterprise users of AI models today.
In interviews with a dozen enterprise teams, we found that selecting models
for tool-assisted queries remains challenging: teams default to large,
widely used models, leading to overspending -- a problem for
customers who overpay and for providers facing oversubscribed GPU
capacity even when smaller models would suffice.

\textbf{Model routing} addresses this by dynamically selecting an
appropriate LLM per query, routing simple requests to smaller models and
reserving large models for harder ones. Prior work shows substantial
savings; e.g., an AWS service reports $\sim 43.9\%$ cost reduction through
intelligent routing~\cite{aws_ipr}. However, most existing model
routers~\cite{hybrid_llm,routellm,routerarena} target chat completion, and
those built for agents optimize end-task
success~\cite{zhang2026mtroutercostawaremultiturnllm}, not the precise tool
invocations agents must generate at each step. A per-step router
instead needs to know which models will get each
tool call right, as one wrong call can derail the task
(Section~\ref{sec:motivation}) or irrevocably change external state, as in
a finance tool call.

We introduce a different approach for routers that support agentic tool
calling. Invoked on every agent turn, the router sits on the critical path and
must decide within milliseconds. This rules out designs suited to
latency-tolerant settings, such as LLM-based
routers~\cite{zhang2025routerr1teachingllmsmultiround} or cascades that
escalate from a cheap model on failure~\cite{cascade_routing}. We instead use
one forward pass of a small encoder. To train and
test it, we curate six public function-calling benchmarks -- Berkeley Function
Calling Leaderboard (BFCL v3)~\cite{bfcl}, AWS ConFETTI~\cite{confetti},
Salesforce xLAM-60K~\cite{xlam}, Hermes~\cite{hermes3},
FunReason-MT~\cite{funreason_mt}, and Glaive Function
Calling~\cite{glaive_fc}, which we hold out entirely as an unseen
test -- covering tool use, multi-turn interaction, and parallel API calls.
We build a unified evaluation framework that normalizes
these datasets, executes each query across candidate LLMs, and uses an
abstract syntax tree (AST)-based checker to score tool invocations
robustly. Using these signals, we fine-tune \textbf{Switchcraft}, a
DistilBERT-based router (67M parameters) that takes an agent's query and
context as input and either selects the cheapest model likely to answer
correctly or abstains when none is.

\textbf{Key Results.} On our in-distribution test set, Switchcraft
achieves 78.3\% answer accuracy -- 2.6\,pp above the best single model
(GPT-5.3-chat at 75.6\%) and above kNN, MIRT, heuristic, and random routers -- while
reducing inference cost by 59\%, saving over \$3{,}800 per million queries.
It abstains on 9\% of queries, catching 57\% of those no model can answer;
with these correct abstentions credited, 86.4\% of its routing decisions are
correct. A single cost budget sets the operating point: a larger budget raises
routing accuracy to 87.9\% (79.7\% answer accuracy) at 34.1\% lower cost than
GPT-5.3-chat, and a smaller one still beats every single model at 70.1\%
lower cost. On held-out Glaive, it matches or exceeds every individual model at
5--7$\times$ lower cost than the strongest ones. Routing takes at most
16.7\,ms (P99) on a commodity T4 GPU. On BFCL's run-to-completion tasks, executed in
simulated environments, Switchcraft beats kNN and every model but the
3.8$\times$ costlier GPT-5.4. With KV caching, routing every turn beats both
favoring a conversation's current model (7.3\% costlier at similar accuracy) and
routing only its first turn (5.7\,pp less accurate;
Section~\ref{sec:multi-turn}). We further find that costlier models do not
consistently outperform cheaper ones, nominally cheaper models can incur
higher cost due to verbose output, and an earlier chat-oriented router of the
same architecture trails Switchcraft by 5.3\,pp
(Appendix~\ref{app:old-model-basket}).

To the best of our knowledge, this is the first router fine-tuned and
evaluated for strict tool-call correctness in agentic workloads. We make three
main contributions:
\begin{itemize}
\item We formulate model routing for agentic use cases and identify key
  challenges.
\item We build a unified evaluation framework for agentic routing,
  including multi-benchmark normalization, corrected ground-truth
  annotations, an AST-based comparison framework for robust
  tool-calling evaluation, and metrics that separate answer, routing, and
  decisive accuracy.
\item We develop Switchcraft, a calibrated DistilBERT-based model router
  that runs inline in milliseconds, can abstain, and exceeds the best
  individual model's answer accuracy at 59\% lower cost,
  remaining cost-effective on a held-out benchmark and under KV caching.
\end{itemize}

\section{Motivation}
\label{sec:motivation}

\begin{figure*}[t]
\centering
\small
\setlength{\fboxsep}{1pt}
\newlength{\mexlabelw}
\settowidth{\mexlabelw}{\textbf{Wrong parameters:}\enspace}

\begin{tikzpicture}[
    node distance=0.2cm,
    box/.style={rectangle, rounded corners=2pt, text width=\linewidth-6.4pt, align=left, inner sep=3pt},
    turnbox/.style={box, draw=gray!60, fill=gray!5},
    callbox/.style={box, draw=blue!50, fill=blue!5},
    errorbox/.style={box, draw=red!50, fill=red!5},
]

\node[anchor=west, text width=\linewidth-6.4pt, inner sep=3pt] (title) at (0, 0) {\textbf{Available tools:} get\_symbol\_by\_name, add\_to\_watchlist, get\_stock\_info, place\_order, get\_order\_details, send\_message, ...};

\node[turnbox, below=0.08cm of title.south west, anchor=north west] (t1) {
\textbf{Turn 1 -- User:} ``Integrate stock for `Omega Industries' into my watchlist effectively.''
};

\node[callbox, below=0.06cm of t1.south west, anchor=north west] (c1) {
\textbf{Correct:}\enspace get\_symbol\_by\_name(name="Omega Industries") $\rightarrow$ \{"symbol": "OMEG"\} \\
\phantom{\textbf{Correct:}}\enspace add\_to\_watchlist(stock="OMEG")
};

\node[turnbox, below=0.08cm of c1.south west, anchor=north west] (t3) {
\textbf{Turn 3 -- User:} ``Execute a transaction for 150 shares at the present market value for the stock we just added.''
};

\node[callbox, below=0.06cm of t3.south west, anchor=north west] (c3) {
\textbf{Correct:}\enspace get\_stock\_info(symbol="OMEG") $\rightarrow$ \{"price": 457.23, ...\} \\
\phantom{\textbf{Correct:}}\enspace place\_order(order\_type="Buy", symbol="OMEG", price=457.23, amount=150)
};

\node[errorbox, below=0.08cm of c3.south west, anchor=north west] (e1) {
\makebox[\mexlabelw][l]{\textbf{Wrong function:}}place\_order(order\_type="Buy", symbol="OMEG", price=557.23, amount=150) \hspace{0.2cm}{\color{red!70!black}\textbf{$\times$}} \\
{\color{red!70!black}Skipping get\_stock\_info and using a hallucinated or stale price causes the order to be placed at wrong price.}
};

\node[errorbox, below=0.06cm of e1.south west, anchor=north west] (e2) {
\makebox[\mexlabelw][l]{\textbf{Wrong parameters:}}place\_order(order\_type=\colorbox{red!15}{"Sell"}, symbol="OMEG", price=457.23, amount=150) \hspace{0.2cm}{\color{red!70!black}\textbf{$\times$}} \\
{\color{red!70!black}Selling instead of buying: a single wrong parameter value causes the \emph{opposite} of the intended action.}
};

\node[errorbox, below=0.06cm of e2.south west, anchor=north west] (e3) {
\makebox[\mexlabelw][l]{\textbf{Wrong value:}}place\_order(order\_type="Buy", symbol="OMEG", price=457.23, amount=\colorbox{red!15}{1500}) \hspace{0.2cm}{\color{red!70!black}\textbf{$\times$}} \\
{\color{red!70!black}An order-of-magnitude error in amount causes a 10$\times$ larger financial commitment.}
};

\end{tikzpicture}
\caption{Motivating example from BFCL v3~\cite{bfcl}. Agentic queries demand \emph{correct functions} and \emph{precise parameter values} at every step; small errors (wrong type, wrong value, wrong order) produce consequential failures.}
\label{fig:motivation-example}
\end{figure*}

Figure~\ref{fig:motivation-example} illustrates why routing for agentic tool
calling differs from routing for chat completion. A user asks an AI agent to
add ``Omega Industries'' to a watchlist and place a market order; fulfilling
this request requires a \emph{sequence} of tool invocations (resolve ticker,
add to watchlist, fetch price, place order) where each step depends on the
previous one. Unlike chat, where a paraphrase may be acceptable, agentic
errors compound and can be irreversible: hallucinating a price puts the
order at the wrong value; confusing \texttt{"Buy"} with \texttt{"Sell"}
executes the \emph{opposite} of the user's intent; a single-digit error in
\texttt{amount} causes a tenfold larger financial commitment. At the same
time, not all variation is an error---independent parallel calls can be
issued in any order, and string parameters often admit semantically
equivalent forms (e.g., \texttt{"Illinois"} vs.\ \texttt{"IL"}); a correct
evaluator must accept these (full description in
Appendix~\ref{app:motivation-variation}). These properties -- sequential
dependencies, strict parameter precision, and selective tolerance for
variation -- motivate a router fine-tuned on agentic data with metrics that
capture the correctness criteria of every tool call, not just the final answer.

\section{Design}
\label{sec:design}

We explored a large space of architectures, input representations, scoring
methods, and routing strategies (frozen-embedding MLPs, larger encoders,
LLM-as-a-router, similarity routing, cost-weighted losses, BLEU and
LLM-as-judge scoring); rejected alternatives are detailed in
Appendix~\ref{app:design-space}. Figure~\ref{fig:system-architecture} shows
the final architecture: a fine-tuning pipeline that ingests function-calling
benchmarks, runs every query against every candidate LLM, scores outputs via
AST comparison, and distills the resulting preference data into a lightweight
DistilBERT classifier; and an inference pipeline that runs the calibrated
classifier plus a cost model to select the cheapest predicted-correct LLM, or to
abstain when no LLM is likely to answer correctly.
We build a router specialized for function calling to avoid diluting it
across heterogeneous workloads such as chat. Dispatching to the correct router is trivial --
function-calling requests carry a \texttt{tools} parameter.

\begin{figure*}[t]
\centering
\begin{tikzpicture}[
    node distance=0.5cm and 0.8cm,
    box/.style={rectangle, draw, rounded corners, minimum width=1.6cm, minimum height=0.5cm, align=center, font=\scriptsize},
    dataset/.style={box, fill=blue!15, minimum width=1.1cm},
    process/.style={box, fill=orange!20},
    model/.style={box, fill=green!20, minimum width=1.2cm, minimum height=0.4cm},
    output/.style={box, fill=purple!15},
    routerbox/.style={rectangle, draw, rounded corners, fill=green!10, minimum width=2.4cm, minimum height=1.2cm, align=center},
    innerbox/.style={rectangle, draw, rounded corners, fill=green!25, minimum width=2.0cm, minimum height=0.4cm, align=center, font=\scriptsize},
    arrow/.style={->, >=stealth, semithick},
    dashedarrow/.style={->, >=stealth, semithick, dashed},
    lbl/.style={font=\tiny, align=center}
]

\node[font=\scriptsize\bfseries] at (-3.6, 2.7) {Fine-Tuning Pipeline};

\node[dataset] (bfcl) at (-5.4, 2.0) {BFCL};
\node[dataset] (confetti) at (-4.2, 2.0) {ConFETTI};
\node[dataset] (xlam) at (-3.0, 2.0) {xLAM};
\node[dataset] (hermes) at (-1.8, 2.0) {Hermes};
\node[dataset] (funreason) at (-0.3, 2.0) {FunReason-MT};

\node[process] (ingest) at (-3.0, 1.0) {Data Ingestion};

\node[process] (inference) at (-5.4, 0.1) {LLM Inference};
\node[model, font=\tiny] (llm1) at (-5.4, -0.6) {GPT-5.4-nano};
\node[model, font=\tiny] (llm2) at (-4.1, -0.6) {Kimi-K2.5};
\node[model, font=\tiny] (llm3) at (-2.9, -0.6) {GPT-5.4};
\node[model, font=\tiny, minimum width=0.6cm] (llm4) at (-2.0, -0.6) {...};

\node[process] (ast) at (-3.0, -1.4) {AST Scoring};

\node[process] (traindata) at (-4.5, -2.3) {Preference Data};
\node[process] (train) at (-1.5, -2.3) {Router Fine-Tuning\\+ Calibration};

\foreach \d in {bfcl, confetti, xlam, hermes, funreason} {
    \draw[arrow] (\d) -- (ingest);
}
\draw[arrow] (ingest) -| (inference);
\draw[arrow] (inference) -- (llm1);
\draw[arrow] (inference) -- (llm2);
\draw[arrow] (inference) -- (llm3);
\draw[arrow] (inference) -- (llm4);
\draw[arrow] (llm1.south) -- ++(0,-0.1) -| ([xshift=-0.3cm]ast.north);
\draw[arrow] (llm3.south) -- ++(0,-0.1) -| ([xshift=0.3cm]ast.north);
\draw[arrow] (ingest) -- (ast);
\draw[arrow] (ast) -| (traindata);
\draw[arrow] (traindata) -- (train);

\node[font=\scriptsize\bfseries] at (4.5, 2.7) {Inference Pipeline};

\node[output] (query) at (4.5, 2.0) {Agent Query};

\node[routerbox, minimum height=1.4cm, minimum width=3.0cm] (agentic_inf) at (4.5, 0.7) {};
\node[font=\scriptsize\bfseries] at (4.5, 1.15) {Agentic Router};
\node[innerbox, minimum width=2.6cm] (classifier_inf) at (4.5, 0.7) {Calibrated DistilBERT};
\node[innerbox, minimum width=2.6cm] (costmodel_inf) at (4.5, 0.2) {Cost model};

\node[process] (select) at (4.5, -0.7) {Select cheapest\\predicted-correct LLM};
\node[output, font=\tiny, minimum width=1.2cm] (abstain) at (6.75, -0.7) {Abstain\\(no LLM call)};

\node[model, font=\tiny] (inf1) at (3.1, -1.7) {GPT-5.4-nano};
\node[model, font=\tiny] (inf2) at (4.2, -1.7) {Kimi-K2.5};
\node[model, font=\tiny] (inf3) at (5.3, -1.7) {GPT-5.4};
\node[model, font=\tiny, minimum width=0.6cm] (inf4) at (6.2, -1.7) {...};

\node[output, minimum width=4.2cm] (response) at (4.5, -2.6) {Tool Call Response};

\draw[arrow] (query) -- (agentic_inf);
\draw[arrow] (agentic_inf) -- (select);
\draw[dashedarrow] (select) -- (abstain);
\draw[arrow] (select) -- (inf1);
\draw[arrow] (select) -- (inf2);
\draw[arrow] (select) -- (inf3);
\draw[arrow] (select) -- (inf4);
\draw[arrow] (inf1) -- (inf1 |- response.north);
\draw[arrow] (inf2) -- (inf2 |- response.north);
\draw[arrow] (inf3) -- (inf3 |- response.north);
\draw[arrow] (inf4) -- (inf4 |- response.north);

\draw[dashedarrow] (train.east) -- node[above, font=\scriptsize] {deploy} (agentic_inf.west);

\end{tikzpicture}
\caption{System architecture. \textbf{Left:} fine-tuning pipeline (ingest in-distribution benchmarks, run inference across LLMs, score via AST, fine-tune and calibrate router). \textbf{Right:} inference-time routing (calibrated DistilBERT classifier and cost model select the cheapest predicted-correct LLM, or abstain when no LLM is likely to answer correctly).}
\label{fig:system-architecture}
\end{figure*}

\subsection{Input Representation}
\label{sec:input-representation}

A key design choice is how to pack an agentic query -- multi-turn conversation,
tool definitions, and metadata -- into DistilBERT's 512-token window. We pack
in five steps:

\begin{enumerate}
    \item \textbf{Latest user turn} (immediate intent, always included).
    \item \textbf{Tool signatures} converted to compact
          \texttt{func\_name(param1, param2)} form (descriptions and
          JSON-schema boilerplate stripped, etc.;
          capped at 100 subword tokens with \texttt{[truncated]}).
    \item \textbf{Earlier turns} added greedily in reverse chronological order
          until the budget is exhausted, prioritizing recent context.
    \item \textbf{Metadata preamble} of three numeric features (\texttt{length, num\_tools, num\_turns})
          for complexity-aware routing without relying solely on text.
    \item \textbf{Concatenate and tokenize} (512-token truncation, dynamic padding).
\end{enumerate}

\subsection{Scoring the Tool-Calls}
\label{sec:ast-checker}

Executing each LLM-issued tool call is infeasible at our scale: queries span
thousands of distinct, often private third-party APIs. Instead, we label each
call statically by comparing its abstract syntax tree (AST)---function name,
argument names, types, and values---to that of the ground-truth call. The
exception is BFCL's multi-turn categories, whose environments BFCL simulates:
we execute their calls and score the end-of-turn state, which credits
recovery from failed calls within a turn
(Appendix~\ref{app:multi-turn-scoring}).

The AST checker determines which (model,~query) pairs the router sees as
``correct''. However, designing a general scorer is surprisingly hard: (1) the
same call admits many semantically equivalent forms (set- vs.\ list-valued
arguments, omitted default values, alternate string encodings), (2) tool-call
arguments can be arbitrarily nested objects that must be compared structurally,
not by direct equality, and (3) the same syntactic shape across
benchmarks (e.g., a list of lists) can carry different meanings.

\begin{table*}[t]
	\centering
	\caption{AST checker comparison on GPT-5.3-chat results. \emph{FN}: our checker accepts but BFCL rejects; \emph{FP}: inverse. \textsuperscript{$\dagger$}~Single-turn BFCL categories only (Simple, Multiple, Parallel, Parallel+Multiple, and their \emph{Live} variants); multi-turn categories are excluded because they are scored by execution rather than AST matching.}
	\label{tab:ast_checker_comparison}
	\small
	\setlength{\tabcolsep}{6pt}
	\begin{tabular}{lrrrrr}
		\toprule
		\multirow{2}{*}{\textbf{Dataset}} & \multirow{2}{*}{\textbf{\# Entries}}
		& \multicolumn{2}{c}{\textbf{Accuracy (\%)}}
		& \multirow{2}{*}{\textbf{FN}} & \multirow{2}{*}{\textbf{FP}} \\
		\cmidrule(lr){3-4}
		&        & \textbf{BFCL}   & \textbf{Ours}              &        &     \\
		\midrule
		BFCL~\cite{bfcl}\textsuperscript{$\dagger$}            & 2{,}351   & 84.13 & \textbf{86.73} &     67 &   6 \\
		Glaive~\cite{glaive_fc}     & 83{,}814  & 43.99 & \textbf{88.46} & 37{,}277 &   2 \\
		ConFETTI~\cite{confetti}	& 506    &  0.00 & \textbf{53.16} &    269 &   0 \\
		xLAM-60K~\cite{xlam}        & 60{,}000  & 17.35 & \textbf{82.32} & 39{,}302 & 323 \\
		Hermes~\cite{hermes3}       & 8{,}940   & 59.73 & \textbf{86.49} &  2{,}531 & 139 \\
		\midrule
		\textbf{Overall}    		& 155{,}611 & 35.08 & \textbf{85.84} & 79{,}446 & 470 \\
		\bottomrule
	\end{tabular}
\end{table*}

For these reasons, the existing AST checker from BFCL~\cite{bfcl} handles its
native benchmark well but breaks on every other one: scoring the same
GPT-5.3-chat outputs yields much lower accuracies on the four non-BFCL datasets
(Table~\ref{tab:ast_checker_comparison}). Through a systematic study of rejected
calls across all five datasets, we identify recurring classes of bias in BFCL's
AST checker; we summarize each below, with examples in
Appendix~\ref{app:ast-checker}
(Tables~\ref{tab:ast_checker_examples}--\ref{tab:ast_checker_examples_nested}).

\begin{enumerate}
	\item \textbf{Array order sensitivity.} BFCL compares arrays
	element-wise, rejecting set-valued parameters when the model
	emits them in a different order than the ground truth. We
	compare flat arrays as multisets and lists of dictionaries as sets.
	\item \textbf{No default-parameter awareness.} BFCL treats every
	parameter listed in the ground truth as required. We parse defaults
	from the tool description and accept an omission when the documented
	default matches the ground truth.
	\item \textbf{Brittle string matching.} Differences in case, whitespace,
	punctuation, and ISO-8601 timestamp formatting cause spurious
	mismatches. We canonicalize each string and fall back to DistilBERT
	cosine similarity (threshold 0.85) for multi-word strings.
	\item \textbf{No nested-structure handling.} BFCL does not recognize
	the JSON-Schema \texttt{"object"} type and aborts the entry; even when
	fixed, it has no recursive comparator. We add the missing types and
	recursively validate each nested object against its sub-schema.
	\item \textbf{Ambiguous array semantics.} A list-of-lists in the ground
	truth can mean either a list of \emph{alternative} valid values or a
	true \emph{nested} array. We use the tool-call schema to disambiguate.
\end{enumerate}

With our AST checker, the same models show much more consistent accuracy across
all five datasets (Table~\ref{tab:ast_checker_comparison}). We validate it via
unit tests, manual review of random entries, and LLM-as-judge scoring.

\subsection{Routing Logic}
\label{sec:routing-logic}

At inference time, the router sits on the agent's critical path and must
decide within a 20\,ms budget (Section~\ref{sec:latency}), too little to
consult an LLM (Appendix~\ref{app:design-space}). After one encoder forward
pass, it selects one LLM per query or abstains, in three lightweight stages:

\textbf{Stage 1: calibrated correctness prediction.}
The DistilBERT classifier outputs a logit for each of the $K$ candidates.
Per-model Platt scaling maps each logit to a calibrated probability
$p_m$ that model $m$ answers the query correctly, so one threshold means
the same for every model.

\textbf{Stage 2: abstention.}
We train Switchcraft to abstain: an abstention head---a logistic regression
over $(p_1,\dots,p_K)$---estimates the probability $p_\varnothing$ that
\emph{no} candidate answers correctly. If $p_\varnothing \geq \tau_a$, the
router does not dispatch any LLM call and returns the query to the agent, which can
ask the user for clarification or escalate to the most expensive model nonetheless.
14.3\% of our test queries defeat every model.

\textbf{Stage 3: cost-aware selection.}
Otherwise, among models with $p_m \geq \tau_r$, the router selects the one
with the \textbf{lowest profiled cost}: the average dollar cost from
the token counts observed when each model answered our
benchmark queries, multiplied by its per-million-token list prices
(Table~\ref{tab:main-results}; full formula in Appendix~\ref{app:cost-model}).
This captures \emph{chattiness}: a model that generates extensive
reasoning or verbose output accumulates a higher profiled cost than a concise
model at the same per-token rate (Section~\ref{sec:chattiness}). If no model clears $\tau_r$, the router falls
back to the likeliest model (argmax).

\textbf{Calibration and policy tuning.}
We split the validation set into two halves, stratified by dataset and by
query difficulty (trivial, decisive, or impossible;
Section~\ref{sec:eval-setup}). The \emph{calibration} half fits the Platt
scalers and the abstention head. The \emph{selection} half picks the
fine-tuning checkpoint and tunes $(\tau_r, \tau_a)$ on a 0.05-step grid,
maximizing decisive accuracy subject to an average profiled-cost budget and a
minimum fraction of impossible queries abstained on. Separate halves keep
the thresholds from overfitting the calibration data. The test set is never
used here.

\textbf{Oracle router.}
Our oracle upper bound has perfect hindsight information: for all queries, including
the test set, it knows which models answer correctly and selects the cheapest
correct one by realized per-query cost. When no model is correct, the oracle
abstains at zero cost, providing a tight upper bound.

\section{Evaluation}
\label{sec:evaluation}

We evaluate Switchcraft against individual LLMs, random, kNN, MIRT,
heuristic, and oracle routers.

\subsection{Datasets}
\label{sec:datasets}

We combine six function-calling benchmarks spanning a broad range of
tool-calling complexity (Table~\ref{tab:datasets}); diversity avoids
overfitting to any single benchmark. Five form our
in-distribution corpus (99{,}125 examples, 97{,}369 after deduplication, in
14 splits). The sixth, Glaive, is held out entirely as an unseen
out-of-distribution (OOD) test (Section~\ref{sec:main-results}): it is large and mostly
easy (74\% of its queries are solved by every model), so pooling it would dominate
in-distribution accuracy. FunReason-MT~\cite{funreason_mt,hardgen} instead
contributes harder multi-turn queries.

\begin{table*}[t]
\centering
\caption{Datasets and splits used in our evaluation. \emph{ST} = single-turn,
\emph{MT} = multi-turn, \emph{par} = parallel calls.}
\label{tab:datasets}
\small
\begin{tabular}{llrr}
\toprule
\textbf{Source} & \textbf{Split} & \textbf{Examples} & \textbf{Type} \\
\midrule
\multirow{10}{*}{BFCL~v3~\cite{bfcl}}
  & Simple                 &    400 & ST \\
  & Multiple               &    200 & ST \\
  & Parallel               &    200 & ST, par \\
  & Parallel+Multiple      &    200 & ST, par \\
  & Live Simple            &    258 & ST \\
  & Live Multiple          &  1{,}053 & ST \\
  & Live Parallel          &     16 & ST, par \\
  & Live Par.+Multiple     &     24 & ST, par \\
  & Multi-turn Base (per turn) &  745 & MT \\
  & Multi-turn Long Ctx (per turn) & 745 & MT \\
\midrule
ConFETTI~\cite{confetti}
  & Conversations (cleaned) &   506 & MT \\
\midrule
xLAM-60K~\cite{xlam}
  & Function Calling 60K   & 60{,}000 & ST \\
\midrule
Hermes~\cite{hermes3}
  & Function Calling v1    &  8{,}940 & ST/MT \\
\midrule
FunReason-MT~\cite{funreason_mt,hardgen}
  & HardGen (per turn)     & 25{,}838 & MT \\
\midrule
\multicolumn{4}{l}{\emph{Held out (OOD test only)}} \\
Glaive~\cite{glaive_fc}
  & Function Calling v2    & 83{,}814 & MT \\
\bottomrule
\end{tabular}
\end{table*}

The corpus spans synthetic and human-authored data, single- and multi-turn
conversations. We decompose multi-turn conversations into
per-turn evaluation records, which is reflected in Table~\ref{tab:datasets}.
FunReason-MT's records come from 13{,}530 trajectories in BFCL-style
environments (file system, messaging, math) with a median of 32 tools per
query; as its turns are ordered, state-mutating call sequences, we score them
with an order-sensitive variant of our AST checker.
Several datasets required substantial
cleaning---most notably ConFETTI, where we corrected 27\% of entries
(Appendix~\ref{app:confetti})---and format normalization to a common
BFCL-style JSONL schema (Appendix~\ref{app:per-dataset}). In-distribution
datasets pass through a unified pipeline: deduplication on (query, tools),
per-dataset stratified 80/10/10 splits, and multi-label annotation by running
each query through all candidate LLMs and recording correctness per our AST
framework (Section~\ref{sec:ast-checker}). Glaive is annotated identically
but never split.

\subsection{Experimental Setup}
\label{sec:eval-setup}

We route among eight LLMs spanning three model families
(Table~\ref{tab:main-results}), accessed through API endpoints (default
temperature; all values in USD).
For every dataset split, including Glaive,
we run every query once against all eight models using the OpenAI-compatible
function-calling API (tool definitions via the native \texttt{tools}
parameter, no system prompt) and score outputs with our AST framework.
Single-run labels are mostly stable:
over 20 runs on an earlier basket, 78--90\% of each model's outcomes never
changed, and soft labels gave no clear gain (Appendix~\ref{app:probabilistic}).
Per-dataset stratified 80/10/10 splits of the in-distribution corpus yield
77{,}891 fine-tuning, 9{,}732 validation, and 9{,}746 held-out test queries.
We fine-tune DistilBERT-base-uncased~\cite{distilbert} (67M parameters)
end-to-end as a multi-label classifier producing eight logits (one per
candidate LLM) from the input representation in
Section~\ref{sec:input-representation}. To test whether a larger encoder or a
longer input helps, we fine-tune ModernBERT-base~\cite{modernbert} (150M, 8K
tokens) and DeBERTa-v3-base~\cite{debertav3} (184M, 512 tokens) under the same
protocol; unless noted, Switchcraft refers to the DistilBERT router, our
default as the smallest and fastest (Section~\ref{sec:latency}). Each encoder
uses 20 random seeds that vary both initialization and data split
(hyperparameters in Appendix~\ref{app:training-config}; DistilBERT's
fine-tuning curves in Appendix~\ref{app:training-curves}). For each seed, we
calibrate and tune the policy on the two validation halves
under a budget of $17 \times 10^{-4}$\,\$ per
query in profiled cost---about
a quarter of GPT-5.3-chat's---and a 55\% floor on impossible queries abstained
on; we report test results for the seed with the highest validation answer
accuracy. Router parameter counts refer to the full fine-tuned model,
including token embeddings and the classification head.

\textbf{Baselines.} We compare against (i) \textbf{single-model} routers
(eight baselines); (ii) a \textbf{random} router that picks a candidate
uniformly per query; (iii) a \textbf{kNN} router that embeds each query with a
frozen DistilBERT encoder, retrieves its $k \in \{1, 10\}$ most similar
training queries by cosine similarity, and selects the cheapest model correct
on all neighbors (else the cheapest of those correct most often);
(iv) an \textbf{MIRT} router~\cite{irt_router} that predicts each model's
success from frozen DistilBERT embeddings of the query and of a short model
description, and selects the cheapest model with predicted success of at
least 0.5 (else the likeliest; Appendix~\ref{app:mirt});
(v) \textbf{heuristic} routers using a single feature (input token length,
number of tool definitions, or conversation turns) with thresholds profiled
on training data; and (vi) an \textbf{oracle} that selects the cheapest correct
model per query and abstains at zero cost when none is correct.

\textbf{Metrics.} Since the benchmarks mix easy and impossible queries,
we report three accuracies. \textbf{Answer accuracy} is the fraction of
queries whose selected model answers correctly; abstention counts as wrong.
\textbf{Routing accuracy} additionally credits abstaining on an
\emph{impossible} query, which no candidate model answers correctly; it equals
answer accuracy for methods that never abstain (all but Switchcraft and the
oracle). \textbf{Decisive accuracy} is
answer accuracy on the \emph{decisive} subset---queries that some candidates
answer correctly and others do not, the only ones on which model choice
matters (58.8\% of the test set; 26.9\% are \emph{trivial}, solved by every
model, and 14.3\% impossible). \textbf{Cost} is the realized average dollar
cost per query; abstentions cost zero (Appendix~\ref{app:additional-limitations}).

\subsection{Main Results: Accuracy--Cost Trade-Offs}
\label{sec:main-results}

Table~\ref{tab:main-results} presents in-distribution results on the held-out
test set, and Figure~\ref{fig:pareto} plots routing accuracy against cost in
distribution~(a) and on held-out Glaive~(b). The cost column reports the
actual API-billed dollar cost per query; a token-level decomposition of these
costs and an analysis of model chattiness are given in
Section~\ref{sec:chattiness}.

\begin{table*}[t]
\centering
\caption{In-distribution results on the held-out test set (9{,}746 queries).
In/Out are list prices; cost is the realized average per query (abstention
costs zero). Answer accuracy counts abstention as wrong, routing accuracy
credits abstention on impossible queries, and decisive accuracy is answer
accuracy on queries that some but not all models answer correctly. For our
routers we report the validation-selected seed ($\pm$: std of test answer
accuracy over 20 seeds; Appendix~\ref{app:seed-stability}); the three
encoders are within seed variance.}
\label{tab:main-results}
\small
\setlength{\tabcolsep}{3.5pt}
\begin{tabular}{ll@{\hspace{2pt}}rrcccc}
\toprule
\textbf{Type} & \textbf{Entity} & \textbf{In} & \textbf{Out}
  & \textbf{Answer} & \textbf{Routing} & \textbf{Decisive} & \textbf{Cost} \\
 & & \multicolumn{2}{c}{\textbf{(\$/M)}} & \textbf{(\%)} & \textbf{(\%)}
  & \textbf{(\%)} & \textbf{($10^{-4}$\,\$)} \\
\midrule
\multirow{8}{*}{Single LLM}
  & GPT-5.3-chat  & 1.75  & 14.00  & 75.63 & 75.63 & 82.87 & 66.2 \\
  & GPT-5.4       & 2.50  & 15.00  & 74.62 & 74.62 & 81.15 & 113.3 \\
  & GPT-5.4-mini  & 0.75  &  4.50  & 72.46 & 72.46 & 77.49 & 33.3 \\
  & GPT-5-mini    & 0.25  &  2.00  & 72.29 & 72.29 & 77.19 & 12.8 \\
  & GPT-5.4-nano  & 0.20  &  1.25  & 70.02 & 70.02 & 73.33 & 9.5 \\
  & GPT-5-nano    & 0.05  &  0.40  & 67.75 & 67.75 & 69.48 & 2.9 \\
  & Qwen-3.5-9B   & 0.05  &  0.15  & 64.94 & 64.94 & 64.70 & 3.5 \\
  & Kimi-K2.5     & 0.60  &  3.00  & 36.39 & 36.39 & 16.18 & 16.1 \\
\midrule
\multirow{4}{*}{Baseline}
  & kNN ($k{=}10$) & \multicolumn{2}{c}{---} & 76.49 & 76.49 & 84.34 & 30.0 \\
  & kNN ($k{=}1$)  & \multicolumn{2}{c}{---} & 70.89 & 70.89 & 74.82 & 4.7 \\
  & MIRT           & \multicolumn{2}{c}{---} & 70.68 & 70.68 & 74.45 & 11.6 \\
  & Random         & \multicolumn{2}{c}{---} & 66.89 & 66.89 & 68.02 & 29.5 \\
\midrule
\multirow{3}{*}{Heuristic}
  & Num.\ Turns   & \multicolumn{2}{c}{---} & 76.13 & 76.13 & 83.73 & 82.6 \\
  & Num.\ Tools   & \multicolumn{2}{c}{---} & 73.05 & 73.05 & 78.48 & 72.9 \\
  & Length        & \multicolumn{2}{c}{---} & 67.11 & 67.11 & 68.40 & 5.8 \\
\midrule
\multirow{3}{*}{\textbf{Ours}}
  & ModernBERT (150M)  & \multicolumn{2}{c}{---}
    & 78.38\,{\scriptsize($\pm$0.36)} & 86.95 & 87.60 & 36.1 \\
  & DistilBERT (67M)   & \multicolumn{2}{c}{---}
    & 78.27\,{\scriptsize($\pm$0.46)} & 86.40 & 87.39 & 27.5 \\
  & DeBERTa-v3 (184M)$^\dagger$ & \multicolumn{2}{c}{---}
    & 78.58\,{\scriptsize($\pm$0.60)} & 87.48 & 87.79 & 27.0 \\
\midrule
Upper Bound
  & Oracle         & \multicolumn{2}{c}{---} & 85.71 & 100.00 & 100.00 & 4.0 \\
\bottomrule
\end{tabular}
\par\smallskip
\parbox{\linewidth}{\footnotesize $^\dagger$Validation selects seed 247,
  whose test partition differs; on it, GPT-5.3-chat scores 76.17\% and the
  oracle 85.77\%.}
\end{table*}

\begin{figure*}[t]
\centering
\begin{tikzpicture}
\begin{groupplot}[
    group style={group size=2 by 1, horizontal sep=1.25cm},
    scale only axis,
    width=0.4\linewidth,
    height=5.0cm,
    xmode=log,
    log ticks with fixed point,
    grid=major,
    grid style={gray!20},
    tick label style={font=\scriptsize},
    label style={font=\small},
    title style={font=\small},
    xlabel={Avg cost per query ($10^{-4}$\,\$, log scale)},
    clip=false,
]
\nextgroupplot[
    title={(a) In-distribution test (9{,}746 queries)},
    ylabel={Routing accuracy (\%)},
    xmin=2, xmax=250, ymin=62, ymax=100,
    xtick={2,5,10,20,50,100,200},
    legend columns=8,
    legend to name=paretolegend,
    legend style={font=\scriptsize, draw=none, column sep=3pt},
]
\addplot[only marks, mark=*, mark size=1.8pt, blue!70!black]
    coordinates {(66.2,75.63) (113.3,74.62) (33.3,72.46) (12.8,72.29)
                 (9.5,70.02) (2.9,67.75) (3.5,64.94)};
\addlegendentry{Single LLM}
\addplot[only marks, mark=triangle*, mark size=2.4pt, violet]
    coordinates {(82.6,76.13) (72.9,73.05) (5.8,67.11)};
\addlegendentry{Heuristic}
\addplot[only marks, mark=diamond*, mark size=2.4pt, orange!85!black]
    coordinates {(30.0,76.49) (4.7,70.89)};
\addlegendentry{kNN}
\addplot[only marks, mark=pentagon*, mark size=2.4pt, brown!80!black]
    coordinates {(11.6,70.68)};
\addlegendentry{MIRT}
\addplot[only marks, mark=+, mark size=2.8pt, very thick, teal]
    coordinates {(29.5,66.89)};
\addlegendentry{Random}
\addplot[red!80!black, densely dashed, thin, mark=o, mark size=1.2pt,
         mark options={solid}, forget plot]
    coordinates {(3.5,78.93) (7.9,80.87) (12.6,83.17) (19.8,84.88)
                 (27.5,86.40) (37.7,87.37) (43.6,87.88)};
\addplot[only marks, mark=square*, mark size=2.2pt, red!80!black]
    coordinates {(27.5,86.40)};
\addlegendentry{Switchcraft}
\addlegendimage{red!80!black, densely dashed, thin, mark=o, mark size=1.2pt,
                mark options={solid}}
\addlegendentry{Budget sweep}
\addplot[only marks, mark=star, mark size=3.2pt, black, thick]
    coordinates {(4.0,100)};
\addlegendentry{Oracle}
\draw[red!80!black, thin, |-|] (axis cs:15.4,86.40) -- (axis cs:41.4,86.40);
\draw[red!80!black, thin, |-|] (axis cs:27.5,85.52) -- (axis cs:27.5,87.83);
\node[font=\tiny\bfseries, red!80!black, anchor=south] at (axis cs:27.5,88.0) {Switchcraft};
\node[font=\tiny, anchor=east] at (axis cs:27,76.49) {kNN ($k{=}10$)};
\node[font=\tiny, anchor=north, inner sep=0.5pt] at (axis cs:66.2,75.2) {GPT-5.3-chat};
\node[font=\tiny, anchor=west] at (axis cs:90,76.13) {Num.~turns};
\node[font=\tiny, anchor=west] at (axis cs:125,74.62) {GPT-5.4};
\node[font=\tiny, anchor=west] at (axis cs:80,73.05) {Num.~tools};
\node[font=\tiny, anchor=north west] at (axis cs:30,71.9) {GPT-5.4-mini};
\node[font=\tiny, anchor=south] at (axis cs:12.8,72.7) {GPT-5-mini};
\node[font=\tiny, anchor=west] at (axis cs:10.3,69.2) {GPT-5.4-nano};
\node[font=\tiny, anchor=west] at (axis cs:12.6,70.68) {MIRT};
\node[font=\tiny, anchor=south] at (axis cs:4.7,71.3) {kNN ($k{=}1$)};
\node[font=\tiny, anchor=south west] at (axis cs:2.1,68.1) {GPT-5-nano};
\node[font=\tiny, anchor=west] at (axis cs:6.4,67.11) {Length};
\node[font=\tiny, anchor=west] at (axis cs:3.9,64.94) {Qwen-3.5-9B};
\node[font=\tiny, anchor=west] at (axis cs:33,66.89) {Random};
\node[font=\tiny, anchor=west] at (axis cs:4.6,98.6) {Oracle};
\node[font=\tiny, gray, anchor=south east] at (rel axis cs:0.98,0.02)
    {Kimi-K2.5 (36.4\%) off-scale};
\nextgroupplot[
    title={(b) Held-out Glaive (83{,}814 queries)},
    xmin=0.25, xmax=20, ymin=83, ymax=100,
    xtick={0.3,1,3,10},
]
\addplot[red!80!black, densely dashed, thin, mark=o, mark size=1.2pt,
         mark options={solid}]
    coordinates {(0.42,87.13) (0.48,87.49) (0.65,88.24) (0.90,89.38)
                 (1.46,89.39) (1.97,89.49) (4.16,89.58)};
\addplot[only marks, mark=*, mark size=1.8pt, blue!70!black]
    coordinates {(10.27,88.53) (7.58,88.46) (3.06,88.04) (0.42,86.98)
                 (3.95,86.89) (0.39,86.26) (0.82,85.47) (1.77,83.94)};
\addplot[only marks, mark=triangle*, mark size=2.4pt, violet]
    coordinates {(8.59,88.20) (0.42,86.98) (0.48,86.87)};
\addplot[only marks, mark=diamond*, mark size=2.4pt, orange!85!black]
    coordinates {(1.27,89.43) (0.46,87.24)};
\addplot[only marks, mark=pentagon*, mark size=2.4pt, brown!80!black]
    coordinates {(0.48,87.05)};
\addplot[only marks, mark=+, mark size=2.8pt, very thick, teal]
    coordinates {(3.52,86.83)};
\addplot[only marks, mark=square*, mark size=2.2pt, red!80!black]
    coordinates {(1.46,89.39)};
\addplot[only marks, mark=star, mark size=3.2pt, black, thick]
    coordinates {(0.37,100)};
\node[font=\tiny\bfseries, red!80!black, anchor=south west] at (axis cs:1.40,89.7) {Switchcraft};
\node[font=\tiny, anchor=north west] at (axis cs:1.1,89.15) {kNN ($k{=}10$)};
\node[font=\tiny, anchor=south west] at (axis cs:6.9,88.75) {GPT-5.3-chat};
\node[font=\tiny, anchor=west] at (axis cs:11,88.53) {GPT-5.4};
\node[font=\tiny, anchor=north] at (axis cs:8.59,88.0) {Num.~turns};
\node[font=\tiny, anchor=north east] at (axis cs:3.0,87.85) {GPT-5.4-mini};
\node[font=\tiny, anchor=north west] at (axis cs:4.0,86.65) {Kimi-K2.5};
\node[font=\tiny, anchor=north east] at (axis cs:3.45,86.6) {Random};
\node[font=\tiny, anchor=west] at (axis cs:1.9,83.94) {GPT-5-mini};
\node[font=\tiny, anchor=west] at (axis cs:0.9,85.47) {GPT-5.4-nano};
\draw[gray, thin] (axis cs:0.405,87.35) -- (axis cs:0.405,91.0);
\node[font=\tiny, anchor=south west, align=left] at (axis cs:0.30,91.0)
    {GPT-5-nano, Qwen, MIRT,\\kNN ($k{=}1$), Length,\\Num.~tools};
\node[font=\tiny, anchor=west] at (axis cs:0.42,99.4) {Oracle};
\end{groupplot}
\node[anchor=north] at ($(group c1r1.south)!0.5!(group c2r1.south) + (0,-0.85cm)$)
    {\pgfplotslegendfromname{paretolegend}};
\end{tikzpicture}
\caption{Routing accuracy versus cost (log scale). Routing accuracy credits
abstention on impossible queries, so it equals answer accuracy for every
method except Switchcraft and the oracle, which reaches 100\% by construction.
(a)~In-distribution test set; Switchcraft's error bars span the minimum and
maximum over 20 seeds, and the dashed line traces its operating points as the
cost budget varies from 3 to $40 \times 10^{-4}$\,\$
(Appendix~\ref{app:adaptive-threshold}).
(b)~Held-out Glaive, never used for fine-tuning, calibration, or selection;
the dashed line applies the same per-budget thresholds.
Switchcraft is Pareto-optimal in distribution at every budget; on Glaive,
Switchcraft and kNN ($k{=}10$) both beat every single model at a fraction of
the strongest models' cost.}
\label{fig:pareto}
\end{figure*}

\textbf{Switchcraft is more accurate than every single model, at a fraction
of the cost.} Switchcraft reaches \textbf{86.40\%} routing accuracy at
\textbf{$27.5 \times 10^{-4}$\,\$} per query---\textbf{10.77\,pp} above the
best single model (GPT-5.3-chat, 75.63\%) at \textbf{59\%} lower cost,
saving \$3{,}875 per million queries. No baseline comes within 9.9\,pp;
every cheaper one trails by ${\geq}14.1$\,pp.

\textbf{A larger budget buys more accuracy.} Re-tuning only the two
thresholds on the selection half under other cost budgets traces a frontier
(dashed in Figure~\ref{fig:pareto}(a);
Appendix~\ref{app:adaptive-threshold}). At its most accurate point, which
ignores cost and sends every query it does not abstain on to its
most-likely-correct model, Switchcraft reaches \textbf{87.88\%} routing
accuracy (79.75\% answer, 89.90\% decisive)---\textbf{12.25\,pp} above
GPT-5.3-chat---yet still costs 34\% less ($43.6 \times 10^{-4}$\,\$), as over
half the queries it routes go to cheaper models.
Conversely, a $12 \times 10^{-4}$\,\$ budget keeps 84.88\% routing accuracy
(76.74\% answer, still above every single model) at
$19.8 \times 10^{-4}$\,\$, 70\% below GPT-5.3-chat's cost.

\textbf{Abstention avoids paying for impossible queries.} 14.3\% of test
queries defeat every model, so a router that always routes pays for a wrong
answer on each. Switchcraft instead returns such queries to the agent at zero
cost: it abstains on 9.0\% of queries,
catching 56.9\% of the impossible ones while abstaining on only 1.0\% of
answerable ones. Correct abstentions supply 8.14\,pp of its 10.77\,pp lead
over GPT-5.3-chat; the rest comes from answering more correctly.

\textbf{Decisive queries reveal routing skill.} Where
model choice matters and abstention earns no credit, Switchcraft leads
GPT-5.3-chat by \textbf{4.52\,pp} and kNN ($k{=}10$) by 3.05\,pp.

\textbf{Baselines.} No baseline abstains, so we also compare on answer
accuracy, which scores Switchcraft's abstentions as wrong and lower-bounds a
deployment that falls back to a default model. Switchcraft's
\textbf{78.27\%} still beats every single model (+2.64\,pp over GPT-5.3-chat)
and baseline: it is more accurate and cheaper than GPT-5.3-chat, GPT-5.4,
GPT-5.4-mini, the random and kNN ($k{=}10$) routers, and the turn- and
tool-count heuristics, and every cheaper baseline is ${\geq}5.9$\,pp less
accurate. Random routing spends about as much as Switchcraft but is 11.4\,pp
less accurate: what matters is where each query goes. kNN retrieval is the
strongest baseline: with $k{=}10$ it beats every single model (76.49\%) yet
trails Switchcraft by 1.78\,pp at 9\% higher cost, cannot abstain, and must
search an index that grows with training data, whereas Switchcraft needs one
encoder forward pass; with $k{=}1$ it is cheap but 7.38\,pp less accurate.
MIRT  (Appendix~\ref{app:mirt}), which trains a 39K-parameter head over frozen embeddings, reaches
70.68\% at $11.6 \times 10^{-4}$\,\$; kNN ($k{=}1$) is slightly more accurate
at 2.5$\times$ lower cost. The
best heuristic (number of turns) costs $3\times$ more than Switchcraft for
2.14\,pp less accuracy: no single surface feature captures the complexity of
agentic routing.

\textbf{Gap to the oracle.} The oracle reaches 100\% routing accuracy
(85.71\% answer accuracy) at $4.0 \times 10^{-4}$\,\$. From the
cheapest single model (GPT-5-nano, 67.75\%), Switchcraft closes 57.8\% of the
routing-accuracy gap and 58.6\% of the answer-accuracy gap. The remaining 13.6\,pp are routing errors
(Section~\ref{sec:error-analysis}). Per dataset, the advantage is largest on
FunReason-MT, +7.9\,pp in answer accuracy over the best single model
(Appendix~\ref{app:per-dataset}).

\textbf{Generalization to an unseen benchmark.} On all 83{,}814 held-out
Glaive queries, with the same weights, calibration, and policy
(Figure~\ref{fig:pareto}(b); Appendix~\ref{app:glaive-ood}),
all three encoders beat every single model (89.39--89.88\% routing accuracy
vs.\ 88.53\% for GPT-5.4; +3.7--6.1\,pp on decisive queries), and Switchcraft
costs 5.2$\times$ less than GPT-5.3-chat and 7.1$\times$ less than GPT-5.4,
sending 74\% of queries to GPT-5-nano. The same per-budget thresholds trace a
flatter frontier (dashed): every budget of at least $12 \times 10^{-4}$\,\$
beats every single model, and the budget-12 point matches the default at 38\%
lower cost (Table~\ref{tab:glaive-budget-sweep}). Glaive is easier and only partly
unseen: 74.1\% of its queries are trivial, and 24.5\% of its records repeat
Hermes training text, though on the rest Switchcraft still matches the best
single model at 4.7$\times$ lower cost. Two results do not transfer. kNN
($k{=}10$) matches our routers (89.43\%), erasing Switchcraft's lead over
retrieval; and the abstention head, calibrated in distribution, catches only
2.8\% of impossible Glaive queries (vs.\ 56.9\%). It fails safe (0.4\% false
abstentions) but should be recalibrated on a small labeled sample of a new
workload, without fine-tuning.

\subsection{Key Findings}
\label{sec:key-findings}

Tables~\ref{tab:main-results}--\ref{tab:chattiness-summary} and Glaive
(Appendix~\ref{app:glaive-ood}) reveal three findings.

\textbf{Finding 1: costlier is not always better.}
GPT-5.3-chat (75.63\%) outperforms the newer, more expensive GPT-5.4
(74.62\%) at 58\% of its cost: GPT-5.4 over-elaborates, producing verbose
chain-of-thought that occasionally corrupts the structured tool-call output.
GPT-5.4-mini costs 2.6$\times$ as much as GPT-5-mini for a 0.17\,pp gain, and
on Glaive, GPT-5-nano (86.98\%) outperforms both the 4.2$\times$ costlier
GPT-5-mini (83.94\%) and the newer GPT-5.4-nano (85.47\%), consistent with
our observation that nano follows tool-calling instructions more directly
while mini paraphrases arguments or adds explanations. Blindly routing to the
newest or most expensive model---a common enterprise default---can therefore
decrease both accuracy and cost-efficiency.

\textbf{Finding 2: model rankings shift across workloads.}
No single model is best everywhere: GPT-5-mini ranks fourth in distribution
but last on Glaive, Kimi-K2.5 last in distribution but fifth on Glaive, and
the best model per dataset group alternates among GPT-5.3-chat, GPT-5.4, and
Qwen-3.5-9B (Appendix~\ref{app:per-dataset}). Routing exploits exactly this
heterogeneity.

\textbf{Finding 3: nominally cheaper models can cost more.}
\label{sec:chattiness}
Per-token pricing is a misleading proxy for per-query cost. We define \textbf{chattiness} as the ratio of actual
per-query cost to an \emph{expected cost} assuming each model consumes the
cross-model average number of output tokens (above 1.0$\times$ = more
expensive than expected). Table~\ref{tab:chattiness-summary} summarizes
per-model results (derivation in Appendix~\ref{app:token-usage}).
Kimi-K2.5 is the chattiest (1.20$\times$; output is 49\% of its cost),
followed by GPT-5-nano (1.18$\times$), whose hidden reasoning tokens inflate
its cost. Qwen-3.5-9B generates the most tokens (299 avg), yet its chattiness
is only 1.06$\times$ because input cost dominates (86\%)---the large output
surplus barely moves the ratio. Qwen-3.5-9B also has the lowest list price in
our pool, yet costs 23\% more per query than GPT-5-nano (Table~\ref{tab:main-results}) because it consumes
more tokens. GPT-5.3-chat is the most concise (0.81$\times$, 71 avg tokens):
despite the second-highest output rate, it is both cheaper and
more accurate than GPT-5.4. A router selecting on list price would
misjudge verbose budget models and concise mid-tier ones;
using profiled per-query cost (Section~\ref{sec:routing-logic}) avoids this
trap.

\begin{table}[ht]
\centering
\caption{Per-model chattiness on the validation split (seed~6).}
\label{tab:chattiness-summary}
\small
\begin{tabular}{l r l}
\toprule
\textbf{Model} & \textbf{Chattiness} & \textbf{Pattern} \\
\midrule
Kimi-K2.5      & 1.20$\times$ & Verbose but budget \\
GPT-5-nano     & 1.18$\times$ & Hidden reasoning cost \\
Qwen-3.5-9B    & 1.06$\times$ & Verbose but cheap rate \\
GPT-5-mini     & 1.01$\times$ & Near average \\
GPT-5.4        & 0.90$\times$ & Concise \\
GPT-5.4-mini   & 0.90$\times$ & Concise \\
GPT-5.4-nano   & 0.87$\times$ & Concise \\
GPT-5.3-chat   & 0.81$\times$ & Most concise \\
\bottomrule
\end{tabular}
\end{table}

\subsection{Router Overhead and Build Cost}
\label{sec:latency}

\begin{table}[t]
\centering
\caption{Router inference on one NVIDIA T4 GPU.}
\label{tab:latency-summary}
\small
\setlength{\tabcolsep}{3pt}
\begin{tabular}{lrrrrr}
\toprule
& \multicolumn{3}{c}{\textbf{P99 latency (ms)}} & \textbf{Peak} & \textbf{Mem.} \\
\cmidrule(lr){2-4}
\textbf{Encoder} & \textbf{64 tok} & \textbf{200} & \textbf{512} & \textbf{(qps)}
  & \textbf{(MB)} \\
\midrule
DistilBERT (67M)  & 3.4  & 7.2  & 16.7 & 739 & 284 \\
ModernBERT (150M) & 19.4 & 22.2 & 53.0 & 243 & 871 \\
DeBERTa-v3 (184M) & 16.5 & 24.2 & 54.6 & 240 & 815 \\
\bottomrule
\end{tabular}
\end{table}

\textbf{Latency and serving cost.} On a commodity T4 GPU
(\textasciitilde\$0.35/hr spot), DistilBERT routes a query within 16.7\,ms
(P99) at up to 512 tokens, over an order of magnitude below typical LLM
generation latency (Table~\ref{tab:latency-summary}). ModernBERT and
DeBERTa-v3 are 3.1--5.8$\times$ slower, missing the 20\,ms budget at 200+
tokens, with no meaningful accuracy gain (Appendix~\ref{app:latency}).
Within the budget, one T4 makes 62--654 routing decisions per
second, so routing costs at most \$1.57 per million queries, under 0.06\% of
Switchcraft's \$2{,}750 LLM spend per million queries.
Calibration, abstention, and selection add only a few vector operations.

\textbf{Build cost.} Labeling our 99{,}125 in-distribution queries with all
eight models cost about \$2{,}300 at list prices, 70\% of it for GPT-5.4 and
GPT-5.3-chat, and fine-tuning DistilBERT takes 55 minutes on four A100 GPUs
(Appendix~\ref{app:training-config}). At \$3{,}875 saved per million queries
over GPT-5.3-chat, labeling pays for itself after about 600{,}000 routed
queries.

\subsection{Multi-Turn Execution and KV Caching}
\label{sec:multi-turn}

\textbf{Execution with recovery.} Our labels score each call given the
ground-truth history, but agents act on tool outputs. BFCL's multi-turn
categories simulate their environments, so we execute every call and score
the end-of-turn state, crediting recovery from failed calls, which 12--45\%
of each model's correct turns include (Appendix~\ref{app:multi-turn-scoring}).
Here, Switchcraft (72.7\%) beats kNN (68.0\%) and every model except GPT-5.4
(77.3\%), which costs 3.8$\times$ as much.

\textbf{KV caching.} Switching models between turns forfeits cached prompt
prefixes. We replay consecutive multi-turn
test turns of 20 seeds with a KV cache per model and published cache-read
prices, which discount cached input by 90\% for OpenAI models and 83\% for
Kimi-K2.5 (none assumed for Qwen-3.5-9B;
Appendix~\ref{app:kv-cache-routing}). Caching cuts per-turn routing's
cost by 65\%. Keeping the current model while it stays predicted-correct costs
7.3\% more at similar accuracy, and routing only the first turn saves 19\% but
loses 5.7\,pp of accuracy: neither improves on per-turn routing.

\subsection{Robustness and Ablations}
\label{sec:robustness}

\textbf{Seed stability and encoder choice.}
Across 20 random seeds, each with its own data split, DistilBERT's test answer
accuracy varies by only 1.66\,pp (77.33\%--78.99\%, mean $78.30 \pm 0.46$\%),
and it beats both GPT-5.3-chat and kNN ($k{=}10$) on every seed's partition,
by at least 1.43\,pp and 1.03\,pp, respectively. Its cost varies more, as
each seed tunes its thresholds, but averages $(29.0 \pm 7.2) \times
10^{-4}$\,\$ (seed~6: 27.5) and is below GPT-5.3-chat's
on every seed. ModernBERT and DeBERTa-v3
are equivalent (means 78.34\% and
78.10\%; Appendix~\ref{app:seed-stability}); we
prefer the faster DistilBERT.

\textbf{Decision-policy ablation.}
Table~\ref{tab:policy-ablation} varies only the decision policy on one
fine-tuned checkpoint. A fixed 0.5 threshold on raw sigmoid outputs (with
argmax fallback and no abstention) reaches only 73.33\%, below GPT-5.3-chat.
Calibrating the probabilities and tuning the reliability threshold under the
cost budget adds 4.59\,pp of answer accuracy (7.79\,pp on decisive queries)
at 2.2$\times$ the cost. Tuning abstention jointly, under the same budget,
adds 8.48\,pp of routing accuracy; because abstentions are free, it also lets
the policy afford a stricter reliability threshold (0.90 vs.\ 0.85), for a
further 0.35\,pp of answer accuracy.

\begin{table*}[t]
\centering
\small
\caption{Decision-policy ablation on one fine-tuned checkpoint (DistilBERT,
seed~6, test set).}
\label{tab:policy-ablation}
\begin{tabular}{lcccc}
\toprule
\textbf{Decision policy} & \textbf{Answer} & \textbf{Routing}
  & \textbf{Decisive} & \textbf{Cost} \\
 & \textbf{(\%)} & \textbf{(\%)} & \textbf{(\%)}
  & \textbf{($10^{-4}$\,\$)} \\
\midrule
Raw sigmoid, fixed threshold 0.5     & 73.33 & 73.33 & 78.97 & 11.6 \\
Calibrated, tuned $\tau_r$ (0.85)    & 77.92 & 77.92 & 86.76 & 25.6 \\
+ jointly tuned abstention (ours)    & 78.27 & 86.40 & 87.39 & 27.5 \\
\bottomrule
\end{tabular}
\end{table*}

\textbf{Token packing ablation.}
Our input representation (Section~\ref{sec:input-representation}) packs
decision-relevant content into 512 tokens. Fine-tuning the
same router without packing, on the conversation followed by all tool
signatures, right-truncated, lowers answer accuracy on 16 of the 20 seeds, by
0.50\,pp on average (0.85\,pp on decisive queries), and raises cost on 15
(mean 31.8 vs.\ $29.0 \times 10^{-4}$\,\$; Appendix~\ref{app:ablation-packing}).
Most inputs fit (median 86 tokens), limiting the gain;
packing matters most when earlier turns would push out the latest one.

\subsection{Error Analysis}
\label{sec:error-analysis}

We classify every test query into six mutually exclusive \textbf{routing
outcomes} (Table~\ref{tab:error-breakdown}). Switchcraft answers
\textbf{78.3\%} of queries correctly and correctly abstains on another 8.1\%.
Its 7.44\,pp answer-accuracy gap to the oracle is avoidable error:
\emph{wrong model} (6.6\%) and \emph{false abstention} (0.9\%). The remaining 6.2\% are
impossible queries that Switchcraft routes instead of abstaining on; they
waste money but cannot cost answer accuracy, since every model fails them.

\begin{table*}[t]
\centering
\caption{Routing outcomes on the test set (9{,}746 queries, DistilBERT,
seed~6).}
\label{tab:error-breakdown}
\small
\begin{tabular}{lrrl}
\toprule
\textbf{Outcome} & \textbf{Count} & \textbf{\%} & \textbf{Meaning} \\
\midrule
Correct--optimal     & 2{,}542 & 26.1 & Cheapest correct model selected \\
Correct--suboptimal  & 5{,}086 & 52.2 & A correct but costlier model selected \\
Correct abstention   &   793   &  8.1 & No model is correct; router abstains \\
Wrong model          &   640   &  6.6 & A correct model exists, but router picks one that fails \\
False abstention     &    85   &  0.9 & A correct model exists, but router abstains \\
Missed abstention    &   600   &  6.2 & No model is correct, but router routes anyway \\
\bottomrule
\end{tabular}
\end{table*}

Among correctly answered queries, 66.7\% go to a non-cheapest model. The
typical overhead is small (median $1.6 \times 10^{-4}$\,\$) but concentrated:
88 BFCL multi-turn queries (0.9\% of the test set) account for 74\% of it,
and multi-turn queries (1.5\%) make up 69\% of Switchcraft's spend.
Switchcraft sends 61\% of queries to the three cheapest models (GPT-5-nano,
GPT-5.4-nano, Qwen-3.5-9B) and only 15\% to GPT-5.4 or GPT-5.3-chat, relieving
the largest models, yet GPT-5.4's 6.1\% of queries account for 73\% of its
spend; the oracle would send 38\% to Qwen-3.5-9B, vs.\ 9\% for
Switchcraft. Ranking
models by \emph{average} profiled cost cannot see that a long multi-turn query
costs far more than average, which is why realized cost exceeds the
$17 \times 10^{-4}$\,\$ budget; per-query cost prediction is a natural next
step. Two robust difficulty patterns emerge
(Appendix~\ref{app:misrouted-details}): fewer correct models raise
error rates (62.7\% misrouted with one correct model vs.\ 0.1\% with eight), and
longer queries are harder (7.4\% below 15 tokens vs.\ 25.7\% at 108+).

\subsection{Chat Router and Other Model Pools}
\label{sec:chat}
With an earlier eight-model basket and protocol
(Appendix~\ref{app:old-model-basket}), Switchcraft likewise came within 0.5\,pp
of the best single model (82.73\% vs.\ 83.22\%) at 88\% lower cost.
Without retraining, it is also more accurate than two newer models, GPT-6-Sol
and GPT-6-Luna (by 6.4 and 4.5\,pp), and 67\% cheaper than GPT-6-Sol
(Appendix~\ref{app:newer-model-basket}). GPT-6-Luna costs 6.6$\times$
less, but as the cheapest correct model on 31.7\% of queries, it should
lower Switchcraft's cost if added to its basket. Switchcraft's value is thus
not tied to one model basket. An earlier chat-oriented DistilBERT
router for this pool, trained on 64 public, mostly chat-completion
benchmarks (Appendix~\ref{app:old-model-basket}), reached only 77.47\%
(5.26\,pp below Switchcraft), suggesting that routing learned mainly from chat
transfers poorly to agentic tool calling.

\section{Limitations}
\label{sec:discussion}

\textbf{Oracle gap.}
Switchcraft's 7.44\,pp answer-accuracy gap to the oracle, mostly wrong-model
choices, is the primary improvement opportunity, followed by its cost gap
($27.5$ vs.\ $4.0 \times 10^{-4}$\,\$), which is concentrated in a few
expensive multi-turn queries. Promising directions include asymmetric loss
shaping and per-query cost prediction.

\textbf{Distribution shift and contamination.}
Our in-distribution test set shares benchmarks with fine-tuning---a
\emph{known agentic workload mix} typical of enterprise settings with
representative production logs available for fine-tuning. On held-out Glaive
the cost advantage survives, but the accuracy margin narrows, abstention needs
recalibration, and Glaive partly overlaps Hermes (Appendix~\ref{app:glaive-ood}); more
distant shifts remain future work. All benchmarks except FunReason-MT predate
the 2025-08-31 GPT-5.3/5.4 training cutoff, so memorization may inflate
single-model accuracies. However, Switchcraft learns from
\emph{realized model behavior}, and its gain is largest on FunReason-MT,
released after that cutoff.

\textbf{Generalization to new models.}
Switchcraft must be retrained whenever its eight-model pool changes, though its design carries over to
an older pool and, without retraining, still compares favorably to two
newer models (Section~\ref{sec:chat}). MIRT represents models by text
descriptions and could in principle route to new ones without
retraining, but trails Switchcraft even on our pool
(Appendix~\ref{app:mirt}).

\section{Related Work}
\label{sec:related-work}

\noindent{\bf Tool calling} extends LLMs with access to external data and
computation, and prior work has proposed methods that teach
individual LLMs to invoke tools~\cite{toolformer,gorilla}. Scoring those calls,
however, remains the bottleneck. Prior work either executes each
call~\cite{xlam} or matches its abstract syntax tree against a reference using
BFCL's checker~\cite{bfcl,confetti}. The former is infeasible at our scale as it
requires implementing all the APIs, and the latter, as we show in
\S\ref{sec:ast-checker}, is significantly biased on every non-BFCL benchmark we
test. Prior work also focuses on synthetic data generation for tool
calls~\cite{RouteNator}.

\noindent{\bf Model routing} reduces inference cost by dispatching each query to
the cheapest model in a pool that can answer it with reasonable quality. A
growing body of work learns this routing policy from per-query quality signals,
using neural networks~\cite{hybrid_llm,fly-swat,routerdc,automix}, k-nearest
neighbors~\cite{llmrouting,routerbench,tensoropera}, matrix
factorization~\cite{embedllm,routellm}, graph neural
networks~\cite{graphrouter}, k-means clustering~\cite{uniroute}, item response
theory~\cite{irt_router}, or constrained optimization~\cite{omnirouter}.
Complementary lines target cost estimation directly~\cite{carrot,metallm},
enrich the action space with token-budget control~\cite{r2-router} or
best-of-$n$ sampling~\cite{best-route}, learn more expressive query
representations~\cite{icl_router}, or improve
explainability~\cite{explainableRouting}. Two recent benchmarks also target
model routing~\cite{routerarena,llmrouterbench}, and serving layers now
implement it~\cite{semantic_router}. Some routers predict per-model
correctness~\cite{llmrouting,irt_router}, and LLMRouterBench includes a tool-use
task, but they score final answers or completed tasks, not individual tool
calls. As we
argue in \S\ref{sec:motivation}, tool calls must be fully correct: a single
incorrect argument can have consequences. Routing has also been extended to multi-step tasks:
Router-R1~\cite{zhang2025routerr1teachingllmsmultiround} trains an LLM router
with reinforcement learning to answer questions by calling candidate models over
multiple rounds, and MTRouter~\cite{zhang2026mtroutercostawaremultiturnllm}
selects, at each turn of an agent episode, the model with the highest predicted
episode outcome. Both optimize task-level success rather than per-turn
correctness, and neither abstains.

We bridge these two threads with Switchcraft---to the best of our knowledge, the
first router fine-tuned and evaluated for strict tool-call correctness in
agentic workloads, supervised by the AST-level correctness of each call and
able to abstain when no candidate is likely to succeed.
A separate line of work attacks agent inference cost through orthogonal
mechanisms such as plan caching~\cite{plan_caching}, which compose with model
routing rather than replacing it.

\section{Conclusions}
\label{sec:conclusions}

We presented Switchcraft, an agentic model router using a DistilBERT
classifier (67M parameters) to route tool-calling queries among eight
candidates or abstain. In distribution, it exceeds the
best individual model's answer accuracy (78.3\% vs.\ 75.6\%; 86.4\% routing
accuracy with correct abstentions credited) at 59\% lower cost
(\$3{,}800+ saved per million queries) and outperforms kNN, MIRT, heuristic, and
random routers; on a held-out benchmark, it matches or exceeds every
individual model at 5--7$\times$ lower cost than the strongest ones. Key
enablers are agentic training data with an improved AST framework, calibrated
cost-aware routing with abstention, a chattiness-aware cost model, and
sub-20\,ms inference.
As available LLMs proliferate and their cost--capability
trade-offs diversify, intelligent routing becomes critical for scalable AI
deployment. Code generation and multi-step planning are natural future targets.

\section*{Acknowledgments}
\label{sec:acknowledgments}

We are grateful to Vijay Aski, Rupeshkumar Mehta, Sethu Raman, and Steve Sweetman
for supporting and enabling deep collaboration between Microsoft
Research and Microsoft Foundry on this effort.

\bibliographystyle{mlsys2025}
\bibliography{main}

\section*{AI Usage Disclosure}

\textbf{Ideation and literature search.} No AI tools were used for ideation: the
research question, system design, and evaluation methodology originated with the
authors. The authors identified the related work. GitHub Copilot CLI was used
only to check our descriptions of cited works against those papers and check for
bugs in the bibliography.

\textbf{Experiment design.} The authors designed and implemented the
experiments. GitHub Copilot CLI assisted in looking for bugs and diagnosing
unexpected output logs.

\textbf{Code generation and data analysis.} The authors designed and implemented
the code. GitHub Copilot CLI fixed bugs in the routing and evaluation pipeline,
which the authors reviewed before accepting.

\textbf{Data generation and annotation.} The candidate LLMs are the object of
study: their responses to benchmark queries, scored by our AST checker, form the
router's labels (Section~\ref{sec:datasets}). ConFETTI ground-truth corrections
used a human-in-the-loop process in which GPT-5 flagged suspected label errors
and proposed fixes that a human reviewer accepted, declined, or edited
(Appendix~\ref{app:confetti}). GPT-4o served as an LLM judge, one of the checks
used to validate our AST checker alongside unit tests and manual review
(Section~\ref{sec:ast-checker}). Our design-space exploration evaluated an LLM
router and LLM-as-judge labeling as alternatives and rejected both
(Appendix~\ref{app:design-space}). We used AI to generate or modify no other
training or evaluation data; some public benchmarks we use were synthesized by
their creators (Section~\ref{sec:datasets}).

\textbf{Writing and figures.} The authors drafted the text. GitHub Copilot CLI
edited it to fix formatting, spelling, and grammar issues and to adjust length. GitHub Copilot CLI also
audited the manuscript for inconsistencies and clarity. All those changes were
carefully reviewed by the authors.

\textbf{Human oversight.} The authors reviewed every AI-produced change before
accepting it, traced reported numbers to the logs and scripts that produce them,
and take full responsibility for the content of this paper.

\clearpage
\makeatletter
\immediate\write\@mainaux{\string\gdef\string\appendixstartpage{\the\c@page}}\makeatother
\appendix
\renewcommand{\dbltopfraction}{0.95}
\renewcommand{\dblfloatpagefraction}{0.8}
\renewcommand{\topfraction}{0.95}
\renewcommand{\floatpagefraction}{0.8}
\renewcommand{\textfraction}{0.05}
\setcounter{dbltopnumber}{3}
\setcounter{topnumber}{3}
\makeatletter
\setlength{\@fptop}{0pt}
\setlength{\@dblfptop}{0pt}
\setlength{\@fpsep}{\floatsep}
\setlength{\@dblfpsep}{\dblfloatsep}
\makeatother
\section{Acceptable Variations in Tool Calls}
\label{app:motivation-variation}

\begin{figure*}[t]
\centering
\small
\setlength{\fboxsep}{1pt}

\begin{tikzpicture}[
    node distance=0.2cm,
    box/.style={rectangle, rounded corners=2pt, text width=\linewidth-6.4pt, align=left, inner sep=3pt},
    turnbox/.style={box, draw=gray!60, fill=gray!5},
    callbox/.style={box, draw=blue!50, fill=blue!5},
    flexbox/.style={box, draw=green!60!black, fill=green!5},
    okannot/.style={text width=\linewidth-6.4pt, align=left, inner sep=3pt, text=green!50!black},
]

\node[anchor=west, text width=\linewidth-6.4pt, inner sep=3pt] (title) at (0, 0) {\textbf{Available tool:} calculate\_sales\_tax(purchase\_amount, city, state)};

\node[turnbox, below=0.08cm of title.south west, anchor=north west] (t1) {
\textbf{User:} ``Calculate the amount of sales tax to be added on a purchase amount of \$30.45 in Chicago, Illinois, \$52.33 in Sacramento, California and \$11.23 in Portland, Oregon.''
};

\node[callbox, below=0.08cm of t1.south west, anchor=north west] (c1) {
\textbf{Response A} {\color{green!50!black}\checkmark} (three parallel calls):\\
\begin{tabular}[t]{@{\hspace{0.3cm}}l@{~}l@{\ }l@{\ }l@{}}
1. & calculate\_sales\_tax(purchase\_amount=30.45, & city="Chicago", & state="Illinois") \\
2. & calculate\_sales\_tax(purchase\_amount=52.33, & city="Sacramento", & state="California") \\
3. & calculate\_sales\_tax(purchase\_amount=11.23, & city="Portland", & state="Oregon")
\end{tabular}
};

\node[flexbox, below=0.08cm of c1.south west, anchor=north west] (c2) {
\textbf{Response B} {\color{green!50!black}\checkmark} (also correct---reordered calls with abbreviated parameter values):\\
\begin{tabular}[t]{@{\hspace{0.3cm}}l@{~}l@{\ }l@{\ }l@{}}
1. & calculate\_sales\_tax(purchase\_amount=11.23, & city="Portland", & state=\colorbox{green!15}{"OR"}) \\
2. & calculate\_sales\_tax(purchase\_amount=30.45, & city=\colorbox{green!15}{"CHI"}, & state=\colorbox{green!15}{"IL"}) \\
3. & calculate\_sales\_tax(purchase\_amount=52.33, & city="Sacramento", & state=\colorbox{green!15}{"CA"})
\end{tabular}
};

\node[okannot, below=0.06cm of c2.south west, anchor=north west] (annot) {
{\bfseries Both responses are correct.}
Because these three calls are \emph{independent} (no data flows between them), any ordering is valid.
Additionally, semantically equivalent parameter values---``Illinois'' vs.\ ``IL'', ``Chicago'' vs.\ ``CHI''---are equally acceptable.
};

\end{tikzpicture}
\caption{Acceptable variation in an agentic parallel tool-calling scenario (BFCL v3~\cite{bfcl}). The user's request requires three independent \texttt{calculate\_sales\_tax} calls. Response~A and Response~B differ in (i)~the \emph{order} of the parallel calls and (ii)~the surface form of string parameters (e.g., ``Illinois'' vs.\ ``IL''), yet both are semantically correct. An agentic evaluation framework must recognize such benign variation rather than penalizing it as error.}
\label{fig:motivation-variation}
\end{figure*}

While agentic tool calls demand strict precision, not all variation constitutes
an error. Consider the sales-tax scenario in
Figure~\ref{fig:motivation-variation}, where the user asks an agent to compute
sales tax for purchases in three cities. The agent must issue three independent
calls to \texttt{calculate\_sales\_tax}, but because no call depends on the
output of another, any permutation of the three calls is equally valid.
Furthermore, string parameters admit semantically equivalent surface forms:
\texttt{"Illinois"} and \texttt{"IL"} refer to the same state, as do
\texttt{"Chicago"} and \texttt{"CHI"}. A correct evaluation must treat both
orderings and both spellings as acceptable, rather than penalizing responses
that happen to differ from a single canonical reference. These flexibility
requirements directly motivate the AST-checker design choices in
Section~\ref{sec:ast-checker}.

\section{Design Space Exploration}
\label{app:design-space}

The architecture described in Section~\ref{sec:design} emerged from systematic
exploration of a substantially larger design space.
Table~\ref{tab:design-space} summarizes the principal alternatives we evaluated
and the reasons we did not retain them in the final system.

\begin{table*}[t]
\centering
\caption{Design alternatives explored during development. Each row is an
approach we prototyped or carefully evaluated before discarding.}
\label{tab:design-space}
\small
\setlength{\tabcolsep}{4pt}
\begin{tabularx}{\linewidth}{p{3.5cm}X}
\toprule
\textbf{Alternative} & \textbf{Outcome and rationale for discarding} \\
\midrule
OpenAI embedding + MLP
  & Replaced the DistilBERT encoder with frozen OpenAI text embeddings fed to a
    small MLP (1--4M parameters). Accuracy was comparable on single-turn data
    but degraded on multi-turn benchmarks because the frozen embedding could not
    be fine-tuned to attend to conversation structure. Also requires an external
    API call at inference time, adding latency and a dependency. \\
\addlinespace
Larger encoders (RoBERTa-base/large, BERT-large)
  & Marginal accuracy gains ($<$0.5\,pp) over DistilBERT but doubled or tripled
    inference latency, violating our sub-20\,ms latency budget for a production
    router. \\
\addlinespace
LLM-as-a-router
  & Using a language model (e.g., GPT-4o-mini) to select the target model per
    query. Latency is 100--500$\times$ higher than an encoder classifier's,
    cost scales linearly with traffic, and accuracy was not meaningfully better.
    Dismissed early. \\
\addlinespace
RouteLLM-style similarity routing
  & Cosine-similarity-weighted scores using LMSYS preference data.
    Informative for error analysis but did not improve over the fine-tuned
    classifier: plain cosine similarity could not separate confusable cases in
    our function-calling domain. A kNN router over our own training queries
    is a stronger retrieval variant; we report it as a baseline
    (Section~\ref{sec:main-results}). \\
\addlinespace
Ensemble / heuristic hybrids
  & Combining rule-based heuristics (e.g., tool count, conversation length) with
    learned classifiers. The classifier alone subsumed the heuristic signals via
    the metadata preamble (Section~\ref{sec:input-representation}). \\
\addlinespace
Cost-weighted loss
  & Asymmetric loss that penalizes misrouting to expensive models more heavily.
    Improved cost slightly but reduced accuracy; the two-stage
    predict-then-select architecture already separates correctness from cost. \\
\addlinespace
BLEU / text-similarity scoring
  & Evaluated as an alternative to AST comparison for labeling correctness.
    Unsuitable: high BLEU scores can co-occur with functionally incorrect tool
    calls (e.g., wrong argument order), and low BLEU can accompany correct calls
    with cosmetic differences. \\
\addlinespace
LLM-as-judge labeling
  & Used a strong LLM to label correctness instead of AST comparison.
    Expensive, non-deterministic, and surprisingly unreliable on structured
    output: the judge often marked calls as correct when parameter values were
    subtly wrong. \\
\addlinespace
RL / contextual-bandit online routing
  & Learning a routing policy from live agent success signals. Promising in
    principle but requires a deployed agent generating reward signals;
    incompatible with our offline benchmark evaluation setting. Identified as
    future work. \\
\addlinespace
Distillation from oracle router
  & Training a fast student router to mimic oracle routing decisions.
    The oracle is already the implicit supervision signal in our multi-label
    formulation (correct/incorrect labels), so explicit distillation would provide no
    additional benefit. \\
\bottomrule
\end{tabularx}
\end{table*}

This exploration consumed the majority of our effort and informed
several key decisions: (i)~a fine-tuned encoder outperforms frozen embeddings;
(ii)~latency constraints eliminate larger backbones and LLM-based routers;
(iii)~AST-based scoring is essential for reliable labels in the function-calling
domain; and (iv)~separating correctness prediction from cost-aware selection
(two-stage routing) dominates single-objective alternatives such as
cost-weighted losses or $\alpha/\beta$ trade-off parameters.

\section{Profiled Cost Model}
\label{app:cost-model}

For every candidate model $m$, we average its actual dollar cost over all
$|\mathcal{D}| = 99{,}125$ raw examples of our in-distribution corpus (before
deduplication and splitting; held-out Glaive is excluded):
\begin{align*}
  C_{\text{profiled}}(m) &= \frac{1}{|\mathcal{D}|}\sum_{q \in \mathcal{D}}
    \text{cost}(q, m), \\
  \text{cost}(q, m) &= \frac{n_{\text{in}}(q, m) \cdot r_{\text{in}}(m)
    + n_{\text{out}}(q, m) \cdot r_{\text{out}}(m)}{10^6}.
\end{align*}
$n_{\text{in}}$ and $n_{\text{out}}$ are the actual input and output token
counts observed when model $m$ answered query $q$; $r_{\text{in}}$ and
$r_{\text{out}}$ are the per-million-token list prices from
Table~\ref{tab:main-results}. For multi-turn conversations, costs are summed
across all turns. This profiled cost captures \emph{chattiness}: a model that
generates extensive reasoning tokens or verbose explanations accumulates a
higher profiled cost than a concise model at the same per-token rate. The
router's cost-aware selection, the kNN and MIRT baselines, and the policy's cost budget
(Section~\ref{sec:eval-setup}) all use these per-model averages. Since they
are averages, they do not capture per-query variation: the realized cost of a
long multi-turn query can far exceed its model's average
(Section~\ref{sec:error-analysis}), which is why Switchcraft's realized test
cost ($27.5 \times 10^{-4}$\,\$) exceeds its $17 \times 10^{-4}$\,\$ profiled
budget.

\section{AST Checker Examples}
\label{app:ast-checker}

Tables~\ref{tab:ast_checker_examples}--\ref{tab:ast_checker_examples_nested}
present representative cases where the BFCL AST checker produces false
negatives, along with our corresponding fixes.

\begin{table*}[t]
	\centering
	\caption{Representative cases where the BFCL AST checker fails with our corresponding fix. Each row is a false-negative entry selected from parsing the GPT-5.3-chat responses. \emph{out} shows the model output, and \emph{GT} is the ground truth.}
	\label{tab:ast_checker_examples}
	\small
	\begin{tabularx}{\linewidth}{>{\raggedright\arraybackslash}p{0.12\linewidth} >{\raggedright\arraybackslash}p{0.27\linewidth} >{\raggedright\arraybackslash}X >{\raggedright\arraybackslash}p{0.22\linewidth}}
	\toprule
	\textbf{Limitation} & \textbf{Failing example} &
	\textbf{Why BFCL fails} & \textbf{Our resolution} \\
	\midrule
	
	Array order sensitivity &
	\emph{out:}\, \texttt{search\_books(} \newline
	\,\,\texttt{keywords=["haunted} \newline
	\,\,\texttt{house", "mystery"],} \newline
	\,\,\texttt{genre="mystery")} \newline
	\emph{GT:}\;\;\, \texttt{keywords=["mystery",} \newline
	\qquad\,\,\texttt{"haunted house"]} &
	BFCL compares the model's list to the ground-truth list with Python's
	\texttt{==} after exact-membership lookup, which is order-sensitive.
	The two lists contain the same two strings but in a different order, so
	equality is false and the call is marked wrong. &
	Order-insensitive multiset comparison with
	per-element string normalization. \\
	\midrule
	
	No default-parameter awareness &
	\emph{out:}\, \texttt{top\_players\_by\_} \newline
	\,\,\texttt{matchmaking(limit=50)} \newline
	\emph{GT:}\;\;\, \texttt{limit=50, page=0} \newline
	\emph{schema:}\, \texttt{page} described as \newline
	\,\,\emph{``Default is \texttt{0}''}  &
	BFCL treats every parameter that appears in the ground truth as
	required. Because the model omitted \texttt{page}, BFCL reports it
	missing; it never reads the schema description to learn that the
	documented default (\texttt{0}) is exactly the expected ground-truth
	value. &
	Default value parsed from the schema description;
	an omission is accepted iff the documented default matches the GT. \\
	\midrule
	
	Brittle string matching &
	\emph{out:}\, \texttt{birthdate=} \newline
	\,\,\texttt{"1990-05-15T00:00:00"} \newline
	\emph{GT:}\;\;\, \texttt{"1990-05-15T00:00:00Z"} \newline \newline
	\emph{out:}\, \texttt{date="15 June",} \newline
	\,\,\texttt{location="Office} \newline
	\,\,\texttt{conference room"} \newline
	\emph{GT:}\;\;\, \texttt{date="15th June",} \newline
	\,\,\texttt{location="office} \newline
	\,\,\texttt{conference room"} &
	BFCL does only case-insensitive byte equality on strings. The trailing
	\texttt{Z} (UTC marker) makes the two ISO-8601 timestamps unequal even
	though they denote the identical instant; likewise, \texttt{"15 June"}
	vs.\ \texttt{"15th June"} differs only by an ordinal suffix and
	\texttt{"Office conference room"} differs only in capitalization.
	BFCL has no whitespace, punctuation, ISO-8601, or paraphrase
	normalization. &
	Canonicalize case, whitespace,
	punctuation, and ISO-8601 (\texttt{Z}, \texttt{+00:00}); a DistilBERT
	cosine $\geq 0.85$ serves as a paraphrase fallback when GT contains
	spaces. \\
	\bottomrule
	\end{tabularx}
\end{table*}

\begin{table*}[t]
	\centering
	\caption{Further BFCL AST checker false negatives involving nested structures, with our corresponding fixes. Examples use the same GPT-5.3-chat responses and notation as Table~\ref{tab:ast_checker_examples}.}
	\label{tab:ast_checker_examples_nested}
	\small
	\begin{tabularx}{\linewidth}{>{\raggedright\arraybackslash}p{0.12\linewidth} >{\raggedright\arraybackslash}p{0.27\linewidth} >{\raggedright\arraybackslash}X >{\raggedright\arraybackslash}p{0.22\linewidth}}
	\toprule
	\textbf{Limitation} & \textbf{Failing example} &
	\textbf{Why BFCL fails} & \textbf{Our resolution} \\
	\midrule
	No nested-structure comparison &
	\emph{out:}\, \texttt{sync\_salesforce\_} \newline
	\,\,\texttt{data(authentication\_} \newline
	\,\,\texttt{details=\{} \newline
	\,\,\texttt{salesforce:\{client\_id,} \newline
	\,\,\texttt{...\}, pega:\{api\_key,} \newline
	\,\,\texttt{...\}\})} \newline
	\emph{GT:}\;\;\, byte-identical nested object \newline
	\emph{schema:}\, \texttt{authentication\_} \newline
	\,\,\texttt{details:\{type:"object",} \newline
	\,\,\texttt{properties:\{...\}\}} &
	BFCL maps each schema type string to a Python class through a
	hard-coded lookup table that has no entry for the JSON-Schema
	standard type \texttt{"object"} (and also omits \texttt{"number"}).
	The first time the checker meets such a parameter it raises
	\texttt{KeyError:\,'object'} and aborts the entire entry, marking it
	wrong regardless of content. Even if the lookup succeeded, BFCL has
	no recursive comparator and could not have validated the nested
	sub-objects. &
	Add \texttt{"object":dict} (and \texttt{"number":int}) to the type
	table; recursively validate each
	nested key against the corresponding sub-schema. \\
	\midrule

	Ambiguous array semantics &
	\emph{out:}\, \texttt{find\_common\_} \newline
	\,\,\texttt{elements(arrays=} \newline
	\,\,\texttt{[[1,2,3,4,5],} \newline
	\,\,\texttt{[2,3,4,6,7],} \newline
	\,\,\texttt{[2,3,8,9,10]])} \newline
	\emph{GT:}\;\;\, byte-identical 2-D matrix \newline
	\emph{schema:}\, \texttt{arrays:\{type:} \newline
	\,\,\texttt{"array", items:} \newline
	\,\,\texttt{\{type:"array"\}\}} &
	BFCL ground truth wraps every value in a list, which makes a true
	list-of-lists indistinguishable in shape from a list of alternative
	flat answers. BFCL has no schema-aware branch: it always interprets
	the outer list as alternatives. Here the parameter's actual type is
	\texttt{array of arrays}, so BFCL silently re-reads the GT as
	\emph{three alternative flat integer lists} and demands the model's
	output equal one of them. It can never match. &
	Consults \texttt{items.type}: alternatives only when the items are scalars;
	a true 2-D structure when items are themselves \texttt{array}. \\
	\midrule

	List-of-dict handling and ordering &
	\emph{out:}\, \texttt{calculate\_gpa(} \newline
	\,\,\texttt{grades=[} \newline
	\,\,\texttt{\{course:"Math",} \newline
	\,\,\texttt{grade:"B"\},} \newline
	\,\,\texttt{\{course:"Science",} \newline
	\,\,\texttt{grade:"A"\}, ...])} \newline
	\emph{GT:}\;\;\, same dicts reordered \newline
	\emph{schema:}\, \texttt{grades:\{type:} \newline
	\,\,\texttt{"array", items:} \newline
	\,\,\texttt{\{type:"object",...\}\}} &
	BFCL has no entry for \texttt{"object"} in its
	type table, so any \texttt{array of object} parameter is dispatched
	into a code path that immediately raises \texttt{KeyError:\,'object'}
	and aborts. Even if the crash were fixed, BFCL would still compare
	the two lists element-wise in order, so any reordering of the
	per-course dictionaries would also be rejected. &
	Each inner dict is normalized and the two lists are compared as multisets, so neither key order
	within a dict nor element order across the list affects the verdict. \\
	
	\bottomrule
	\end{tabularx}
\end{table*}

\section{Run-to-Completion Scoring on BFCL Multi-Turn}
\label{app:multi-turn-scoring}

Most of our datasets are scored one turn at a time by AST matching
(Section~\ref{sec:ast-checker}): given the ground-truth history, the model's
calls must match the reference calls. BFCL's multi-turn categories (base and
long context; 1{,}490 per-turn records, 150 of them in the test set) are scored
differently, because BFCL simulates their environments (file system,
messaging, trading, travel booking, and others). Within a turn, the model may
keep calling tools, seeing each call's execution result before the next, until
it stops or reaches BFCL's step limit (21 calls). We then execute all of its
calls and mark the turn correct if the final environment state equals the
state produced by the ground-truth calls and every ground-truth execution
result appears among the model's results. Intermediate calls are never
compared with the reference, so a model that makes a mistake and recovers
within the turn is credited. Such recovery is common:
Table~\ref{tab:multi-turn-recovery} shows that 12--45\% of each model's correct
turns include at least one failed call.

\begin{table*}[t]
\centering
\caption{Behavior on all 1{,}490 BFCL multi-turn per-turn records, over the
turns whose inference completed: median API calls per turn, share of turns the
model ended itself rather than at the step limit, and share of correct turns
that recovered from at least one failed call.}
\label{tab:multi-turn-recovery}
\small
\begin{tabular}{lccc}
\toprule
\textbf{Model} & \textbf{Calls per turn} & \textbf{Ended by model (\%)}
  & \textbf{Correct with recovery (\%)} \\
\midrule
GPT-5.3-chat & 21 & 0.0  & 26.5 \\
GPT-5.4      & 21 & 0.0  & 12.3 \\
GPT-5.4-mini & 21 & 0.0  & 25.2 \\
GPT-5-mini   & 21 & 0.0  & 29.4 \\
GPT-5.4-nano & 21 & 0.0  & 27.0 \\
GPT-5-nano   & 21 & 0.0  & 29.9 \\
Qwen-3.5-9B  & 21 & 0.0  & 45.0 \\
Kimi-K2.5    & 3  & 97.2 & 17.5 \\
\bottomrule
\end{tabular}
\end{table*}

Two limitations remain. First, recovery \emph{across} turns is not credited:
each per-turn record starts from the ground-truth state and history, so a
mistake can neither carry into nor be repaired in a later turn. BFCL's own
conversation-level metric, which requires every turn to pass, does not credit
it either. Second, our harness requires a tool call in every API request, which
should keep a model from ending a turn with a text reply. As
Table~\ref{tab:multi-turn-recovery} shows, every model except Kimi-K2.5 keeps
calling tools until the step limit (GPT-5.3-chat, for instance, repeatedly
calls \texttt{echo} to report its answer). Kimi-K2.5's API does not enforce the
requirement: in 97\% of its turns, Kimi-K2.5 eventually replies without a tool
call, almost always in natural language summarizing what it did, and the
harness treats such a reply as the end of the turn. The extra calls are
executed and scored, so one that changes the
state can break a turn the model had already solved; they are also billed, so
multi-turn costs (Table~\ref{tab:per-dataset-cost}) exceed what a natural
stopping point would give for every model except Kimi-K2.5. Records whose
inference fails count as incorrect. In 98 of the 745 long-context records
(13\%), every model fails before its first call, so these count as impossible
queries: rebuilding the long-context file system from the saved ground-truth
state re-adds BFCL's filler files, raising a duplicate-file error. Another
9--53 long-context records per model exceed the model's context limit;
Kimi-K2.5 never does, but loses 66 multi-turn records to API errors.

On this subset, Switchcraft (72.7\%) is more accurate than kNN (68.0\%) and
every single model except GPT-5.4 (77.3\%), which costs 3.8$\times$ as much
(Tables~\ref{tab:per-dataset} and~\ref{tab:per-dataset-cost}). No operating
point closes this gap: larger budgets (Table~\ref{tab:budget-sweep}) leave
Switchcraft at 72.7\% while raising its cost on this subset by 50\%, to
2.5$\times$ below GPT-5.4's, and smaller budgets are less accurate. With
its correct abstentions credited, its routing accuracy on this subset (78.0\%) is
within one query of GPT-5.4's.

\section{ConFETTI Ground-Truth Corrections}
\label{app:confetti}

The ConFETTI dataset~\cite{confetti} provides 506 multi-turn conversational
function-calling entries that we adopted for use in our evaluation
(Section~\ref{sec:evaluation}).
When manually examining results from preliminary experiments,
we observed that some ground-truth labels
contained errors---incorrect function calls or parameter values that do not
match the conversation context.
Because our routing evaluation scores every model against these labels, even
a small number of incorrect labels can unfairly penalize correct model
behavior and distort accuracy comparisons.
We therefore conducted a systematic review and corrected the ground truth
where necessary.

\subsection{Correction Methodology}

We performed two review rounds using an LLM-assisted human-in-the-loop
process.
In each round, every entry was submitted to a verifier model
(GPT-5) that received the full conversation history,
the available function definitions, and the current ground-truth label.
The model assessed whether the ground truth correctly represents the next
function call given the conversational context and flagged entries it
considered incorrect, along with a suggested correction and explanation.

Entries flagged by the model were then presented to a human reviewer in an
interactive terminal interface.
For each flagged entry the reviewer could \emph{accept} the correction
(updating the ground-truth file), \emph{decline} it (keeping the original
label), or \emph{skip} it so that the human could manually edit that entry, typically
because the suggested correction was not quite correct or the input needed correction rather
than the ground truth.
Entries the model deemed correct were automatically marked as reviewed
and not surfaced for human inspection.
A second round re-examined entries that were skipped and manually edited in round~1 and served
as a consistency check.

Across two rounds, a total of
\textbf{71 ground-truth corrections} (14.0\% of the dataset) were made.
Combined with the 86 input disambiguations described below, 138 of 506
entries (27\%) were modified in total.
We will open-source all corrections as a PR against the ConFETTI
repository, and separately open-source the LLM-assisted
correction tool used to produce them.

\subsection{Error Categories}

Table~\ref{tab:confetti-errs} classifies the 71 corrections into six
categories.  Each correction was assigned a single primary category; when
multiple issues co-occurred, we categorized by the most impactful error
(e.g., calling the wrong function takes precedence over a casing mismatch
in one of its parameters).
These categories matter for function-calling evaluation because scoring
typically relies on exact or near-exact matching of function names and
parameter values, so even minor mismatches in the ground truth can cause
correct model outputs to be marked as failures.

\begin{table*}[t]
\centering
\small
\caption{Categories of ground-truth errors corrected in the ConFETTI
dataset (71 corrections out of 506 entries).}
\label{tab:confetti-errs}
\begin{tabularx}{\linewidth}{p{3.0cm} r X X}
\toprule
\textbf{Category} & \textbf{Count} & \textbf{Description} & \textbf{Example} \\
\midrule
Wrong function
  & 14
  & Ground truth invokes a function that does not match the
    user's intent or the available conversational context.
  & User asks what their insurance \emph{covers};
    ground truth calls \texttt{get\_policy} (returns general
    policy info) instead of \texttt{get\_available\_coverage}.
\\[4pt]
Incorrect parameter value
  & 21
  & Correct function, but a structured parameter value is
    wrong---e.g., wrong entity ID, swapped coordinates, or
    incorrect reference from prior tool results.
  & User asks to check their \emph{savings} account balance;
    ground truth passes the \emph{checking} account number.
\\[4pt]
Incorrect content
  & 7
  & Free-text content in the function call does not match
    what the user actually said.
  & User says ``white elephant RSVP''; ground truth sets the
    email subject to ``Secret Santa RSVP.''
\\[4pt]
Format / schema error
  & 17
  & Value violates the function's schema constraints:
    wrong enum casing, incorrect data type, invalid
    date format, trailing whitespace, or
    invalid enum value.
  & Function expects date as \texttt{YYYY-MM-DD};
    ground truth passes \texttt{"2024-04-09T24:00:00Z"}
    (invalid date-time with hour 24).
\\[4pt]
Temporal error
  & 6
  & Date or time value is incorrect or unresolvable from the
    conversation---e.g., wrong UTC offset, unspecified year
    defaulting to an arbitrary past date.
  & User requests a 4:20\,PM Eastern appointment;
    ground truth books \texttt{09:20 UTC} instead of
    \texttt{20:20 UTC}.
\\[4pt]
Logical / state-tracking error
  & 6
  & Ground truth reflects flawed reasoning about the
    conversation state or skips a prerequisite step.
  & Assistant computed distances to JFK and EWR but not LGA;
    ground truth searches flights to LGA without first
    computing its distance.
\\
\bottomrule
\end{tabularx}
\end{table*}

\subsection{Input Disambiguation}

In addition to the 71 ground-truth corrections, we modified the
\emph{conversation input} of 86 entries (17\% of the dataset), all but one to
resolve temporal ambiguity.
Many ConFETTI conversations reference dates without specifying a year
(e.g., ``fly on the 25th of December'' or ``check hotel availability for
June'').
The ground-truth function calls in these entries use specific dates with
hardcoded years (e.g., \texttt{2023-12-25}), yet the conversation text
provides no basis for inferring which year was intended.
This creates an unfair evaluation setup: a model that produces the correct
function call with a different year---or that reasonably asks for
clarification---would be scored as incorrect.

To resolve this, we appended the year (or month and year where needed) to
the user utterance so that the conversation unambiguously specifies the date
referenced by the ground-truth label.
For example, ``fly on the 25th of December'' becomes ``fly on the 25th of
December 2023.''
The remaining entry received a non-temporal disambiguation: we added a
location qualifier (``in California'') because the ground truth assumed that
state.
These input changes do not alter the ground-truth labels; they ensure that
the information needed to produce the expected function call is present in
the conversation.

In total, 138 of 506 entries (27\%) received either an input
disambiguation, a ground-truth correction, or both (19 entries required
both).

\section{Dataset Details and Per-Dataset Breakdown}
\label{app:per-dataset}

This appendix provides detailed descriptions and cleaning procedures for
each dataset, followed by a per-dataset accuracy and cost breakdown and the
full results on held-out Glaive.

\subsection{Dataset Descriptions}

\textbf{BFCL~v3.}
The Berkeley Function Calling Leaderboard v3~\cite{bfcl} is the de facto
standard for evaluating tool-calling capabilities. We use all single-turn
categories (\texttt{simple}, \texttt{multiple}, \texttt{parallel},
\texttt{parallel\_multiple}) and their ``live'' counterparts, which test
against real-world API signatures. For the multi-turn categories
(\texttt{multi\_turn\_base}, \texttt{multi\_turn\_long\_context}), we
decompose each multi-turn conversation into per-turn evaluation records
using a custom script that replays ground-truth function executions to
build the conversation history progressively. This yields 745 per-turn
examples per category, providing fine-grained multi-turn evaluation.

\textbf{ConFETTI.}
ConFETTI~\cite{confetti} provides 109 human-simulated multi-turn
conversations spanning 86 unique APIs, totaling 506 turns after
decomposition. Unlike BFCL's synthetic multi-turn data, ConFETTI
conversations are authored by human annotators who simulate realistic
dialogue flows, including clarifications, corrections, and multi-step
task completions. We apply a format normalization step that restructures
conversation turns into the nested list format expected by our evaluation
framework. The ConFETTI dataset required substantial data cleaning: we
disambiguated 86 conversation inputs (17\%), 85 of them where temporal
references lacked a year, and corrected 71 ground-truth annotations (14.0\%) where the
expected function call was incorrect or incomplete---138 entries (27\%) in
total (Appendix~\ref{app:confetti}). We frame these corrections as
\emph{benchmark infrastructure}, not a methodological contribution: we will
open-source the corrected ConFETTI annotations, the LLM-assisted
correction tool used to produce them, and our improved AST checker, so
that any third party can reproduce, audit, or reject individual edits
without re-deriving the corrections.

\textbf{xLAM-60K.}
The Salesforce xLAM Function Calling 60K dataset~\cite{xlam} provides
60{,}000 single-turn function-calling examples. Each example consists of
a user query, a set of tool definitions, and ground-truth function calls.
This dataset contributes scale and diversity of tool schemas, serving as
the bulk of our single-turn training data.

\textbf{Hermes Function Calling v1.}
The Hermes Function Calling v1 dataset~\cite{hermes3} from Nous Research
contributes 8{,}940 examples drawn from three subsets:
\texttt{func\_calling\_singleturn}, \texttt{func\_calling}, and
\texttt{glaive\_func\_calling}. This provides a mix of single-turn and
multi-turn tool-calling scenarios, including both tool-call prediction
(selecting the right function and arguments) and tool-response
understanding (interpreting the result of a prior tool call). The
\texttt{glaive\_func\_calling} subset is derived from Glaive; we quantify
its overlap with our held-out Glaive set in Section~\ref{app:glaive-ood}.

\textbf{FunReason-MT.}
FunReason-MT~\cite{funreason_mt} is a data-synthesis framework for
multi-turn tool use; we use its public demonstration release,
HardGen~\cite{hardgen}, which contains multi-turn trajectories generated in
BFCL-style environments (file system, messaging, math, and similar APIs).
Since the release does not include the initial environment state needed by
BFCL's stateful multi-turn checker, we follow our Hermes and xLAM-60K
pipeline: we flatten each trajectory into per-turn tool-call prediction
records, skipping trajectories flagged as containing deliberately broken
tool responses. This yields 25{,}838 records from 13{,}530 conversations,
with a median of 32 tool definitions per query. FunReason-MT turns are
ordered, state-mutating call sequences (e.g., \texttt{cd} followed by
\texttt{cat}), so we score them with an order-sensitive variant of our AST
checker that matches the $i$-th predicted call against the $i$-th
ground-truth call, retaining the per-argument flexibility of
Section~\ref{sec:ast-checker}.

\textbf{Glaive Function Calling v2 (held out).}
The Glaive Function Calling v2 dataset~\cite{glaive_fc} is the largest
(83{,}814 examples) and covers multi-turn function-calling conversations
with diverse tool definitions. We convert the dataset to BFCL format and
apply two cleaning steps: (i)~replacing \texttt{None} placeholder values in
ground-truth arguments with appropriate defaults (e.g., empty objects for
functions like \texttt{get\_current\_time} that take no arguments), and
(ii)~removing empty-string optional parameters from ground-truth annotations
that would cause spurious evaluation failures. We hold Glaive out entirely:
it is evaluated with every method but never used for fine-tuning,
calibration, checkpoint or seed selection, or the kNN index.

\subsection{Per-Dataset Accuracy Breakdown}

Table~\ref{tab:per-dataset} breaks down answer accuracy by dataset group on
the in-distribution test set (seed-6 partition, 9{,}746 examples).

\begin{table*}[t]
\centering
\caption{Per-dataset-group answer accuracy (\%) on the in-distribution test
set. Dataset groups aggregate the sources in Table~\ref{tab:datasets}:
\emph{Single} = BFCL single-turn (simple, multiple, parallel,
parallel\_multiple),
\emph{Live} = BFCL live variants,
\emph{MT} = BFCL multi-turn (base + long context),
\emph{ConF.} = ConFETTI, \emph{FunR.} = FunReason-MT.
Sample sizes ($n$) shown below each column header.
Models sorted by overall accuracy; best single-model result per column
in \textbf{bold}.}
\label{tab:per-dataset}
\small
\setlength{\tabcolsep}{4pt}
\begin{tabular}{lrrrrrrrr}
\toprule
& \textbf{Single} & \textbf{Live} & \textbf{MT}
  & \textbf{ConF.} & \textbf{FunR.} & \textbf{Hermes}
  & \textbf{xLAM} & \textbf{All} \\
$n$ & 100 & 139 & 150 & 52 & 2559 & 783 & 5963 & 9746 \\
\midrule
GPT-5.3-chat  & \textbf{96.0} & \textbf{79.1} & 62.0 & 38.5 & 57.4 & 88.1 & \textbf{82.1} & \textbf{75.6} \\
GPT-5.4       & \textbf{96.0} & 77.7 & \textbf{77.3} & 32.7 & \textbf{59.9} & \textbf{88.8} & 78.9 & 74.6 \\
GPT-5.4-mini  & 93.0 & 76.3 & 60.7 & 34.6 & 55.9 & 88.6 & 77.6 & 72.5 \\
GPT-5-mini    & 90.0 & 74.1 & 64.7 & 34.6 & 59.7 & 85.7 & 76.1 & 72.3 \\
GPT-5.4-nano  & 90.0 & 76.3 & 65.3 & 34.6 & 50.6 & 86.5 & 76.1 & 70.0 \\
GPT-5-nano    & 87.0 & 76.3 & 58.0 & 32.7 & 44.2 & 87.7 & 75.3 & 67.8 \\
Qwen-3.5-9B   & 86.0 & 73.4 & 39.3 & \textbf{48.1} & 58.3 & 86.1 & 65.3 & 64.9 \\
Kimi-K2.5     & 89.0 & 63.3 & 68.0 & 38.5 & 11.8 & 79.7 & 39.0 & 36.4 \\
\midrule
MIRT          & 88.0 & 77.0 & 58.7 & 38.5 & 52.1 & 87.7 & 76.6 & 70.7 \\
kNN ($k{=}10$) & 90.0 & 79.9 & 68.0 & 42.3 & 62.6 & 90.3 & 80.9 & 76.5 \\
DistilBERT    & 94.0 & 80.6 & 72.7 & 40.4 & 67.8 & 90.2 & 81.4 & 78.3 \\
Oracle        & 100.0 & 94.2 & 88.7 & 67.3 & 76.1 & 95.7 & 88.2 & 85.7 \\
\bottomrule
\end{tabular}
\end{table*}

\textbf{Observations.}

\begin{itemize}
    \item \textbf{Routing helps most on the hardest benchmark.} On
          FunReason-MT, the router (67.8\%) beats the best single model
          (GPT-5.4, 59.9\%) by 7.9 percentage points and kNN by 5.2\,pp. It
          also beats every single model on BFCL live (80.6\% vs.\ 79.1\%)
          and Hermes (90.2\% vs.\ 88.8\%), confirming that routing can
          improve accuracy, not just reduce cost.

    \item \textbf{BFCL multi-turn (MT) remains hard.} The router (72.7\%)
          trails only GPT-5.4 (77.3\%), and the oracle (88.7\%) shows a
          16.0\,pp gap. This suggests that long multi-turn context is
          difficult for the router to capture within its 512-token window.

    \item \textbf{ConFETTI is challenging} due to its
          small size (52 queries) and conversational nature. The oracle itself
          reaches only 67.3\%, meaning a third of ConFETTI queries defeat all
          eight models. The router's 40.4\% is between the best single model
          (Qwen-3.5-9B, 48.1\%) and the median single model (34.6\%).

    \item \textbf{FunReason-MT and xLAM-60K show the largest model spreads}:
          Kimi-K2.5 (11.8\% and 39.0\%) vs.\ the best model (59.9\% and
          82.1\%), gaps of 48.1 and 43.1~pp. On xLAM-60K, the largest group,
          the router achieves 81.4\%, close to the best model.

    \item \textbf{BFCL-Single} queries are relatively easy (86--96\% across
          all models), so routing adds less value; the router's 94.0\% is
          slightly below the best single models (96.0\%).
\end{itemize}

Table~\ref{tab:per-dataset-cost} shows the corresponding per-query cost
breakdown. Costs are reported in milli-dollars ($10^{-3}$\,\$).

\begin{table*}[t]
\centering
\caption{Per-dataset-group average cost per query on the in-distribution
test set (9{,}746 examples), in milli-dollars ($10^{-3}$\,\$).
For most models, multi-turn queries are 130--320$\times$ more
expensive than single-turn ones because each of their up to 21 API calls
re-sends the conversation (Appendix~\ref{app:multi-turn-scoring}).
Cheapest single-model result per column in \textbf{bold}.
Costs here are computed as the per-query average of API-billed dollar
amounts (abstentions cost zero); they differ from the cost-decomposition
view in Table~\ref{tab:cost-decomp}, which is computed on a validation split
from average token counts and so does not preserve per-query variance.}
\label{tab:per-dataset-cost}
\small
\setlength{\tabcolsep}{4pt}
\begin{tabular}{lrrrrrrrr}
\toprule
& \textbf{Single} & \textbf{Live} & \textbf{MT}
  & \textbf{ConF.} & \textbf{FunR.} & \textbf{Hermes}
  & \textbf{xLAM} & \textbf{All} \\
\midrule
GPT-5.3-chat  & 1.29 & 1.89 & 271.1 & 4.54 & 6.91 & 1.06 & 1.37 & 6.62 \\
GPT-5.4       & 1.79 & 2.88 & 511.4 & 7.23 & 9.74 & 1.48 & 2.04 & 11.33 \\
GPT-5.4-mini  & 0.52 & 0.85 & 151.4 & 2.15 & 2.91 & 0.45 & 0.57 & 3.33 \\
GPT-5-mini    & 0.33 & 0.39 & 51.4 & 0.83 & 1.24 & 0.22 & 0.32 & 1.28 \\
GPT-5.4-nano  & 0.14 & 0.22 & 45.1 & 0.55 & 0.76 & 0.12 & 0.15 & 0.95 \\
GPT-5-nano    & 0.08 & 0.10 & \textbf{10.3} & \textbf{0.18} & \textbf{0.31} & 0.06 & 0.09 & \textbf{0.29} \\
Qwen-3.5-9B   & \textbf{0.06} & \textbf{0.09} & 15.0 & 0.19 & 0.39 & \textbf{0.05} & \textbf{0.07} & 0.35 \\
Kimi-K2.5     & 0.69 & 1.11 & 14.8 & 1.81 & 3.03 & 0.61 & 0.86 & 1.61 \\
\midrule
MIRT          & 0.15 & 0.43 & 53.4 & 3.28 & 0.81 & 0.12 & 0.26 & 1.16 \\
kNN ($k{=}10$) & 0.38 & 0.61 & 149.7 & 1.93 & 2.27 & 0.22 & 0.38 & 3.00 \\
DistilBERT    & 0.58 & 0.85 & 134.3 & 3.35 & 2.10 & 0.36 & 0.37 & 2.75 \\
Oracle        & 0.10 & 0.17 & 14.0 & 0.59 & 0.45 & 0.06 & 0.09 & 0.40 \\
\bottomrule
\end{tabular}
\end{table*}

\textbf{Cost observations.}

\begin{itemize}
    \item \textbf{Multi-turn queries dominate cost.} MT queries cost
          130--320$\times$ more than single-turn queries for the same model
          (e.g., GPT-5.4: \$0.511 vs.\ \$0.002 per query) because every
          model except Kimi-K2.5 runs each multi-turn record to the 21-call
          step limit, re-sending the conversation each time; Kimi-K2.5
          (21$\times$) is the exception because it ends its turns after a
          few calls. Although MT
          comprises only 1.5\% of test queries, it accounts for 69\% of
          Switchcraft's total spending.

    \item \textbf{Switchcraft saves 50--73\% on most dataset groups} compared
          to the best single model (GPT-5.3-chat), with savings ranging from
          50\% (MT) to 73\% (xLAM-60K). The exception is ConFETTI,
          where savings are only 26\%---the router routes many ConFETTI
          queries to expensive models, reflecting the difficulty of these
          conversational queries.

    \item \textbf{GPT-5-nano is the cheapest model overall}, while
          Qwen-3.5-9B is cheapest on the four mostly single-turn groups but
          not on the multi-turn ones, where its tokenizer inflates long input
          contexts.
          Switchcraft costs $9.6\times$ as much as GPT-5-nano but is
          10.5\,pp more accurate, and costs 59\% less than GPT-5.3-chat
          while being 2.6\,pp more accurate.
\end{itemize}

\subsection{Held-Out Glaive Results}
\label{app:glaive-ood}

Table~\ref{tab:glaive-full} reports every method on held-out Glaive. Glaive
is dominated by trivial queries (74.1\%; 19.1\% decisive, 6.8\% impossible),
which compresses answer accuracy into an 83.9--89.7\% band for every
deployable method while decisive accuracy spans 51.6--81.8\%. The
turn-count heuristic sends every Glaive query to GPT-5.3-chat or GPT-5.4 and
costs $8.59 \times 10^{-4}$\,\$, while the tool-count heuristic sends every
query to GPT-5-nano and matches it exactly.

\begin{table*}[t]
\centering
\caption{All methods on held-out Glaive (83{,}814 queries), using the same
router weights, calibration, and policy as Table~\ref{tab:main-results}. kNN
and MIRT use the seed-6 training split (MIRT's validation-selected seed);
DeBERTa-v3 uses its validation-selected seed (247).}
\label{tab:glaive-full}
\small
\begin{tabular}{llcccc}
\toprule
\textbf{Type} & \textbf{Entity} & \textbf{Answer} & \textbf{Routing}
  & \textbf{Decisive} & \textbf{Cost} \\
 & & \textbf{(\%)} & \textbf{(\%)} & \textbf{(\%)}
  & \textbf{($10^{-4}$\,\$)} \\
\midrule
\multirow{8}{*}{Single LLM}
  & GPT-5.4        & 88.53 & 88.53 & 75.64 & 10.27 \\
  & GPT-5.3-chat   & 88.46 & 88.46 & 75.29 & 7.58 \\
  & GPT-5.4-mini   & 88.04 & 88.04 & 73.06 & 3.06 \\
  & GPT-5-nano     & 86.98 & 86.98 & 67.53 & 0.42 \\
  & Kimi-K2.5      & 86.89 & 86.89 & 67.04 & 3.95 \\
  & Qwen-3.5-9B    & 86.26 & 86.26 & 63.73 & 0.39 \\
  & GPT-5.4-nano   & 85.47 & 85.47 & 59.59 & 0.82 \\
  & GPT-5-mini     & 83.94 & 83.94 & 51.55 & 1.77 \\
\midrule
\multirow{4}{*}{Baseline}
  & kNN ($k{=}10$) & 89.43 & 89.43 & 80.36 & 1.27 \\
  & kNN ($k{=}1$)  & 87.24 & 87.24 & 68.90 & 0.46 \\
  & MIRT           & 87.05 & 87.05 & 67.87 & 0.48 \\
  & Random         & 86.83 & 86.83 & 66.74 & 3.52 \\
\midrule
\multirow{3}{*}{Heuristic}
  & Num.\ Turns    & 88.20 & 88.20 & 73.93 & 8.59 \\
  & Num.\ Tools    & 86.98 & 86.98 & 67.53 & 0.42 \\
  & Length         & 86.87 & 86.87 & 66.92 & 0.48 \\
\midrule
\multirow{3}{*}{\textbf{Ours}}
  & ModernBERT (150M) & 89.40 & 89.51 & 80.23 & 1.29 \\
  & DistilBERT (67M)  & 89.20 & 89.39 & 79.32 & 1.46 \\
  & DeBERTa-v3 (184M) & 89.68 & 89.88 & 81.79 & 1.80 \\
\midrule
Upper Bound
  & Oracle         & 93.17 & 100.00 & 100.00 & 0.37 \\
\bottomrule
\end{tabular}
\end{table*}

\textbf{Abstention under shift.} The routers' abstention heads,
calibrated in distribution, rarely fire on Glaive: DistilBERT, ModernBERT,
and DeBERTa-v3 abstain on 0.60\%, 0.17\%, and 0.34\% of queries and catch
2.8\%, 1.5\%, and 2.9\% of the 5{,}723 impossible ones, with false-abstention
rates of at most 0.43\%. Routing accuracy therefore barely exceeds answer
accuracy; recalibrating the abstention threshold on a small labeled sample of
the new workload would restore it without fine-tuning.

\textbf{Overlap with training.} Hermes's \texttt{glaive\_func\_calling}
subset is derived from Glaive. Comparing conversation texts (ignoring tool
definitions) against the seed-6 Hermes training split, we find that 24.5\% of Glaive
records (20{,}572) repeat the text of some Hermes training record; Glaive is
highly templated (its 83{,}814 records contain only 28{,}817 distinct
conversation texts), and the tool sets differ (only 1.3\% of Glaive records
share both the opening request and the tool names with a training record).
Table~\ref{tab:glaive-overlap} splits the results accordingly. On the
overlapping records, every router leads the strongest models, by 6--8\,pp on
decisive queries. On the 75.5\% without overlap, Switchcraft (DistilBERT)
matches the best single model there, GPT-5.3-chat (87.45\% vs.\ 87.37\%
answer; 77.14\% vs.\ 76.51\% decisive), at 4.7$\times$ lower cost, while
DeBERTa-v3 and kNN ($k{=}10$) exceed GPT-5.3-chat by 0.72 and 0.46\,pp. The
cost advantage is robust to the overlap; the accuracy margin is not.

\begin{table*}[t]
\centering
\caption{Held-out Glaive split by textual overlap with Hermes training
records: answer and decisive accuracy (\%) and cost ($10^{-4}$\,\$).
GPT-5.3-chat is the best single model on the non-overlapping records.}
\label{tab:glaive-overlap}
\small
\setlength{\tabcolsep}{4pt}
\begin{tabular}{l ccc ccc}
\toprule
& \multicolumn{3}{c}{\textbf{Overlap (20{,}572)}}
  & \multicolumn{3}{c}{\textbf{No overlap (63{,}242)}} \\
\cmidrule(lr){2-4} \cmidrule(lr){5-7}
\textbf{Method} & \textbf{Answer} & \textbf{Decisive} & \textbf{Cost}
  & \textbf{Answer} & \textbf{Decisive} & \textbf{Cost} \\
\midrule
GPT-5.4         & 93.19 & 78.77 & 8.59 & 87.01 & 74.66 & 10.82 \\
GPT-5.3-chat    & 91.82 & 71.40 & 6.77 & 87.37 & 76.51 & 7.85 \\
kNN ($k{=}10$)  & 94.34 & 84.90 & 0.99 & 87.83 & 78.92 & 1.36 \\
ModernBERT      & 94.69 & 86.79 & 0.96 & 87.69 & 78.17 & 1.40 \\
DistilBERT      & 94.57 & 86.24 & 0.82 & 87.45 & 77.14 & 1.66 \\
DeBERTa-v3      & 94.60 & 86.37 & 1.15 & 88.08 & 80.35 & 2.01 \\
Oracle          & 97.15 & 100.00 & 0.32 & 91.88 & 100.00 & 0.38 \\
\bottomrule
\end{tabular}
\end{table*}

\textbf{Operating points under shift.}
Table~\ref{tab:glaive-budget-sweep} applies the per-budget thresholds of
Table~\ref{tab:budget-sweep}, tuned in distribution, unchanged to Glaive. The
trade-off carries over but flattens: at every budget of at least
$12 \times 10^{-4}$\,\$, Switchcraft beats every single model, yet raising it
from 12 to 40 adds only 0.20\,pp of routing accuracy for 4.6$\times$ the cost,
and the budget-12 point matches the default at 38\% lower cost. Realized costs
stay far below the budgets, as every model is cheaper per
Glaive query (GPT-5.4 costs $10.27$ vs.\ $113.3 \times 10^{-4}$\,\$ in
distribution): the thresholds transfer, not a spending level. Budgets of
$8 \times 10^{-4}$\,\$ or less trail the strongest single models, though
the budget-3 point roughly matches GPT-5-nano at its cost, and kNN
($k{=}10$; 89.43\% at $1.27 \times 10^{-4}$\,\$) sits just above the frontier.

\begin{table*}[t]
\centering
\caption{Operating points of the tuned policy on held-out Glaive (DistilBERT,
seed~6), using the in-distribution thresholds of Table~\ref{tab:budget-sweep}.
Accuracies (\%) and realized cost ($10^{-4}$\,\$) are measured on all
83{,}814 Glaive queries.}
\label{tab:glaive-budget-sweep}
\small
\begin{tabular}{rcccccr}
\toprule
\textbf{Budget} & \textbf{$\boldsymbol{\tau_r}$} & \textbf{$\boldsymbol{\tau_a}$}
  & \textbf{Answer} & \textbf{Routing} & \textbf{Decisive} & \textbf{Cost} \\
\midrule
3                 & 0.35 & 0.60 & 86.76 & 87.13 & 66.64 & 0.42 \\
5                 & 0.55 & 0.70 & 87.25 & 87.49 & 69.09 & 0.48 \\
8                 & 0.70 & 0.75 & 88.05 & 88.24 & 73.28 & 0.65 \\
12                & 0.80 & 0.75 & 89.19 & 89.38 & 79.27 & 0.90 \\
\textbf{17} (default) & 0.90 & 0.75 & 89.20 & 89.39 & 79.32 & 1.46 \\
25                & 0.95 & 0.75 & 89.30 & 89.49 & 79.82 & 1.97 \\
40                & 1.00 & 0.75 & 89.39 & 89.58 & 80.33 & 4.16 \\
\bottomrule
\end{tabular}
\end{table*}

\section{Router Fine-Tuning Configuration}
\label{app:training-config}

This appendix details the hyperparameters and fine-tuning settings used for all
three encoder models evaluated in Section~\ref{sec:robustness}.
All models share the same fine-tuning framework (Hugging Face
Transformers), loss function, and data pipeline;
they differ only in the base model, sequence length, batch size, learning
rate, and numeric precision.

\subsection{Training Data}

Training labels are generated from the evaluation results of all 8 target
models on the 14 in-distribution dataset splits (Section~\ref{sec:evaluation}).
Held-out Glaive contributes no training, validation, or test examples.
For each query, a model is labeled as \emph{correct} if its AST-match score
$\geq 0.75$ (the same threshold used by BFCL v3 scoring).
This produces a \textbf{multi-label} binary target vector per query---multiple
models may be correct for the same input.

\textbf{Stratified 80/10/10 splits.}
Each in-distribution dataset split (the 14 non-Glaive rows of
Table~\ref{tab:datasets}) is partitioned
\emph{independently} into 80\% train / 10\% validation / 10\% test, and then the
per-dataset slices are concatenated and shuffled to form the global splits.
Stratifying by dataset guarantees that every in-distribution benchmark and
every category within BFCL (Simple, Multiple, Parallel, Live*, Multi-turn,
etc.) appears in the train, validation, and test sets in its native 80/10/10
ratio; only Glaive is held out entirely, by design.
Deduplication on the (\textit{query, tools}) pair is performed
\emph{within each dataset} before splitting, preventing near-duplicates
from leaking across the train/validation/test boundary. The final sizes
are 77{,}891 training, 9{,}732 validation, and 9{,}746 test examples.

\textbf{Seed-dependent splits.}
The seed value controls \emph{both} the random initialization of the
classifier head \emph{and} the per-dataset shuffle that produces the
80/10/10 split. Concretely, each seed $s$ deterministically produces a
distinct train/validation/test partition (\texttt{train\_data\_*\_$s$.csv},
\texttt{val\_data\_*\_$s$.csv}, \texttt{test\_data\_*\_$s$.csv}) by seeding
\texttt{random.seed(s)} before shuffling each dataset. The 80/10/10 ratio
and per-dataset stratification are preserved across seeds---only the
specific examples assigned to each split change. As a result, the
$\pm0.46$\,pp seed variance in test answer accuracy reported in
Section~\ref{sec:robustness} reflects variation across both classifier
initializations \emph{and} train/validation/test partitions, providing a
stronger generalization signal than a fixed-split protocol with only
initialization-level seed variation.

To assess seed stability, each model is fine-tuned independently with 20 random
seeds (see Appendix~\ref{app:seed-stability}). The full list of seeds is
given at the end of this appendix.

\textbf{Calibration and selection halves.}
Each seed's validation split is further divided into a \emph{calibration}
half and a \emph{selection} half (Section~\ref{sec:routing-logic}),
stratified by dataset and by query difficulty (trivial, decisive, or
impossible) and assigned deterministically by hashing query IDs with a fixed
split seed (2025). The calibration half fits the per-model Platt scalers and
the abstention head; the selection half scores fine-tuning checkpoints and
tunes the routing thresholds. The test split is used only for reporting, and
we select the reported seed by validation answer accuracy.

\textbf{Earlier evaluation protocol.}
Two appendix experiments---the earlier model basket
(Appendix~\ref{app:old-model-basket}) and probabilistic labels
(Appendix~\ref{app:probabilistic})---were run with our earlier protocol. It used five
benchmarks---BFCL~v3, ConFETTI, Glaive, xLAM-60K, and Hermes---in 14 splits,
each split 80/10/10, so Glaive was in distribution and FunReason-MT was not
yet included (122{,}267 examples after deduplication; 12{,}267 validation and
12{,}282 test). It selected checkpoints on validation macro-F1, applied a
fixed 0.5 threshold to raw sigmoid outputs with argmax fallback, never
abstained, and used an oracle that routed impossible queries to the most
expensive model. Comparisons within these experiments are internally
consistent, but their absolute numbers are not comparable to
Tables~\ref{tab:main-results} and~\ref{tab:glaive-full}.

\subsection{Input Representation}

Each training example is tokenized using the token packing strategy described
in Section~\ref{sec:input-representation}.
Table~\ref{tab:packing-budget} shows the per-model token budgets.

\begin{table*}[t]
\centering
\small
\caption{Token packing budgets by model.}
\label{tab:packing-budget}
\begin{tabular}{lccc}
\toprule
\textbf{Setting} & \textbf{DistilBERT} & \textbf{DeBERTa-v3} & \textbf{ModernBERT} \\
\midrule
Max sequence length      & 512   & 512   & 8{,}192 \\
Max tool tokens          & 100   & 100   & 1{,}600 \\
\bottomrule
\end{tabular}
\end{table*}

Within the budget, the packing algorithm prioritizes the latest turn, compact
tool signatures, and earlier turns newest-first, as detailed in
Section~\ref{sec:input-representation}.

\subsection{Shared Hyperparameters}

Table~\ref{tab:shared-hparams} lists settings common to all three models.

\begin{table*}[t]
\centering
\small
\caption{Shared training hyperparameters.}
\label{tab:shared-hparams}
\begin{tabular}{ll}
\toprule
\textbf{Hyperparameter} & \textbf{Value} \\
\midrule
Optimizer                  & AdamW \\
LR scheduler               & Linear decay with warmup \\
Warmup ratio               & 0.1 \\
Weight decay               & 0.01 \\
Max epochs                 & 30 \\
Early stopping patience    & 3 epochs (on checkpoint metric) \\
Loss function              & BCEWithLogitsLoss \\
Class balancing (pos\_weight) & Disabled \\
Checkpoint metric          & Decisive accuracy at budget (selection half) \\
Probability calibration    & Per-model Platt scaling (calibration half) \\
Abstention head            & Logistic regression on calibrated probabilities \\
Routing thresholds         & $(\tau_r, \tau_a)$ on a 0.05-step grid (selection half) \\
Cost budget                & $17 \times 10^{-4}$\,\$ per query (profiled) \\
Impossible-detection floor & 55\% \\
Mixed precision            & FP16 (DeBERTa-v3: BF16 with \texttt{torch.compile}) \\
\bottomrule
\end{tabular}
\end{table*}

\subsection{Per-Model Hyperparameters}

Table~\ref{tab:per-model-hparams} reports each model's configuration and
parameter counts.
The learning rate and batch size were selected via preliminary experiments;
DistilBERT tolerates a higher learning rate due to its smaller depth, while
DeBERTa-v3 and ModernBERT fine-tune stably at $2 \times 10^{-5}$.
ModernBERT uses a smaller batch size to accommodate its longer sequences
within GPU memory.

\begin{table*}[t]
\centering
\small
\caption{Per-model training hyperparameters and parameter counts.}
\label{tab:per-model-hparams}
\begin{tabular}{lccc}
\toprule
\textbf{Setting} & \textbf{DistilBERT} & \textbf{DeBERTa-v3} & \textbf{ModernBERT} \\
\midrule
Base model            & \texttt{distilbert-base-uncased} & \texttt{deberta-v3-base} & \texttt{ModernBERT-base} \\
Total parameters      & 67.0M & 184.4M & 149.6M \\
Token embedding params & 23.44M & 98.38M & 38.68M \\
Classification head params & 0.597M & 0.597M & 0.597M \\
Hidden size           & 768   & 768   & 768 \\
Layers                & 6     & 12    & 22 \\
Attention heads       & 12    & 12    & 12 \\
Learning rate         & $5 \times 10^{-5}$ & $2 \times 10^{-5}$ & $2 \times 10^{-5}$ \\
Train batch size      & 16    & 32    & 8 \\
Eval batch size       & 16    & 64    & 8 \\
\bottomrule
\end{tabular}
\end{table*}

Parameter counts are for the full fine-tuned sequence-classification model.
The token embedding row counts only the vocabulary embedding matrix
(vocabulary size $\times$ 768), excluding positional and relative-position
embeddings. Both the token embedding and classification head counts are
included in the total, not added to it.

\subsection{Classification Head}

We fine-tune the encoder and its Hugging Face sequence-classification head
end-to-end. Each head maps a 768-dimensional representation to eight logits,
one per candidate LLM. The final linear layer has $768 \times 8 + 8 = 6{,}152$
parameters; model-specific projections (a pre-classifier for DistilBERT, a
pooler for DeBERTa-v3, and a dense layer plus normalization for ModernBERT)
bring each full head to 596{,}744 parameters. DistilBERT's head also applies
dropout, which adds no parameters.
The loss is binary cross-entropy with logits, computed independently for each
class, enabling multi-label prediction.

\subsection{Checkpoint Selection and Inference-Time Policy}

During fine-tuning, each epoch's checkpoint is scored on the selection half
by the decisive accuracy of the best budget-feasible routing policy, computed
from the raw sigmoid outputs, with the probability that every model fails,
$\textstyle\prod_m (1-p_m)$, standing in for the not-yet-fitted abstention head; the best checkpoint is restored at
the end.

At inference time, the model outputs 8 logits (one per target model).
Per-model Platt scalers convert them to calibrated probabilities $p_m$, and
the abstention head---a logistic regression over the eight calibrated
probabilities, fit on the calibration half to predict whether no model is
correct---outputs $p_\varnothing$. The router abstains if
$p_\varnothing \geq \tau_a$; otherwise it selects the model with the lowest
profiled cost among those with $p_m \geq \tau_r$, falling back to the
highest-probability model (\texttt{argmax}) if none qualifies. On the
selection half, we evaluate every $(\tau_r, \tau_a)$ pair on a 0.05-step grid
and keep the one with the highest decisive accuracy among pairs whose average
profiled cost is at most $17 \times 10^{-4}$\,\$ and that abstain on at least
55\% of impossible queries, breaking ties by routing accuracy, answer
accuracy, and lower cost. For the reported seeds, this yields
$\tau_r = 0.90$ for all three encoders, with $\tau_a = 0.75$ (DistilBERT,
DeBERTa-v3) or $0.70$ (ModernBERT).

\subsection{Compute Environment}

All fine-tuning was conducted on a node with 4$\times$ NVIDIA A100 80\,GB PCIe GPUs.
The parallelism strategy differs by model:
DistilBERT uses PyTorch DataParallel across all 4 GPUs per seed
(effective batch size $16 \times 4 = 64$; seeds run sequentially);
DeBERTa-v3 uses DistributedDataParallel (DDP) via \texttt{torchrun} across
all 4 GPUs per seed (effective batch size $32 \times 4 = 128$; seeds run
sequentially);
ModernBERT does not support multi-GPU parallelism, so seeds are distributed
across GPUs (5 seeds per GPU, run sequentially in parallel across devices;
batch size 8 on a single GPU).
Table~\ref{tab:compute-env} lists the software stack.

\begin{table}[ht]
\centering
\small
\caption{Software environment for router fine-tuning.}
\label{tab:compute-env}
\begin{tabular}{ll}
\toprule
\textbf{Component} & \textbf{Version} \\
\midrule
Python          & 3.11 \\
PyTorch         & 2.6.0 (CUDA 12.4) \\
Transformers    & 4.55.2 \\
Accelerate      & 1.10.0 \\
scikit-learn    & 1.7.1 (calibration, abstention head) \\
\bottomrule
\end{tabular}
\end{table}

\textbf{Wall-clock time.}
DistilBERT fine-tunes in ${\sim}$55\,min per seed on 4 GPUs via DataParallel
(18\,hr total for 20 seeds, run sequentially).
DeBERTa-v3 requires ${\sim}$6.7\,hr per seed using 4-GPU DDP (134\,hr total).
ModernBERT requires ${\sim}$5.3\,hr per seed on a single GPU (27\,hr wall-clock with
5 seeds per GPU $\times$ 4 GPUs in parallel).
Total compute across all 60 runs (3 models $\times$ 20 seeds):
${\sim}$73 (DistilBERT, 4 GPUs $\times$ 18\,hr) + ${\sim}$538 (DeBERTa, 4 GPUs
$\times$ 134\,hr) + ${\sim}$105 (ModernBERT) $\approx$ 720 A100 GPU-hours.
No gradient accumulation is used; the per-device batch sizes in
Table~\ref{tab:per-model-hparams} are multiplied by the number of GPUs to obtain
the effective global batch size.

\textbf{Seeds.}
The 20 seeds used across all models are:
\texttt{9999, 1, 2, 3, 4, 5, 6, 7, 8, 9, 997, 420, 1234, 2025, 2024, 666,
247, 42, 31415, 27182}.

\section{Fine-Tuning Curves}
\label{app:training-curves}

\begin{figure*}[t]
\centering
\includegraphics[width=\linewidth]{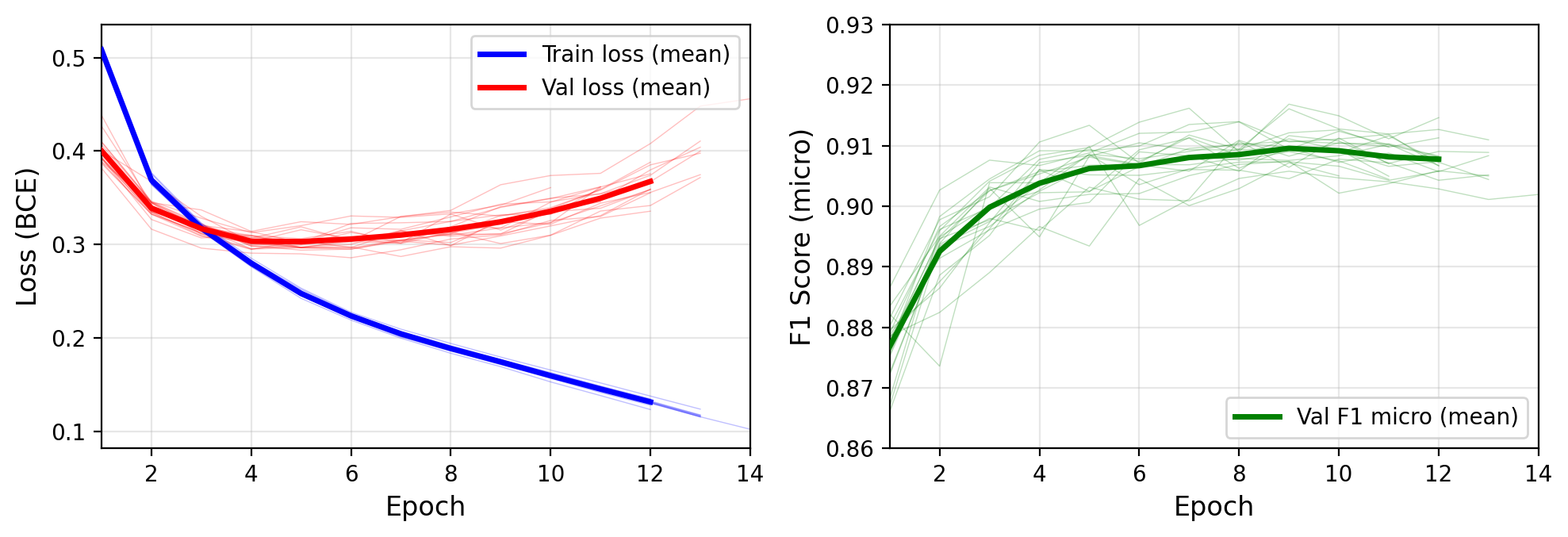}
\caption{DistilBERT fine-tuning curves across 20 seeds. Left: training and
validation loss. Right: validation micro-F1. Faint lines show individual seeds;
bold lines show the mean (over seeds still fine-tuning at each epoch). Early
stopping triggers between epochs 10 and 14 depending on the seed.}
\label{fig:training-curves}
\end{figure*}

Figure~\ref{fig:training-curves} shows the fine-tuning and validation loss curves
for all 20 DistilBERT seeds. Fine-tuning uses binary cross-entropy loss with
AdamW (lr=$5{\times}10^{-5}$, linear schedule with 10\% warmup, weight decay
0.01) for up to 30 epochs, with early stopping (patience~3) on the checkpoint
metric, the decisive accuracy of the best budget-feasible routing policy
(Appendix~\ref{app:training-config}). This metric and the validation curves
below are measured on the selection half of each seed's validation split.

Training loss decreases steadily from 0.51 (epoch~1) to $\sim$0.13 (epoch~12).
Validation loss reaches its minimum at epochs~4--5 (0.303 on average) and rises
thereafter, indicating mild overfitting. Despite the rising validation loss,
validation micro-F1 continues to improve slightly, plateauing around
0.908--0.910 over epochs~7--12, while most seeds are still fine-tuning. This divergence between loss and F1 is
expected for multi-label classification: the loss penalizes calibration
(sigmoid probability), while F1 rewards ranking (threshold at 0.5). We
recalibrate the probabilities after fine-tuning in any case
(Section~\ref{sec:routing-logic}). Early stopping ends fine-tuning between
epochs 10 and 14 across seeds.

The narrow spread across all 20 seeds---standard deviation of F1 is only
0.0030 at epoch~9---confirms that fine-tuning is stable and the final checkpoint
quality is not sensitive to random initialization (see also
Section~\ref{sec:robustness}).

\section{Router Seed Stability}
\label{app:seed-stability}

Every seed draws its own train/validation/test partition
(Appendix~\ref{app:training-config}), so we compare methods seed by seed on
each seed's own test split. Figure~\ref{fig:seed-strip} shows the test answer
accuracy of Switchcraft (DistilBERT encoder), kNN ($k{=}10$), and GPT-5.3-chat
across the 20 seeds.

\begin{figure}[ht]
\centering
\begin{tikzpicture}
\begin{axis}[
    width=0.95\linewidth,
    height=4.2cm,
    xlabel={Test answer accuracy (\%)},
    ytick={0,1,2},
    yticklabels={GPT-5.3-chat, kNN ($k{=}10$), Switchcraft},
    xmin=74.8, xmax=79.6,
    ymin=-0.6, ymax=2.6,
    grid=major,
    grid style={gray!20},
    tick label style={font=\scriptsize},
    label style={font=\small},
    clip=false,
]
\addplot[only marks, mark=*, mark size=2.2pt, red!80!black, opacity=0.7]
    coordinates {
        (78.53,2) (78.72,2) (78.70,2) (78.24,2) (78.13,2)
        (78.27,2) (77.65,2) (78.99,2) (78.08,2) (78.95,2)
        (78.47,2) (78.05,2) (78.24,2) (78.68,2) (77.37,2)
        (78.50,2) (78.57,2) (78.27,2) (77.33,2) (78.21,2)
    };
\addplot[only marks, mark=diamond*, mark size=2.4pt, orange!85!black, opacity=0.7]
    coordinates {
        (76.72,1) (76.95,1) (76.86,1) (76.88,1) (76.44,1)
        (76.49,1) (76.34,1) (76.93,1) (76.71,1) (76.38,1)
        (76.25,1) (76.46,1) (76.80,1) (76.78,1) (76.01,1)
        (76.58,1) (76.50,1) (77.16,1) (76.31,1) (77.13,1)
    };
\addplot[only marks, mark=*, mark size=2.2pt, blue!70!black, opacity=0.7]
    coordinates {
        (75.88,0) (75.70,0) (76.12,0) (76.25,0) (75.43,0)
        (75.63,0) (75.70,0) (76.28,0) (76.66,0) (75.83,0)
        (76.17,0) (75.32,0) (75.83,0) (76.49,0) (75.20,0)
        (76.58,0) (76.38,0) (75.94,0) (75.85,0) (76.36,0)
    };
\addplot[only marks, mark=|, mark size=7pt, black, very thick]
    coordinates {(78.30,2) (76.63,1) (75.98,0)};
\node[font=\scriptsize, anchor=south] at (axis cs:78.30,2.25) {mean 78.30\%};
\node[font=\scriptsize, anchor=south] at (axis cs:76.63,1.25) {mean 76.63\%};
\node[font=\scriptsize, anchor=south] at (axis cs:75.98,0.25) {mean 75.98\%};
\end{axis}
\end{tikzpicture}
\caption{Test answer accuracy across 20 random seeds, each evaluated on its
own test partition. Dots are seeds; black bars mark means. Switchcraft beats
kNN ($k{=}10$) and GPT-5.3-chat on every seed.}
\label{fig:seed-strip}
\end{figure}

Table~\ref{tab:seed-results} reports, for all 20 seeds of each router
encoder, the validation answer accuracy used for seed selection, the test
answer accuracy, and the average test cost per query.

\begin{table*}[t]
\centering
\caption{Answer accuracy (\%) on each seed's validation split (Val, used for
seed selection) and test split (Test), and average test cost per query
($10^{-4}$\,\$), for 20 seeds of DistilBERT, ModernBERT, and DeBERTa-v3
routers. The validation-selected seed per encoder is \textbf{bolded}.}
\label{tab:seed-results}
\small
\setlength{\tabcolsep}{4pt}
\begin{tabular}{r ccc ccc ccc}
\toprule
& \multicolumn{3}{c}{\textbf{DistilBERT}} & \multicolumn{3}{c}{\textbf{ModernBERT}}
& \multicolumn{3}{c}{\textbf{DeBERTa-v3}} \\
\cmidrule(lr){2-4} \cmidrule(lr){5-7} \cmidrule(lr){8-10}
\textbf{Seed} & \textbf{Val} & \textbf{Test} & \textbf{Cost}
              & \textbf{Val} & \textbf{Test} & \textbf{Cost}
              & \textbf{Val} & \textbf{Test} & \textbf{Cost} \\
\midrule
1      & 78.16 & 78.53 & 20.0 & 77.91 & 78.60 & 24.5 & 78.10 & 78.56 & 20.2 \\
2      & 78.30 & 78.72 & 27.8 & 77.90 & 78.32 & 34.4 & 77.89 & 77.73 & 17.0 \\
3      & 77.82 & 78.70 & 34.3 & 77.82 & 78.45 & 24.1 & 77.75 & 78.71 & 15.8 \\
4      & 77.83 & 78.24 & 35.7 & 78.58 & 79.34 & 28.8 & 77.88 & 78.52 & 19.2 \\
5      & 78.72 & 78.13 & 31.1 & 77.87 & 77.67 & 24.3 & 78.64 & 78.46 & 26.0 \\
\textbf{6} & \textbf{79.27} & \textbf{78.27} & \textbf{27.5}
           & \textbf{79.26} & \textbf{78.38} & \textbf{36.1} & 77.68 & 76.48 & 27.8 \\
7      & 77.42 & 77.65 & 24.2 & 78.20 & 78.09 & 35.9 & 77.68 & 77.62 & 21.8 \\
8      & 78.21 & 78.99 & 28.0 & 77.56 & 78.22 & 42.4 & 77.69 & 77.77 & 24.5 \\
9      & 77.50 & 78.08 & 15.4 & 77.92 & 78.41 & 20.6 & 77.58 & 78.31 & 14.4 \\
42     & 78.65 & 78.95 & 34.5 & 78.46 & 78.42 & 44.7 & 78.80 & 78.77 & 37.5 \\
\textbf{247} & 78.84 & 78.47 & 40.2 & 79.06 & 78.50 & 27.6
             & \textbf{78.90} & \textbf{78.58} & \textbf{27.0} \\
420    & 78.16 & 78.05 & 27.9 & 78.07 & 78.19 & 50.0 & 78.11 & 78.09 & 28.1 \\
666    & 78.48 & 78.24 & 32.7 & 78.27 & 77.84 & 22.1 & 78.45 & 78.00 & 30.3 \\
997    & 78.02 & 78.68 & 33.2 & 78.03 & 78.50 & 26.0 & 77.38 & 77.94 & 33.5 \\
1234   & 77.88 & 77.37 & 21.2 & 77.87 & 78.07 & 26.3 & 77.78 & 77.53 & 30.0 \\
2024   & 77.22 & 78.50 & 17.6 & 77.64 & 78.66 & 35.0 & 77.66 & 78.38 & 14.1 \\
2025   & 78.78 & 78.57 & 36.1 & 78.49 & 77.84 & 29.1 & 78.79 & 78.16 & 37.5 \\
9999   & 78.39 & 78.27 & 22.8 & 78.80 & 78.45 & 38.7 & 77.69 & 77.57 & 24.6 \\
27182  & 78.37 & 77.33 & 41.4 & 79.24 & 78.12 & 22.2 & 78.24 & 77.65 & 20.2 \\
31415  & 77.02 & 78.21 & 28.3 & 77.48 & 78.64 & 25.7 & 77.74 & 79.25 & 23.1 \\
\midrule
Mean   & 78.15 & 78.30 & 29.0 & 78.22 & 78.34 & 30.9 & 78.02 & 78.10 & 24.6 \\
Std    & $\pm$0.58 & $\pm$0.46 & $\pm$7.2 & $\pm$0.54 & $\pm$0.36 & $\pm$8.3
       & $\pm$0.46 & $\pm$0.60 & $\pm$7.0 \\
\bottomrule
\end{tabular}
\end{table*}

All 20 DistilBERT seeds achieve test answer accuracy between 77.33\% and
78.99\%, a range of only 1.66 percentage points, with a mean of 78.30\%
($\pm$0.46). ModernBERT shows a similar pattern: its 20 seeds span
77.67\%--79.34\% (1.66\,pp range, mean 78.34\% $\pm$0.36), and DeBERTa-v3 is
comparable: 76.48\%--79.25\% (2.77\,pp range, mean 78.10\% $\pm$0.60).
Validation-selected seeds are not test-optimal---DistilBERT's seed~6
(78.27\%) sits at its mean---so the reported numbers do not benefit from
test-set selection. Average cost varies more across seeds
($14$--$50 \times 10^{-4}$\,\$) because each seed's thresholds are tuned
separately. Seed by seed, DistilBERT beats GPT-5.3-chat on all 20 test
partitions (by 1.43--3.12\,pp, mean 2.32\,pp) at lower cost, and beats kNN
($k{=}10$) on all 20 (by at least 1.03\,pp, and at least 1.89\,pp on decisive
queries). ModernBERT also beats both baselines on every seed, and
DeBERTa-v3 beats GPT-5.3-chat on every seed and kNN on 19 of 20.
This confirms that Switchcraft's advantage over single-model and retrieval
baselines is not an artifact of seed selection.

\section{Decision-Policy Tuning and Operating Points}
\label{app:adaptive-threshold}

\subsection{Motivation}

Our original decision rule applied a fixed sigmoid threshold ($\theta=0.5$)
and selected the cheapest model above it, falling back to argmax. It has two
weaknesses for agentic routing: raw sigmoid outputs are not calibrated, so
one threshold means different things for different models, and a fixed rule
cannot express how much cost a deployment will tolerate for accuracy. An
earlier study under our previous protocol (Appendix~\ref{app:training-config})
already showed that alternative rules expose a broad frontier: raising the
threshold to 0.75, or selecting the cheapest model whose probability is
within $\delta$ of the maximum, moved accuracy from 81.9\% to 85.2\% as cost
grew from $7.3$ to $41.7 \times 10^{-4}$\,\$ per query (means over 20 seeds).
Switchcraft's tuned policy (Section~\ref{sec:routing-logic}) makes this
trade-off explicit: calibration makes thresholds comparable across models,
and the cost budget selects the operating point.

\subsection{Operating Points}

Table~\ref{tab:budget-sweep} varies the budget for one fine-tuned checkpoint
(DistilBERT, seed~6). For each budget, $(\tau_r, \tau_a)$ is tuned on the
selection half exactly as in Section~\ref{sec:routing-logic}, including the
55\% impossible-detection floor, and then evaluated once on the test set
with realized costs. Table~\ref{tab:policy-ablation} isolates the
contributions of calibration, threshold tuning, and abstention at the
default budget.

\begin{table*}[t]
\centering
\caption{Operating points of the tuned policy (DistilBERT, seed~6).
Thresholds are tuned on the selection half under each profiled-cost budget;
accuracies (\%) and realized cost are measured on the test set. Costs are in
$10^{-4}$\,\$.}
\label{tab:budget-sweep}
\small
\begin{tabular}{rcccccr}
\toprule
\textbf{Budget} & \textbf{$\boldsymbol{\tau_r}$} & \textbf{$\boldsymbol{\tau_a}$}
  & \textbf{Answer} & \textbf{Routing} & \textbf{Decisive} & \textbf{Cost} \\
\midrule
3                 & 0.35 & 0.60 & 69.40 & 78.93 & 72.45 & 3.5 \\
5                 & 0.55 & 0.70 & 72.06 & 80.87 & 76.87 & 7.9 \\
8                 & 0.70 & 0.75 & 75.04 & 83.17 & 81.90 & 12.6 \\
12                & 0.80 & 0.75 & 76.74 & 84.88 & 84.79 & 19.8 \\
\textbf{17} (default) & 0.90 & 0.75 & 78.27 & 86.40 & 87.39 & 27.5 \\
25                & 0.95 & 0.75 & 79.23 & 87.37 & 89.03 & 37.7 \\
40                & 1.00 & 0.75 & 79.75 & 87.88 & 89.90 & 43.6 \\
\bottomrule
\end{tabular}
\end{table*}

\subsection{Findings}

\begin{itemize}
    \item \textbf{Over 10 percentage points of accuracy are available}
          by relaxing the budget: the frontier spans 69.4\% at
          $3.5 \times 10^{-4}$\,\$ to 79.7\% at $43.6 \times 10^{-4}$\,\$
          (decisive accuracy 72.4\% to 89.9\%).
    \item \textbf{Every budget from 12 upward beats the best single model.}
          Even the most accurate point ($\tau_r = 1$, which sends every query it does not abstain
          on to its most-likely-correct model) is 4.1\,pp more accurate than
          GPT-5.3-chat at 34\% lower cost, and the budget-12 point is
          1.1\,pp more accurate at 70\% lower cost.
    \item \textbf{Abstention is stable.} Once the budget allows, the
          abstention threshold settles at 0.75, abstaining on 9.0\% of
          queries and catching 56.9\% of impossible ones; tighter budgets
          abstain more (up to 11.5\% of queries), since abstentions are free.
    \item \textbf{Realized cost exceeds the profiled budget} at most
          operating points (by up to 65\%): the queries a stricter policy
          sends to expensive models are disproportionately long multi-turn
          ones, whose realized cost far exceeds the per-model averages the
          policy sees (Appendix~\ref{app:cost-model}).
\end{itemize}

These results suggest that cost-sensitive deployments can lower the budget
toward $8$--$12 \times 10^{-4}$\,\$ while staying near or above the best
single model's accuracy, and accuracy-sensitive ones can raise it to
$25 \times 10^{-4}$\,\$ for roughly one more point of accuracy.

\section{Ablation: Token Packing}
\label{app:ablation-packing}

Switchcraft packs each query into the encoder's 512-token window
(Section~\ref{sec:input-representation}): a metadata line, the latest turn,
tool signatures capped at 100 tokens, and as many earlier turns as fit,
newest first. To isolate this design, we fine-tune DistilBERT routers whose
only change is their input: the conversation followed by all of its tool
signatures, with no metadata and no cap on the signatures, right-truncated at
512 tokens, so a long input loses its tool signatures first and then the end
of its latest turn. Everything else follows the protocol of
Table~\ref{tab:main-results}: hyperparameters, checkpoint selection,
calibration, abstention head, and thresholds tuned under the same cost budget
and impossible-query floor. We train a router without packing for each of the
20 seeds and compare it with the packed router on that seed's own partition.

\begin{table*}[t]
\centering
\caption{Token packing ablation (DistilBERT): answer, routing, and decisive
accuracy (\%) and cost ($10^{-4}$\,\$). Top: test-set mean $\pm$ sample std
over 20 seeds, each router scored on each seed's own partition. The $\Delta$
row gives the mean paired difference, no packing minus packing (for cost, the
mean per-seed relative change), and in parentheses the number of seeds on
which packing is more accurate (for cost, cheaper). Bottom: seed~6, selected
on validation for both routers.}
\label{tab:ablation-packing}
\small
\begin{tabular}{lcccc}
\toprule
\textbf{Input} & \textbf{Answer} & \textbf{Routing} & \textbf{Decisive}
  & \textbf{Cost} \\
\midrule
\multicolumn{5}{l}{\emph{Test set, mean over 20 seeds}} \\
Token packing (ours) & 78.30\,$\pm$\,0.46 & 86.80\,$\pm$\,0.72
  & 87.31\,$\pm$\,0.67 & 29.0\,$\pm$\,7.2 \\
No packing           & 77.80\,$\pm$\,0.59 & 86.08\,$\pm$\,0.62
  & 86.46\,$\pm$\,0.89 & 31.8\,$\pm$\,6.6 \\
$\Delta$ (seeds)     & $-$0.50 (16) & $-$0.72 (16) & $-$0.85 (16)
  & $+$13.6\% (15) \\
\midrule
\multicolumn{5}{l}{\emph{Test set, seed~6}} \\
Token packing (ours) & 78.27 & 86.40 & 87.39 & 27.5 \\
No packing           & 78.42 & 86.48 & 87.70 & 46.6 \\
\midrule
\multicolumn{5}{l}{\emph{Held-out Glaive, seed~6}} \\
Token packing (ours) & 89.20 & 89.39 & 79.32 & 1.46 \\
No packing           & 89.58 & 89.76 & 81.33 & 1.39 \\
\bottomrule
\end{tabular}
\end{table*}

\textbf{Results.}
Packing makes the router modestly but consistently more accurate
(Table~\ref{tab:ablation-packing}). Across the 20 seeds, removing it lowers
answer accuracy by 0.50\,pp on average (sample std 0.52\,pp), routing
accuracy by 0.72\,pp, and decisive accuracy by 0.85\,pp, and each of these
drops on 16 of the 20 seeds; per seed, the change in answer accuracy ranges
from $-$1.34 to $+$0.36\,pp. Packing also lowers cost on 15 of the 20 seeds
(mean 29.0 vs.\ $31.8 \times 10^{-4}$\,\$), but the per-seed change is noisy,
from $-$33.9\% to $+$69.6\%, because in-distribution cost is dominated by the
expensive BFCL multi-turn queries (Appendix~\ref{app:per-dataset}) and shifts
with how they are routed.

\textbf{Seed 6.}
Validation selects seed~6 for both routers (validation answer accuracy
79.27\% with packing and 79.25\% without), so our main results use it. It is
one of the four seeds on which the router without packing is slightly more
accurate ($+$0.15\,pp answer, $+$0.31\,pp decisive), and the one on which it
costs the most extra ($+$69.6\%), almost entirely on the 150 BFCL multi-turn
queries, whose average cost nearly doubles, from $1{,}343$ to
$2{,}602 \times 10^{-4}$\,\$. Both routers select the same thresholds
($\tau_r = 0.90$, $\tau_a = 0.75$) and abstain at similar rates (9.01\% with
packing and 8.91\% without). On held-out Glaive, the seed-6 router without
packing is slightly more accurate ($+$0.38\,pp answer, $+$2.01\,pp decisive)
and 4.5\% cheaper.

\textbf{Where packing matters.}
Most inputs fit in the window: on seed~6's test split, the input without
packing has a median of 86 tokens (mean 207), and 11.7\% of queries exceed
the window. Table~\ref{tab:ablation-packing-truncation} groups these queries
by what right-truncation removes. Even on this seed, where packing does not
help overall, the packed router is 4.18\,pp more accurate (6.29\,pp on
decisive queries) on the 2.5\% of queries whose earlier turns alone fill the
window, so that right-truncation removes the whole latest turn; on the other
groups, the router without packing is slightly more accurate.

\begin{table*}[t]
\centering
\caption{Seed-6 test queries grouped by what right-truncation at 512 tokens
removes from the input without packing: nothing; only tool signatures; all
tool signatures and part of the latest turn; or all tool signatures and the
whole latest turn. Answer accuracy (\%) and cost ($10^{-4}$\,\$) of the packed
router and the router without packing; $\Delta$ is computed before rounding.}
\label{tab:ablation-packing-truncation}
\small
\setlength{\tabcolsep}{4pt}
\begin{tabular}{lrccccc}
\toprule
& & \multicolumn{3}{c}{\textbf{Answer accuracy}}
  & \multicolumn{2}{c}{\textbf{Cost}} \\
\cmidrule(lr){3-5} \cmidrule(lr){6-7}
\textbf{Removed} & \textbf{Queries} & \textbf{Packing}
  & \textbf{No packing} & \textbf{$\Delta$} & \textbf{Packing}
  & \textbf{No packing} \\
\midrule
Nothing                 & 88.3\% & 79.56 & 79.75 & $+$0.19 & 23.16 & 44.10 \\
Tool signatures only    &  8.0\% & 70.66 & 71.56 & $+$0.89 & 69.70 & 74.25 \\
Part of latest turn     &  1.2\% & 59.48 & 61.21 & $+$1.72 & 67.31 & 47.62 \\
Whole latest turn       &  2.5\% & 65.69 & 61.51 & $-$4.18 & 25.14 & 44.98 \\
\bottomrule
\end{tabular}
\end{table*}

\textbf{Implications.}
Packing adds about half a point of answer accuracy, comparable to the
seed-to-seed spread (Appendix~\ref{app:seed-stability}) but consistent in
direction, and usually lowers cost. Its clearest benefit is protecting the
latest turn when earlier turns would fill the window, which is rare in our
corpus but becomes more common as conversations grow longer.

\section{Router Inference Latency}
\label{app:latency}

A practical router should add as little latency as possible before dispatching
to the selected LLM to avoid inflating time-to-first-token (TTFT). We benchmark the three
fine-tuned BERT-based router encoders -- DistilBERT, ModernBERT, and DeBERTa-v3
-- under the same inference workload. This appendix reports their latency,
throughput, and GPU memory use. These measurements focus on inference overhead
rather than correctness.

\textbf{Infrastructure choice.}
We benchmark on a single NVIDIA T4 GPU (16\,GB, Turing architecture), a widely
available inference accelerator in public clouds (AWS \texttt{g4dn}, Azure
\texttt{NC\_T4\_v3}, GCP \texttt{n1-standard + T4}).  The T4 represents a
low-cost GPU option (\textasciitilde\$0.35/hr spot\footnote{Based on AWS
\texttt{g4dn.xlarge} spot pricing as of April 2026:
\url{https://aws.amazon.com/ec2/spot/pricing/}.}); high performance on this
hardware would illustrate cheap inference cost without requiring
specialized accelerators.

\textbf{Methodology.}
We load each router model's validation-selected checkpoint and run inference on synthetic
inputs at three sequence lengths: 64, 200, and 512 tokens. For each
configuration we perform 50 warmup iterations followed by 200 timed iterations,
reporting wall-clock latency percentiles (P50, P95, P99) and sustained
throughput (queries/sec). All measurements use \texttt{torch.cuda.synchronize()}
barriers for accurate timing. We report FP32 precision with PyTorch 2.6 and CUDA
12.4. We report GPU memory allocated and reserved after loading each model. For
most of the results reported below, GPU utilization (the mean SM activity) was
close to 100\%.

\textbf{Results.}
Tables~\ref{tab:latency-t4}, \ref{tab:latency-t4-modernbert}, and
\ref{tab:latency-t4-deberta} present the full measurements for DistilBERT,
ModernBERT, and DeBERTa-v3, respectively.

\begin{table*}[t]
\centering
\caption{DistilBERT router inference (67.0\,M parameters, 283.6\,MB GPU memory allocated, 314.0\,MB reserved) on NVIDIA T4, FP32.}
\label{tab:latency-t4}
\small
\begin{tabular}{rrrrrr}
\toprule
\textbf{Batch} & \textbf{Seq Len} & \textbf{P50 (ms)} & \textbf{P95 (ms)} & \textbf{P99 (ms)} & \textbf{Throughput (qps)} \\
\midrule
1   & 64  & 3.22    & 3.26    & 3.35    & 310 \\
8   & 64  & 12.23   & 12.57   & 12.62   & 654 \\
32  & 64  & 46.73   & 47.45   & 47.58   & 685 \\
64  & 64  & 86.49   & 87.61   & 88.09   & 739 \\
128 & 64  & 187.97  & 191.14  & 192.10  & 682 \\
\midrule
1   & 200 & 7.05    & 7.20    & 7.22    & 142 \\
8   & 200 & 38.74   & 39.28   & 39.29   & 206 \\
32  & 200 & 156.66  & 158.35  & 159.07  & 204 \\
64  & 200 & 326.96  & 331.95  & 332.92  & 196 \\
128 & 200 & 653.38  & 658.64  & 659.50  & 196 \\
\midrule
1   & 512 & 16.02   & 16.43   & 16.65   & 62 \\
8   & 512 & 112.14  & 114.91  & 115.32  & 71 \\
32  & 512 & 485.17  & 486.69  & 487.30  & 66 \\
64  & 512 & 953.77  & 958.28  & 960.47  & 67 \\
128 & 512 & 1909.03 & 1917.14 & 1920.32 & 67 \\
\bottomrule
\end{tabular}
\end{table*}

\begin{table*}[t]
\centering
\caption{ModernBERT router inference (149.6\,M parameters, 870.7\,MB GPU memory allocated, 1124.0\,MB reserved) on NVIDIA T4, FP32.}
\label{tab:latency-t4-modernbert}
\small
\begin{tabular}{rrrrrr}
\toprule
\textbf{Batch} & \textbf{Seq Len} & \textbf{P50 (ms)} & \textbf{P95 (ms)} & \textbf{P99 (ms)} & \textbf{Throughput (qps)} \\
\midrule
1   & 64  & 18.76   & 19.00   & 19.40   & 53 \\
8   & 64  & 32.95   & 33.38   & 33.43   & 243 \\
32  & 64  & 136.56  & 138.48  & 138.98  & 234 \\
64  & 64  & 278.69  & 285.36  & 286.68  & 229 \\
128 & 64  & 545.45  & 554.99  & 556.30  & 235 \\
\midrule
1   & 200 & 21.51   & 21.96   & 22.16   & 46 \\
8   & 200 & 131.31  & 133.00  & 133.35  & 61 \\
32  & 200 & 536.99  & 539.38  & 539.99  & 60 \\
64  & 200 & 945.42  & 960.14  & 961.19  & 67 \\
128 & 200 & 1894.98 & 1901.11 & 1903.77 & 68 \\
\midrule
1   & 512 & 43.99   & 44.68   & 52.95   & 23 \\
8   & 512 & 339.79  & 341.25  & 341.99  & 24 \\
32  & 512 & 1300.24 & 1305.45 & 1307.81 & 25 \\
64  & 512 & 2608.43 & 2617.33 & 2618.18 & 25 \\
128 & 512 & 5222.83 & 5238.58 & 5242.05 & 25 \\
\bottomrule
\end{tabular}
\end{table*}

\begin{table*}[t]
\centering
\caption{DeBERTa-v3 router inference (184.4\,M parameters, 814.9\,MB GPU memory allocated, 834.0\,MB reserved) on NVIDIA T4, FP32.}
\label{tab:latency-t4-deberta}
\small
\begin{tabular}{rrrrrr}
\toprule
\textbf{Batch} & \textbf{Seq Len} & \textbf{P50 (ms)} & \textbf{P95 (ms)} & \textbf{P99 (ms)} & \textbf{Throughput (qps)} \\
\midrule
1   & 64  & 16.21   & 16.36   & 16.46   & 62 \\
8   & 64  & 39.07   & 39.59   & 39.79   & 205 \\
32  & 64  & 141.71  & 144.01  & 144.58  & 226 \\
64  & 64  & 266.91  & 269.25  & 269.53  & 240 \\
128 & 64  & 554.96  & 558.33  & 559.41  & 231 \\
\midrule
1   & 200 & 23.65   & 24.20   & 24.23   & 42 \\
8   & 200 & 123.60  & 124.89  & 125.05  & 65 \\
32  & 200 & 484.78  & 485.97  & 486.65  & 66 \\
64  & 200 & 957.34  & 965.23  & 967.61  & 67 \\
128 & 200 & 1817.45 & 1825.12 & 1827.43 & 70 \\
\midrule
1   & 512 & 53.52   & 54.08   & 54.62   & 19 \\
8   & 512 & 356.02  & 357.78  & 358.09  & 22 \\
32  & 512 & 1458.65 & 1462.86 & 1465.00 & 22 \\
64  & 512 & 2893.57 & 2902.08 & 2904.23 & 22 \\
128 & 512 & 5807.98 & 5825.37 & 5828.51 & 22 \\
\bottomrule
\end{tabular}
\end{table*}

\textbf{Summary.}
For single-query routing at intermediate prompt lengths (200 tokens), P95
latency is 7.20\,ms for DistilBERT, 21.96\,ms for ModernBERT, and 24.20\,ms for
DeBERTa-v3. At 512 tokens, the corresponding P95 latencies are 16.43\,ms,
44.68\,ms, and 54.08\,ms. Across all tested configurations, peak throughput is
739 queries/sec for DistilBERT (batch size 64, 64 tokens), 243 queries/sec for
ModernBERT (batch size 8, 64 tokens), and 240 queries/sec for DeBERTa-v3 (batch
size 64, 64 tokens). On the in-distribution test set
(Table~\ref{tab:main-results}), answer accuracy is 78.3\% for DistilBERT,
78.4\% for ModernBERT, and 78.6\% for DeBERTa-v3, a spread of just 0.3\,pp. The corresponding mean costs of the selected LLMs are
$(27.5,\,36.1,\,27.0)\times 10^{-4}$\,\$ per query, respectively. These results
show similar accuracy and close costs for DistilBERT and DeBERTa-v3, while
ModernBERT has a higher measured LLM cost. When accuracy and LLM cost are
comparable, router latency, throughput, and deployment cost become important
tie-breakers. On this T4 workload, the inference measures alone favor
DistilBERT, though context-length needs would favor ModernBERT.

\textbf{Serving cost.}
Under a 20\,ms P99 budget, the largest DistilBERT batches that fit are 8 at 64
tokens (654 queries/sec), 1 at 200 tokens (142 queries/sec), and 1 at 512
tokens (62 queries/sec) (Table~\ref{tab:latency-t4}). At \textasciitilde\$0.35/hr
spot, one T4 therefore routes a query for $1.5$--$15.7 \times 10^{-7}$\,\$
(\$0.15--\$1.57 per million queries), under 0.06\% of the
$27.5 \times 10^{-4}$\,\$ per query that Switchcraft spends on LLM calls
(Table~\ref{tab:main-results}).

\section{Misrouted Query Breakdowns}
\label{app:misrouted-details}

This appendix provides detailed breakdowns of the 1{,}325 misrouted queries
(13.6\% of the 9{,}746-query test set) from Switchcraft's reported
configuration (DistilBERT, seed~6). A query is \emph{misrouted} if its routing
decision is wrong: the selected model fails although a correct model exists
(\emph{wrong model}: 640 cases), the router abstains although a correct model
exists (\emph{false abstention}: 85), or no model is correct but the router
routes anyway (\emph{missed abstention}: 600). Consistent with routing
accuracy (Section~\ref{sec:eval-setup}), the 793 correct abstentions count as
correctly routed.

\subsection*{Misroute Rate by Number of Tool Definitions}

\begin{table}[ht]
\centering
\caption{Misroute rate by number of tool definitions.}
\label{tab:misroute-tools}
\small
\begin{tabular}{lrrr}
\toprule
\textbf{Tools} & \textbf{Misrouted} & \textbf{Total} & \textbf{Rate (\%)} \\
\midrule
1         &   187 & 1{,}576 & 11.9 \\
2--3      &   418 & 3{,}410 & 12.3 \\
4--6      &   240 & 1{,}780 & 13.5 \\
7--10     &    34 &   242 & 14.0 \\
11--50    &   446 & 2{,}738 & 16.3 \\
\bottomrule
\end{tabular}
\end{table}

As shown in Table~\ref{tab:misroute-tools}, the misroute rate increases
monotonically with the number of tool definitions in the query, rising from
11.9\% for single-tool queries to 16.3\% for queries with 11--50 tools (most
of which come from FunReason-MT). Queries with many tools tend to have longer
function schemas that may exceed the router's 512-token context window,
causing information loss.

\subsection*{Misroute Rate by Number of Conversation Turns}

\begin{table}[ht]
\centering
\caption{Misroute rate by number of conversation turns.}
\label{tab:misroute-turns}
\small
\begin{tabular}{lrrr}
\toprule
\textbf{Turns} & \textbf{Misrouted} & \textbf{Total} & \textbf{Rate (\%)} \\
\midrule
1         & 1{,}017 & 8{,}037 & 12.7 \\
3--4      &    40 &   314 & 12.7 \\
5--8      &   254 & 1{,}326 & 19.2 \\
9--20     &    13 &    68 & 19.1 \\
21+       &     1 &     1 & 100.0 \\
\bottomrule
\end{tabular}
\end{table}

Table~\ref{tab:misroute-turns} shows that short conversations (up to four
turns) are misrouted 12.7\% of the time, while longer ones (5--20 turns) reach
about 19\%; the sample of very long conversations is small (69 queries with 9
or more turns).

\subsection*{Misroute Rate by Text Length}

\begin{table}[ht]
\centering
\caption{Misroute rate by text length (in whitespace-delimited tokens).}
\label{tab:misroute-length}
\small
\begin{tabular}{lrrr}
\toprule
\shortstack[l]{\textbf{Text length}\\\textbf{(tokens)}} & \textbf{Misrouted} & \textbf{Total} & \textbf{Rate (\%)} \\
\midrule
2--15       &   154 & 2{,}072 &  7.4 \\
15--23      &   349 & 2{,}686 & 13.0 \\
23--42      &   381 & 2{,}534 & 15.0 \\
42--108     &   190 & 1{,}477 & 12.9 \\
108--1{,}442 & 251 &   977 & 25.7 \\
\bottomrule
\end{tabular}
\end{table}

Table~\ref{tab:misroute-length} shows that very short queries ($<$15 tokens)
are the easiest to route (7.4\%), likely because they are simple
single-function calls. Mid-range queries are harder, and the longest bucket
(108+ tokens) reaches 25.7\%.

\subsection*{Misroute Rate by Number of Correct Models}

\begin{table}[ht]
\centering
\caption{Misroute rate by number of correct models in the pool.}
\label{tab:misroute-correct-models}
\small
\begin{tabular}{lrrr}
\toprule
\textbf{Models Correct} & \textbf{Misrouted} & \textbf{Total} & \textbf{Rate (\%)} \\
\midrule
0 &   600 & 1{,}393 & 43.1 \\
1 &   249 &   397 & 62.7 \\
2 &   114 &   340 & 33.5 \\
3 &    89 &   381 & 23.4 \\
4 &    63 &   396 & 15.9 \\
5 &    81 &   529 & 15.3 \\
6 &   103 & 1{,}185 &  8.7 \\
7 &    24 & 2{,}506 &  1.0 \\
8 &     2 & 2{,}619 &  0.1 \\
\bottomrule
\end{tabular}
\end{table}

Table~\ref{tab:misroute-correct-models} reveals the strongest predictor of
routing difficulty. When all 8 models answer correctly, the misroute rate is
0.1\%---the only two errors are false abstentions. When only 1 model is
correct, the router must identify that specific model among 8 candidates, and
the error rate climbs to 62.7\%. Of the 1{,}393 queries where no model is
correct, the router abstains on 793 (56.9\%); the other 600 are missed
abstentions and account for 45.3\% of all misroutes.

\subsection*{Predicted vs.\ Oracle Model Confusion}

Among the 640 \emph{wrong-model} errors (where a correct model existed but the
router chose one that failed), Table~\ref{tab:misroute-confusion} shows the
predicted and oracle model distributions:

\begin{table}[ht]
\centering
\caption{Predicted vs.\ oracle model distribution for the 640 wrong-model errors.}
\label{tab:misroute-confusion}
\small
\begin{tabular}{lrr}
\toprule
\textbf{Model} & \textbf{Predicted (\%)} & \textbf{Oracle (\%)} \\
\midrule
GPT-5-nano    & 21.9 & 25.9 \\
Qwen-3.5-9B   & 19.2 & 32.2 \\
GPT-5.3-chat  & 14.4 &  5.8 \\
GPT-5.4-nano  & 13.4 & 13.9 \\
GPT-5-mini    & 12.8 &  9.5 \\
GPT-5.4       &  7.0 &  2.7 \\
GPT-5.4-mini  &  5.8 &  5.3 \\
Kimi-K2.5     &  5.5 &  4.7 \\
\bottomrule
\end{tabular}
\end{table}

Switchcraft over-predicts the two most expensive models, GPT-5.3-chat
(14.4\% of wrong predictions vs.\ 5.8\% of oracle selections) and GPT-5.4
(7.0\% vs.\ 2.7\%), while under-predicting Qwen (19.2\% vs.\ 32.2\%). The top
confusion flows are between the cheapest models:
\begin{itemize}
    \item GPT-5-nano $\rightarrow$ Qwen (75 cases): the router predicts
          nano, while Qwen would have been the cheapest correct
          model.
    \item Qwen $\rightarrow$ GPT-5-nano (54 cases): the reverse confusion.
    \item GPT-5.4-nano $\rightarrow$ Qwen (33 cases), GPT-5-mini
          $\rightarrow$ Qwen (32), and GPT-5.3-chat $\rightarrow$ Qwen (32):
          a costlier prediction where Qwen succeeds.
\end{itemize}

\section{Per-Model Token Usage and Chattiness Analysis}
\label{app:token-usage}

Token counts are extracted from each model's API response metadata.
Every result record includes \texttt{input\_token\_count},
\texttt{output\_token\_count}, and (where applicable)
\texttt{reasoning\_token\_count}.
The output token count corresponds to the API's \texttt{completion\_tokens}
field, which \emph{includes} reasoning tokens; the reasoning token count is
the subset reported via \texttt{completion\_\allowbreak tokens\_details.\allowbreak reasoning\_tokens}.
Thus the visible (non-reasoning) output is
$\text{output} - \text{reasoning}$ tokens.
For multi-turn conversations, token counts are stored as nested lists
(one entry per turn); we sum across all turns to obtain the per-example total.
Reasoning tokens are billed at the output token rate but are not
returned in the response text.

Table~\ref{tab:token-usage} reports the average input, output, and reasoning
token counts per query for each model on the 9{,}732-example validation split
of seed~6 (the partition behind Table~\ref{tab:main-results}), alongside
per-token pricing.

\begin{table*}[t]
\centering
\caption{Per-model token consumption on the seed-6 validation split (9{,}732 examples).
Avg In = average input (prompt) tokens per query.
Avg Out = average total output (completion) tokens per query, which
\emph{includes} reasoning tokens.
Avg Reas = average reasoning tokens per query (a subset of Avg Out).
\% Reas = percentage of queries that consume any reasoning tokens.
Chatt.\ = chattiness (actual / expected cost; derivation below).
Pricing is in USD per million tokens.
See the caveat on input token counts at the end of this section.}
\label{tab:token-usage}
\small
\begin{tabular}{l rr rrrr r}
\toprule
\textbf{Model} & \textbf{In \$/M} & \textbf{Out \$/M}
  & \textbf{Avg In} & \textbf{Avg Out} & \textbf{Avg Reas} & \textbf{\% Reas}
  & \textbf{Chatt.} \\
\midrule
GPT-5.4       & 2.50 & 15.00 & 2{,}919 & 111 &  46 & 66.8 & 0.90$\times$ \\
GPT-5.4-mini  & 0.75 &  4.50 & 2{,}905 & 106 &  40 & 54.6 & 0.90$\times$ \\
GPT-5.4-nano  & 0.20 &  1.25 & 3{,}027 &  90 &  28 & 34.5 & 0.87$\times$ \\
GPT-5.3-chat  & 1.75 & 14.00 & 2{,}929 &  71 &   9 & 11.6 & 0.81$\times$ \\
GPT-5-mini    & 0.25 &  2.00 & 3{,}018 & 180 & 109 & 66.6 & 1.01$\times$ \\
GPT-5-nano    & 0.05 &  0.40 & 2{,}971 & 272 & 206 & 85.9 & 1.18$\times$ \\
Kimi-K2.5     & 0.60 &  3.00 & 1{,}350 & 261 &   0 &  0.0 & 1.20$\times$ \\
Qwen-3.5-9B   & 0.05 &  0.15 & 5{,}642 & 299 &   0 &  0.0 & 1.06$\times$ \\
\bottomrule
\end{tabular}
\end{table*}

Several patterns emerge. First, \textbf{reasoning token usage varies
dramatically}: GPT-5-nano uses reasoning tokens on 86\% of queries (206 avg),
while GPT-5.3-chat uses them on only 12\% (9 avg). This hidden cost
substantially inflates GPT-5-nano's per-query expense beyond what its
ultra-low per-token price (\$0.05/M input) would suggest. Second,
\textbf{output verbosity and reasoning are distinct phenomena}: Qwen-3.5-9B
produces the most total output tokens (299 avg, all visible) but zero
reasoning tokens, whereas GPT-5-nano's 272 avg output tokens include 206
reasoning tokens---leaving only 66 visible output tokens, in line with the
most concise models (61--66).

\textbf{Profiled per-query cost.}
The router's cost-based tie-breaking uses \emph{profiled costs}: the actual
dollar cost of each model computed from its observed token consumption and
per-token pricing, rather than from list price alone.
Table~\ref{tab:profiled-cost} reports the average profiled cost per query
for each model, computed over all 99{,}125 raw in-distribution examples
(before the pipeline's deduplication step that yields 97{,}369 records for
splitting; see Section~\ref{sec:datasets}). The same per-model cost will
differ modestly across splits because the share of expensive multi-turn
queries varies: e.g., GPT-5.3-chat averages $63.7 \times 10^{-4}$\,\$ on the
full raw corpus, $61.1 \times 10^{-4}$\,\$ on the validation split
(Table~\ref{tab:cost-decomp}), and $66.2 \times 10^{-4}$\,\$ on the test set
(Table~\ref{tab:main-results}). These differences reflect split composition,
not pricing changes.

\begin{table}[ht]
\centering
\caption{Profiled average inference cost per query ($10^{-4}$\,\$), computed from observed
token consumption across all 99{,}125 raw in-distribution examples
(pre-deduplication; see Section~\ref{sec:datasets}). Models are sorted
by ascending cost. This ordering determines which model the router selects when
multiple candidates are predicted correct.}
\label{tab:profiled-cost}
\small
\begin{tabular}{l r}
\toprule
\textbf{Model} & \shortstack{\textbf{Avg cost / query}\\\textbf{($10^{-4}$\,\$)}} \\
\midrule
GPT-5-nano    & 2.7 \\
Qwen-3.5-9B   & 3.4 \\
GPT-5.4-nano  & 7.9 \\
GPT-5-mini    & 11.9 \\
Kimi-K2.5     & 16.1 \\
GPT-5.4-mini  & 28.0 \\
GPT-5.3-chat  & 63.7 \\
GPT-5.4       & 96.7 \\
\bottomrule
\end{tabular}
\end{table}

The profiled cost ordering matches the list-price ordering
(Table~\ref{tab:main-results}) for every model except the two cheapest, which
swap places: Qwen-3.5-9B has the lower list price (the same \$0.05/M input
rate as GPT-5-nano and a cheaper \$0.15/M output rate), yet its profiled cost
is 29\% higher ($3.4$ vs.\ $2.7 \times 10^{-4}$\,\$), because its tokenizer
produces about twice as many input tokens and it generates more output. A
price-based ranking would therefore prefer the wrong model in the cheapest
routing tier---the tier that serves the most queries (Switchcraft sends 41\%
of test queries to GPT-5-nano).

\textbf{Cost decomposition.}
Table~\ref{tab:cost-decomp} breaks the per-query cost into its input and
output components, revealing how much each contributes to the total.
For most models input cost accounts for 58--86\% of total cost, meaning that
output token differences---the driver of the chattiness metric
(Section~\ref{sec:chattiness})---are compressed when viewed through the
lens of total cost.

\begin{table*}[t]
\centering
\caption{Per-query cost decomposition on the seed-6 validation split (9{,}732
examples). Input Cost and Output Cost are the average per-query costs
(\texttt{avg\_input\_tokens} $\times$ \texttt{input\_rate} and
\texttt{avg\_output\_tokens} $\times$ \texttt{output\_rate}, respectively).
Input~\% is the share of total cost attributable to input tokens.
Models are sorted by ascending total cost.}
\label{tab:cost-decomp}
\small
\begin{tabular}{l rrr r}
\toprule
\textbf{Model} & \textbf{Input cost} & \textbf{Output cost}
  & \textbf{Total cost} & \textbf{Input \%} \\
 & \textbf{($10^{-4}$\,\$)} & \textbf{($10^{-4}$\,\$)}
  & \textbf{($10^{-4}$\,\$)} & \\
\midrule
GPT-5-nano    & 1.5 & 1.1 & 2.6 & 58 \\
Qwen-3.5-9B   & 2.8 & 0.4 & 3.3 & 86 \\
GPT-5.4-nano  & 6.1 & 1.1 & 7.2 & 84 \\
GPT-5-mini    & 7.5 & 3.6 & 11.1 & 68 \\
Kimi-K2.5     & 8.1 & 7.8 & 15.9 & 51 \\
GPT-5.4-mini  & 21.8 & 4.8 & 26.6 & 82 \\
GPT-5.3-chat  & 51.3 & 9.9 & 61.1 & 84 \\
GPT-5.4       & 73.0 & 16.6 & 89.6 & 81 \\
\bottomrule
\end{tabular}
\end{table*}

The outlier is Kimi-K2.5, whose input share is only 51\%---a combination of
its high output rate (\$3.00/M) and its fewer API calls on multi-turn
records (see caveat below). Since output is half of Kimi's cost, its
chattiness of 1.20$\times$ (Section~\ref{sec:chattiness}) retains much of its
raw output ratio (1.50$\times$).
Conversely, Qwen-3.5-9B has the highest raw output ratio (1.72$\times$) but a
chattiness of only 1.06$\times$ because input accounts for 86\% of its cost.

\textbf{Caveat on input token counts.}
The Avg~In column in Table~\ref{tab:token-usage} and the input costs in
Table~\ref{tab:cost-decomp} should be interpreted with two caveats.
First, different models use different tokenizers, so the same prompt text
produces different input token counts---for instance, on queries answered
with a single API call, Kimi-K2.5's tokenizer produces about as many input
tokens as the GPT models (0.99$\times$), while Qwen-3.5-9B's tokenizer
produces 2.0$\times$ more. These differences reflect tokenizer design, not
meaningful variation in model behavior.
Second, on BFCL multi-turn records Kimi-K2.5 makes far fewer API calls than
other models---a median of 2 versus 21 for every other model---because our
harness requires a tool call in every request, so the other models keep
calling tools until the step limit, whereas Kimi-K2.5, whose API does not
enforce this requirement, ends its turns with a text reply
(Appendix~\ref{app:multi-turn-scoring}). Its fewer cumulative calls explain
Kimi's unusually low average input count (1{,}350 tokens vs.\ $\sim$3{,}000
for GPT models) and its low input cost share (51\%).
Despite these caveats, the profiled per-query costs in
Tables~\ref{tab:profiled-cost} and~\ref{tab:cost-decomp} are computed from
each API's own reported token counts and pricing, so they accurately reflect
what a deployment would actually be billed.

\textbf{Chattiness derivation.}
For each model $m$, we compute an \emph{expected cost} assuming it generates the
cross-model average number of output tokens (173.6, averaged over all eight models
on the validation split) while keeping its actual input cost fixed:
\[
  \begin{aligned}
  \text{expected}(m) ={}&
    \underbrace{\bar{t}_{\text{in}}^{(m)} \cdot r_{\text{in}}(m)}_{\text{actual input cost}} \\
    &{}+ \underbrace{\bar{t}_{\text{out}} \cdot r_{\text{out}}(m)}_{\text{avg-output cost}}
  \end{aligned}
\]
where $\bar{t}_{\text{in}}^{(m)}$ is $m$'s average input token count,
$\bar{t}_{\text{out}} = 173.6$ is the cross-model average output token count,
and $r_{\text{in}}, r_{\text{out}}$ are the per-token rates.
\textbf{Chattiness} is the ratio of actual to expected cost:
\[
  \text{chattiness}(m) = \frac{\text{actual cost per query}(m)}
                               {\text{expected cost}(m)}
\]
Figure~\ref{fig:chattiness} visualizes this metric. Points above the $y{=}x$
diagonal cost more than expected (chattiness ${>}1$); points below cost less
(chattiness ${<}1$).

\begin{figure*}[t]
\centering
\begin{tikzpicture}
\begin{axis}[
    width=\linewidth,
    height=7.5cm,
    xlabel={Expected Cost per Query (\$)},
    ylabel={Actual Cost per Query (\$)},
    xmin=0, xmax=0.011,
    ymin=0, ymax=0.011,
    grid=major,
    grid style={gray!20},
    tick label style={font=\scriptsize},
    label style={font=\small},
    clip=false,
    legend style={
        at={(0.03,0.97)},
        anchor=north west,
        font=\scriptsize,
        draw=gray!50,
        fill=white,
        fill opacity=0.9,
        text opacity=1,
    },
]
\addplot[gray, dashed, domain=0:0.011, samples=2] {x};
\addlegendentry{Expected = Actual}

\addplot[only marks, mark=*, mark size=3pt, blue!70!black]
    coordinates {
        (0.009902, 0.008959)  %
        (0.002960, 0.002656)  %
        (0.000822, 0.000718)  %
        (0.007556, 0.006114)  %
        (0.001102, 0.001114)  %
        (0.000218, 0.000258)  %
        (0.001331, 0.001592)  %
        (0.000308, 0.000327)  %
    };
\addlegendentry{Model}

\node[font=\scriptsize, anchor=south east, blue!70!black]
    at (axis cs:0.009802, 0.009112) {GPT-5.4};
\node[font=\scriptsize, anchor=south west, blue!70!black]
    at (axis cs:0.002960, 0.002793) {GPT-5.4-mini};
\node[font=\scriptsize, anchor=south west, blue!70!black]
    at (axis cs:0.007556, 0.006251) {GPT-5.3-chat};

\node[font=\scriptsize, blue!70!black, anchor=west] (lbl-5mini)
    at (axis cs:0.00380, 0.00175) {GPT-5-mini};
\draw[blue!70!black, thin, shorten >=1pt] (lbl-5mini.west) -- (axis cs:0.001102, 0.001114);

\node[font=\scriptsize, blue!70!black, anchor=east] (lbl-kimi)
    at (axis cs:0.00065, 0.00215) {Kimi};
\draw[blue!70!black, thin, shorten >=1pt] (lbl-kimi.east) -- (axis cs:0.001331, 0.001592);

\node[font=\scriptsize, blue!70!black, anchor=west] (lbl-54nano)
    at (axis cs:0.00380, 0.00125) {GPT-5.4-nano};
\draw[blue!70!black, thin, shorten >=1pt] (lbl-54nano.west) -- (axis cs:0.000822, 0.000718);

\node[font=\scriptsize, blue!70!black, anchor=west] (lbl-qwen)
    at (axis cs:0.00380, 0.00075) {Qwen};
\draw[blue!70!black, thin, shorten >=1pt] (lbl-qwen.west) -- (axis cs:0.000308, 0.000327);

\node[font=\scriptsize, blue!70!black, anchor=west] (lbl-5nano)
    at (axis cs:0.00380, 0.00025) {GPT-5-nano};
\draw[blue!70!black, thin, shorten >=1pt] (lbl-5nano.west) -- (axis cs:0.000218, 0.000258);

\end{axis}
\end{tikzpicture}
\caption{Expected vs.\ actual average cost per query on the seed-6 validation split.
\emph{Expected cost} uses each model's actual input cost plus the cost of
generating the cross-model average output (173.6 tokens) at that model's output
rate. Points above the $y{=}x$ diagonal cost more than expected---indicating
above-average output verbosity; points below cost less---indicating concise
output.}
\label{fig:chattiness}
\end{figure*}

\textbf{Verbose budget models.}
Kimi-K2.5 has the highest chattiness in our pool ($\approx$1.20$\times$): it
generates 261 output tokens on average---well above the cross-model mean of
174---making its actual cost 20\% higher than expected. Because output accounts
for 49\% of Kimi's total cost (Table~\ref{tab:cost-decomp}), its verbose
output translates substantially into higher total cost.
Qwen-3.5-9B is the most verbose model in raw token terms (299 avg), yet its
chattiness is only $\approx$1.06$\times$. The explanation is that 86\% of its
per-query cost is input (Table~\ref{tab:cost-decomp}), so even a large output
surplus barely moves the overall cost ratio.
Despite sharing the cheapest input rate with GPT-5-nano (\$0.05/M) and having
an even cheaper output rate (\$0.15/M vs.\ \$0.40/M), Qwen costs 27\% more per
query ($3.3$ vs.\ $2.6 \times 10^{-4}$\,\$) due to its higher total token
consumption---and achieves lower accuracy (64.94\% vs.\ 67.75\% on the test
set).

\textbf{Concise output vs.\ verbose output.}
GPT-5.3-chat has the second-highest output rate in our pool (\$14.00/M), yet it
is the most concise model (chattiness $\approx$0.81$\times$, 71 avg completion
tokens, only 12\% with reasoning). Its actual per-query cost ($61.1 \times 10^{-4}$\,\$) is
19\% below expected. A router relying on list price alone would rank
GPT-5.3-chat among the most expensive models; in practice its frugal output
makes it markedly cheaper than GPT-5.4 while being the most accurate single
model. The GPT-5.4 family (GPT-5.4, GPT-5.4-mini, GPT-5.4-nano) also
falls below the diagonal (chattiness 0.87--0.90$\times$), generating fewer
output tokens than the cross-model average. For these premium models, input
cost dominates (81--84\% of total; Table~\ref{tab:cost-decomp}), which
compresses chattiness toward~1 even though their output token ratios range from
0.52 to 0.64$\times$ the cross-model mean.

\section{KV Caching and Multi-Turn Routing}
\label{app:kv-cache-routing}

We extend the uncached cost analysis in \S\ref{sec:evaluation} with an
offline replay that accounts for key--value (KV) caching. We test
whether retaining a model across turns improves Switchcraft's
accuracy--cost trade-off.

\textbf{Routing variants.}
We compare three policies using the same DistilBERT checkpoints:
\begin{enumerate}
	\item \textbf{Per-turn routing} applies the original policy at each
	turn, including its abstention gate and cost-aware model selection (\S\ref{sec:routing-logic}).
	\item \textbf{Adaptive sticky routing} also evaluates the router each
	turn and preserves its abstention decisions. When routing, it
	retains the current model if its predicted probability of correctness meets
	$\tau_r$; otherwise, it reselects using the original cost-aware
	rule and fallback.
	\item \textbf{First-turn-only routing} applies the original policy only
	on the segment's first turn and keeps that decision for the entire segment.
	It does not perform eligibility or abstention checks in later turns.
\end{enumerate}

\textbf{Multi-turn replay.}
We evaluate all 20 DistilBERT seeds on their respective test partitions
(\S\ref{sec:eval-setup}). Since splitting the data into train/validation/test 
is per turn, intervening training or validation turns break the sequence.
As a result, we restrict replay to \emph{segments} of at least two consecutive turns from the same reconstructed conversation and from the test data.

\textbf{Cache accounting.}
We assume a separate cache for each model per conversation. A cached prompt is
reusable only if it exactly matches a prefix of the current input and the
tool definitions are unchanged. Consider a three-turn conversation routed
to models $A \to B \to A$. On turn~1, $A$ processes and caches the input.
On turn~2, $B$ processes the full input, including the earlier history,
because it cannot reuse $A$'s cache. On turn~3, $A$ can reuse the matching
prefix cached on turn~1, but must process the input added since then.
Switching to $B$ preserves $A$'s cache but does not extend it. We use OpenAI's published model
prices~\cite{openai_model_pricing} and Novita's Kimi-K2.5
prices~\cite{novita_kimi_pricing}
(Table~\ref{tab:kv-cache-input-prices}). For Qwen-3.5-9B, the operators we considered have not published cached-input prices, so we conservatively apply the uncached input rate.

\begin{table}[H]
	\centering
	\caption{Uncached and cached input prices in USD per million input tokens.}
	\label{tab:kv-cache-input-prices}
	\small
	\setlength{\tabcolsep}{4pt}
	\begin{tabular}{@{}lrr@{}}
		\toprule
		\textbf{Model} & \textbf{Uncached input} & \textbf{Cached input} \\
		& \multicolumn{2}{c}{\textbf{(\$/M tokens)}} \\
		\midrule
		GPT-5.4      & 2.50 & 0.250 \\
		GPT-5.4-mini & 0.75 & 0.075 \\
		GPT-5.4-nano & 0.20 & 0.020 \\
		GPT-5.3-chat & 1.75 & 0.175 \\
		GPT-5-mini   & 0.25 & 0.025 \\
		GPT-5-nano   & 0.05 & 0.005 \\
		Kimi-K2.5    & 0.60 & 0.100 \\
		Qwen-3.5-9B  & 0.05 & 0.050 \\
		\bottomrule
	\end{tabular}
\end{table}

\textbf{Results.}
Table~\ref{tab:kv-cache-routing} shows the mean and standard deviation of accuracy and cost across all seeds. The cached and uncached estimates use identical model choices and token counts.

\begin{table}[H]
	\centering
	\caption{Multi-turn KV-cache replay over 20 DistilBERT seeds
		(mean $\pm$ standard deviation).}
	\label{tab:kv-cache-routing}
	\small
	\setlength{\tabcolsep}{3pt}
	\begin{tabular}{@{}lcccc@{}}
		\toprule
		\textbf{Policy} & \textbf{Answer} & \textbf{Routing}
		& \multicolumn{2}{c}{\textbf{Cost ($10^{-4}$\,\$)}} \\
		& \textbf{(\%)} & \textbf{(\%)}
		& \textbf{No cache} & \textbf{With cache} \\
		\midrule
		\shortstack[l]{Per-turn\\routing}
		& \shortstack{73.59\\$\pm$3.41} & \shortstack{85.61\\$\pm$2.08}
		& \shortstack{115.10\\$\pm$94.52} & \shortstack{40.68\\$\pm$18.30} \\
		\addlinespace
		\shortstack[l]{Adaptive sticky\\routing}
		& \shortstack{73.96\\$\pm$3.36} & \shortstack{85.99\\$\pm$2.01}
		& \shortstack{130.00\\$\pm$95.83} & \shortstack{43.64\\$\pm$18.48} \\
		\addlinespace
		\shortstack[l]{First-turn-only\\routing}
		& \shortstack{67.85\\$\pm$3.17} & \shortstack{80.13\\$\pm$1.92}
		& \shortstack{104.11\\$\pm$72.12} & \shortstack{32.88\\$\pm$14.28} \\
		\bottomrule
	\end{tabular}
\end{table}

Caching reduces the estimated cost of
per-turn routing by 64.7\%. Adaptive sticky routing achieves similar mean
answer and routing accuracy but increases cache-adjusted cost by 7.3\%.
Retaining an eligible model avoids switching, yet a cached expensive model
can still cost more than an uncached cheaper one.

First-turn-only routing trades accuracy for lower cost: relative to
per-turn routing, cache-adjusted cost falls by 19.2\%, while answer and
routing accuracy decline by 5.74 and 5.48\,pp, respectively.

Longer sessions may offer greater opportunities for prefix reuse.
We plan to extend the routing cost model to account for cache state.
When the current model remains eligible, switching to another
predicted-correct model would require a lower estimated total cost after
accounting for cache reuse on both models.

\section{Reproduction with an Older Model Basket}
\label{app:old-model-basket}

To demonstrate that our findings generalize beyond a single set of LLMs, we
replicate the core evaluation with an \emph{earlier model basket} comprising
eight OpenAI models available in 2025 (Table~\ref{tab:old-models}).
This basket predates the GPT-5.4 family and the open-weight models used in
the main paper; it includes four reasoning models (GPT-5, GPT-5-mini,
GPT-5-nano, and o4-mini) and spans a 40$\times$ range of input prices, from
GPT-5-nano (\$0.05/M) to GPT-4.1 (\$2.00/M).

\begin{table}[ht]
\centering
\caption{Earlier model basket: LLMs available for routing, listed with
per-token pricing (USD per million tokens).}
\label{tab:old-models}
\small
\begin{tabular}{lrr}
\toprule
\textbf{Model} & \textbf{Input (\$/M)} & \textbf{Output (\$/M)} \\
\midrule
GPT-5          & 1.25  & 10.00 \\
GPT-5-chat     & 1.25  & 10.00 \\
GPT-4.1        & 2.00  & 8.00  \\
o4-mini        & 1.10  & 4.40  \\
GPT-5-mini     & 0.25  & 2.00  \\
GPT-4.1-mini   & 0.40  & 1.60  \\
GPT-5-nano     & 0.05  & 0.40  \\
GPT-4.1-nano   & 0.10  & 0.40  \\
\bottomrule
\end{tabular}
\end{table}

\textbf{Experimental setup.}
This study uses our earlier evaluation protocol
(Appendix~\ref{app:training-config}): the original 14 dataset splits with
Glaive included, 80/10/10 stratified splits, the same DistilBERT
architecture and hyperparameters, the same 20-seed fine-tuning procedure, a
fixed 0.5 threshold without abstention, and an oracle that routes impossible
queries to the most expensive model. Apart from the protocol, the only other
difference from Section~\ref{sec:eval-setup} is the pool of candidate LLMs.
We report results on that protocol's held-out test set of 12{,}282 examples.

\textbf{Results.}
Table~\ref{tab:old-results} and Figure~\ref{fig:old-pareto} present the
accuracy--cost trade-offs. The key findings from the main paper hold:

\begin{table*}[t]
\centering
\caption{Accuracy and average inference cost per query on the test set
(12{,}282 examples) with the earlier model basket. Entities are grouped by
type and sorted by accuracy.}
\label{tab:old-results}
\small
\begin{tabular}{llcc}
\toprule
\textbf{Type} & \textbf{Entity} & \textbf{Accuracy (\%)}
  & \textbf{Avg Cost ($10^{-4}$\,\$)} \\
\midrule
\multirow{8}{*}{Single LLM}
  & GPT-4.1        & 83.22 & 73.6 \\
  & GPT-4.1-mini   & 82.18 & 10.5 \\
  & GPT-5          & 79.18 & 49.9 \\
  & GPT-5-nano     & 78.80 & 1.8 \\
  & GPT-5-mini     & 77.24 & 8.0 \\
  & GPT-4.1-nano   & 76.67 & 3.2 \\
  & GPT-5-chat     & 62.76 & 22.7 \\
  & o4-mini        & 60.93 & 23.6 \\
\midrule
\multirow{3}{*}{Heuristic Router}
  & Length          & 78.95 & 12.8 \\
  & Num.\ Tools   & 67.84 & 66.9 \\
  & Num.\ Turns    & 62.67 & 54.0 \\
\midrule
Chat Router
  & DistilBERT (chat-oriented) & 77.47 & 7.7 \\
\midrule
\textbf{Agent Router}
  & \textbf{DistilBERT (ours)} & \textbf{82.73} & \textbf{8.7} \\
\midrule
Upper Bound
  & Oracle         & 89.65 & 12.7 \\
\bottomrule
\end{tabular}
\end{table*}

\begin{figure*}[t]
\centering
\includegraphics[width=\linewidth]{graphs/old-models-tradeoffs.png}
\caption{Accuracy--cost Pareto plot for the earlier model basket on the
held-out test set (12{,}282 examples). Our agent-fine-tuned DistilBERT router
(\textcolor{red!70!black}{$\blacksquare$}) achieves 82.73\% accuracy while the
earlier chat-oriented router (\textcolor{green!50!black}{$\blacklozenge$}),
trained mainly on chat-completion benchmarks, reaches only 77.47\% at similar
cost.}
\label{fig:old-pareto}
\end{figure*}

\textbf{Finding 1: the learned router dominates the cost--accuracy frontier.}
The DistilBERT agent router achieves 82.73\% accuracy at $8.7 \times 10^{-4}$\,\$ per
query---comparable to the best individual model (GPT-4.1 at 83.22\%) while
reducing cost by \textbf{88\%} ($73.6 \rightarrow 8.7 \times 10^{-4}$\,\$). This
mirrors the main result in Section~\ref{sec:main-results}: a lightweight
classifier can match frontier-model accuracy at a fraction of the cost.

\textbf{Finding 2: a chat-oriented router is insufficient for agentic
workloads.}
This experiment additionally includes a \emph{Chat Router}: an earlier
DistilBERT router for the same eight models, trained on the 64 public benchmarks in
Table~\ref{tab:chat-router-data}. All but one are chat-completion tasks
(e.g., question answering, reasoning, math, code, and summarization); the
exception, BFCL, is a function-calling benchmark. This router does not have our agentic training
data, AST-based correctness labels, or token packing. Despite achieving
low cost ($7.7 \times 10^{-4}$\,\$), the chat router reaches only \textbf{77.47\%}
accuracy---5.26 percentage points below our agent router (82.73\%) and even
below the cheapest heuristic baseline (Length at 78.95\%).

This gap suggests that routing learned mainly from chat completions transfers
poorly to agentic tool calling, whose queries involve multi-turn conversations,
complex tool schemas, and interleaved reasoning that differ from single-turn
chat. Since the chat router shares our encoder and model pool, the gap reflects
how it was trained---its data, correctness labels, and input
representation---rather than model capacity.

\textbf{Finding 3: heuristic performance varies with the model pool.}
With the earlier model basket, the Length heuristic performs best (78.95\%),
while Num.\ Turns drops to 62.67\%---a reversal from the main results where
Num.\ Turns leads (76.13\%; Table~\ref{tab:main-results}). This suggests that heuristic effectiveness is
tightly coupled to the specific models available and does not generalize,
whereas the learned router consistently performs well regardless of the
underlying model pool.

\begin{table*}[t]
\centering
\caption{The 64 public datasets on which the earlier chat-oriented router was
trained; all but BFCL (function calling) are chat-completion
benchmarks. HF = Hugging Face, GH = GitHub.}
\label{tab:chat-router-data}
\small
\setlength{\tabcolsep}{2pt}
\begin{tabular}{@{}ll@{\hspace{10pt}}c@{\hspace{10pt}}ll@{}}
\toprule
\textbf{Dataset} & \textbf{Source} & & \textbf{Dataset} & \textbf{Source} \\
\midrule
AGIEval              & GH (ruixiangcui/AGIEval)          & & mgsm                 & GH (google-research/url-nlp) \\
ai2\_arc             & HF (allenai/ai2\_arc)             & & mlqa                 & HF (facebook/mlqa) \\
anli                 & HF (facebook/anli)                & & mmlu                 & HF (cais/mmlu) \\
Belebele             & HF (facebook/2M-Belebele)         & & MMLU-Pro             & HF (TIGER-Lab/MMLU-Pro) \\
bigcodebench         & HF (bigcode/bigcodebench)         & & narrativeqa          & HF (deepmind/narrativeqa) \\
BioInstructQA        & HF (BioMistral/BioInstructQA)     & & Natural-Questions    & GH (google-research-datasets) \\
boolq                & HF (google/boolq)                 & & OpenBookQA           & GH (allenai/OpenBookQA) \\
BoolQ\_robustness    & HF (ibm-research/BoolQ\_rob.)     & & pawsx                & GH (google-research-datasets/paws) \\
cnn\_dailymail       & HF (abisee/cnn\_dailymail)        & & piqa                 & GH (ybisk/ybisk.github.io) \\
code\_generation     & HF (livecodebench/code\_gen.)     & & PubMedQA             & HF (qiaojin/PubMedQA) \\
code\_gen.\_lite     & HF (livecodebench/code\_gen.\_l.) & & qasper               & HF (allenai/qasper) \\
commonsense\_qa      & HF (tau/commonsense\_qa)          & & QMSum                & GH (Yale-LILY/QMSum) \\
CyberMetric          & GH (cybermetric/CyberMetric)      & & quac                 & HF (allenai/quac) \\
d2n                  & HF (NTaylor/d2n)                  & & race                 & HF (ehovy/race) \\
function-calling     & HF (gorilla-llm/BFCL)             & & RCT-summarization    & GH (bwallace/RCT-summ.-data) \\
gpqa                 & GH (Idavidrein/gpqa)              & & samsum               & HF (Samsung/samsum) \\
gpt4\_judge\_battles & HF (routellm/gpt4\_judge\_b.)     & & social\_i\_qa        & HF (allenai/social\_i\_qa) \\
gsm8k                & HF (openai/gsm8k)                 & & squad\_v2            & HF (rajpurkar/squad\_v2) \\
HeadQA               & HF (dvilares/head\_qa)            & & SQuALITY             & GH (nyu-mll/SQuALITY) \\
HellaSwag            & GH (rowanz/hellaswag)             & & StoryCloze           & GH (EleutherAI/lm-eval.-harness) \\
HumanEval            & GH (openai/human-eval)            & & SummScreen           & GH (mingdachen/SummScreen) \\
humanevalplus        & HF (evalplus/humanevalplus)       & & ToxiGen              & HF (toxigen/toxigen-data) \\
lambada              & HF (cimec/lambada)                & & trivia\_qa           & HF (mandarjoshi/trivia\_qa) \\
MATH-500             & HF (HuggingFaceH4/MATH-500)      & & TruthfulQA           & GH (sylinrl/TruthfulQA) \\
MathInstruct         & HF (TIGER-Lab/MathInstruct)       & & TyDiQA               & GH (google-research-datasets/tydiqa) \\
mbpp                 & HF (Muennighoff/mbpp)             & & WinoGrande           & GH (allenai/winogrande) \\
mbppplus             & HF (evalplus/mbppplus)            & & wmdp                 & HF (cais/wmdp) \\
med\_qa              & HF (bigbio/med\_qa)               & & xcopa                & HF (cambridgeltl/xcopa) \\
MedCalc-Bench        & GH (ncbi-nlp/MedCalc-Bench)       & & xlsum                & HF (csebuetnlp/xlsum) \\
medbullets           & GH (HanjieChen/ChallengeClQA)     & & xnli                 & HF (facebook/xnli) \\
medical-o1-ver.      & HF (FreedomIntelligence)          & & xquad                & HF (google/xquad) \\
medmcqa              & HF (openlifescienceai/medmcqa)    & & xstory\_cloze        & HF (juletxara/xstory\_cloze) \\
\bottomrule
\end{tabular}
\end{table*}

\section{Impact of Newer Models}
\label{app:newer-model-basket}

Two models newer than our basket, GPT-6-Sol and GPT-6-Luna (\$2.00/\$10.00 and
\$0.10/\$0.50 per million input/output tokens), raise the question of whether
Switchcraft remains useful. Without fine-tuning it again or retuning its
policy, we evaluate both models on the same queries with the same AST
framework and cost model, and compare them as single-model baselines with
Switchcraft (seed~6) on the test set and on held-out Glaive.

A few long multi-turn queries lack token counts for some original models because
those models failed early, while the newer models complete them at high cost. We
compare costs on the queries where both have token counts, which favors the new
GPT-6 models.

\begin{table*}[t]
\centering
\caption{Switchcraft (seed~6), fine-tuned on the original basket only, versus
two newer models as single-model baselines, with the best original model for
reference. Costs marked $\dagger$ are over the test queries where both the
newer model and Switchcraft's chosen model have token counts (all but 19--20);
Switchcraft's own cost is also 27.5 on each of these subsets.}
\label{tab:newer-models}
\small
\begin{tabular}{lcccc}
\toprule
\textbf{Method} & \textbf{Answer} & \textbf{Routing}
  & \textbf{Decisive} & \textbf{Cost} \\
 & \textbf{(\%)} & \textbf{(\%)} & \textbf{(\%)}
  & \textbf{($10^{-4}$\,\$)} \\
\midrule
\multicolumn{5}{l}{\emph{In-distribution test (9{,}746 queries)}} \\
Switchcraft  & 78.27 & 86.40 & 87.39 & 27.5 \\
GPT-6-Sol    & 71.84 & 71.84 & 76.26 & 82.8\rlap{$^\dagger$} \\
GPT-6-Luna   & 73.72 & 73.72 & 79.37 & 4.2\rlap{$^\dagger$} \\
GPT-5.3-chat & 75.63 & 75.63 & 82.87 & 66.2 \\
\midrule
\multicolumn{5}{l}{\emph{Held-out Glaive (83{,}814 queries)}} \\
Switchcraft  & 89.20 & 89.39 & 79.32 & 1.46 \\
GPT-6-Sol    & 84.55 & 84.55 & 57.55 & 6.96 \\
GPT-6-Luna   & 85.37 & 85.37 & 61.07 & 0.35 \\
GPT-5.4      & 88.53 & 88.53 & 75.64 & 10.27 \\
\bottomrule
\end{tabular}
\end{table*}

\textbf{Switchcraft remains more accurate than both.}
Newer is not uniformly better (Table~\ref{tab:newer-models}): GPT-6-Sol is
less accurate than GPT-5.3-chat, and Switchcraft dominates it, with 6.4\,pp
higher answer accuracy (95\% paired-bootstrap interval 5.7--7.2\,pp) at 67\%
lower cost, and 4.6\,pp higher at 79\% lower cost on Glaive. Every operating
point in Table~\ref{tab:budget-sweep} from a budget of 5 upward is also more
accurate and cheaper than GPT-6-Sol. GPT-6-Luna is a trade-off: Switchcraft is
4.5\,pp more accurate (interval 3.8--5.3\,pp; 12.7\,pp in routing accuracy) but
6.6$\times$ costlier ($27.48$ vs.\ $4.17 \times 10^{-4}$\,\$ on the 9{,}727
queries where both have token counts), and on Glaive 3.8\,pp more accurate at
4.2$\times$ the cost. No operating point beats GPT-6-Luna on both answer
accuracy and cost. The cheapest more accurate one, at a budget of 8 (75.0\%),
costs 2.7 times as much ($12.61$ vs.\ $4.60 \times 10^{-4}$\,\$ on the 9{,}727
queries where both have token counts; this subset excludes different queries,
which raises GPT-6-Luna's mean). The lowest-budget point beats it on routing
accuracy (78.9\% vs.\ 73.7\%) at lower cost ($3.51$ vs.\
$4.08 \times 10^{-4}$\,\$ on 9{,}728 such queries). Both accuracy margins hold on all 20 seeds (5.3--6.6\,pp over GPT-6-Sol,
3.5--5.3\,pp over GPT-6-Luna).

\textbf{Newer models enlarge the routing opportunity.}
Adding both models to the basket raises the oracle's answer accuracy only from
85.71\% to 86.29\%: each answers under 0.5\% of test queries that no original
model answers, and on 98.7\% of the queries GPT-6-Sol answers correctly, a
cheaper original model does too. GPT-6-Luna, however, is the cheapest correct
model on 31.7\% of queries, lowering the oracle's cost from 4.0 to
$2.9 \times 10^{-4}$\,\$. The newer models thus add little capability, but
GPT-6-Luna is a cheaper way to answer many queries the basket already
handles---the opportunity a router exploits---so we expect re-fine-tuning
Switchcraft with GPT-6-Luna in its basket to lower its cost further. Its
abstentions also remain sound: the newer models answer only 37--43 of its 878
abstentions, and crediting an abstention only when they also fail lowers its
routing accuracy by just 0.2\,pp. Since the newer models were likely trained
on more recent data, memorization of our public benchmarks
(Section~\ref{sec:discussion}) would, if anything, favor them.

\section{Probabilistic Correctness Labels}
\label{app:probabilistic}

\subsection{Motivation}

The standard evaluation protocol in this paper treats each LLM response as
either correct or incorrect (a binary label). In practice, however, LLMs are
\emph{stochastic}: the same model may produce a correct answer on some
invocations and an incorrect one on others for the \emph{same} query.
A model that answers correctly 95\% of the time is fundamentally more reliable
than one that answers correctly 55\% of the time, yet both receive the same
binary label of ``correct'' if evaluated on a single invocation.

This appendix presents a preliminary investigation into \textbf{probabilistic
correctness labels}: instead of evaluating each model once per query, we
evaluate it \textbf{20 times} and record the fraction of correct responses as
a probability $p \in [0, 1]$. We then fine-tune Switchcraft using these soft labels
and measure whether this richer supervision improves routing quality.

\subsection{Methodology}

\textbf{Multi-invocation evaluation.}
Each of the eight candidate LLMs is called 20 times on every query.
Each response is independently scored using the same AST comparison framework
described in Section~\ref{sec:eval-setup}. The \emph{correctness probability}
for model $m$ on query $q$ is:
\[
  p(m, q) = \frac{\text{correct iterations}}{20}
\]
This probability captures the reliability of a model on a given query, not
just whether it \emph{can} answer correctly.

\textbf{Dataset coverage.}
Due to the 20$\times$ cost multiplier, we evaluated 12 of the 14 dataset splits
(omitting Glaive and xLAM-60K), yielding 1{,}224 test examples.
This is a smaller test set than the 12{,}282 examples of the earlier protocol,
making results noisier but still directionally informative.

\textbf{Soft-label training.}
The
data generation pipeline stores each model's probability directly as the
training label. For example, a query where GPT-4.1 answered correctly on 17/20
invocations and GPT-5-nano on 20/20 receives the label vector:
\begin{center}
\small
\texttt{\{``gpt-4.1'': 0.85, ``gpt-5-nano'': 1.0, ...\}}
\end{center}
The DistilBERT classifier is then fine-tuned with \texttt{BCEWithLogitsLoss}
against these continuous targets. Each output head learns to predict the
\emph{probability of correctness} for the corresponding model, rather than a
hard correct/incorrect label.

\textbf{Inference procedure.}
At inference time, the soft-label fine-tuned model still outputs a sigmoid
probability for each candidate LLM. The prediction procedure is identical to
the binary-label case under our earlier protocol
(Appendix~\ref{app:training-config}):
\begin{enumerate}
  \item Apply a threshold ($\theta = 0.5$) to each sigmoid output to determine
        which models are predicted to be ``reliable'' for this query.
  \item Among the models above the threshold, select the \textbf{cheapest} one
        (cost-aware tie-breaking using profiled per-query costs).
  \item If no model exceeds the threshold, fall back to the model with the
        highest predicted probability (argmax).
\end{enumerate}

\begin{figure*}[t]
\centering
\includegraphics[width=\linewidth,trim=0 280bp 0 38bp,clip]{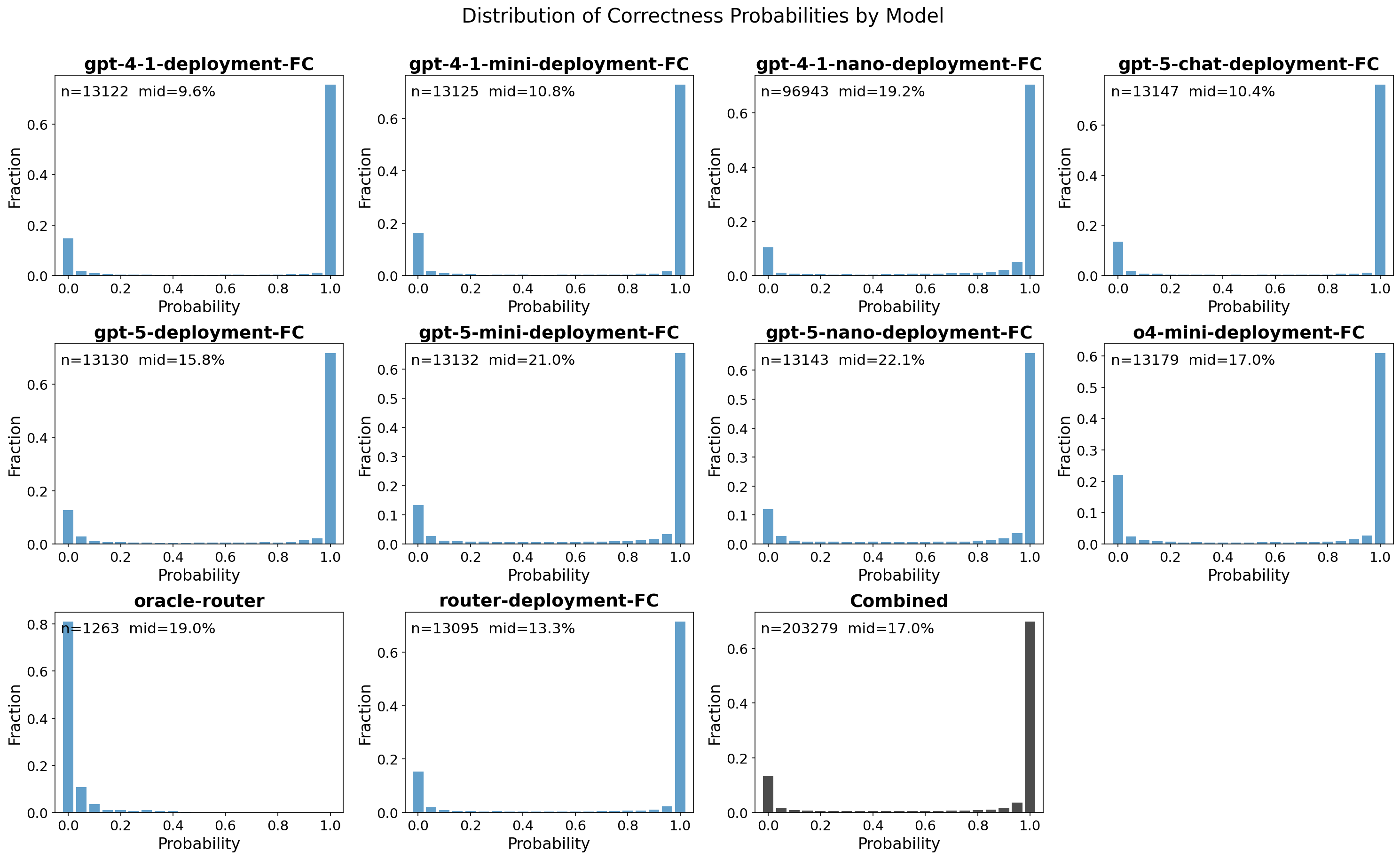}
\caption{Distribution of correctness probabilities across 20 invocations,
grouped by model. Each histogram shows the fraction of queries at each
probability level. ``mid'' denotes the fraction of queries with probability
strictly between 0 and 1.}
\label{fig:prob-by-model}
\end{figure*}

\begin{figure*}[t]
\centering
\includegraphics[width=\linewidth,trim=0 0 0 38bp,clip]{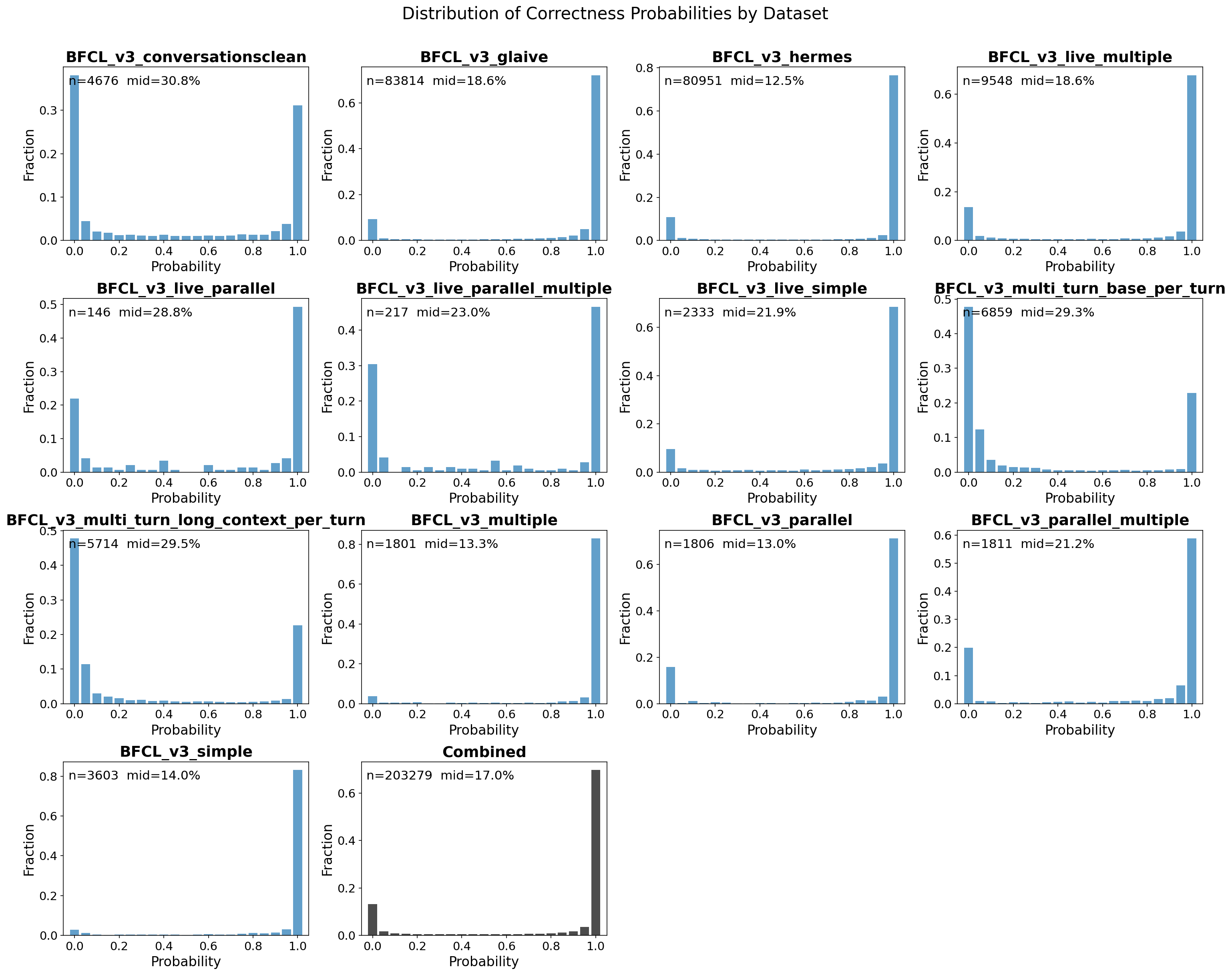}
\caption{Distribution of correctness probabilities across 20 invocations,
grouped by dataset. Multi-turn and conversational datasets show higher
fractions of intermediate probabilities, indicating greater response
variability.}
\label{fig:prob-by-dataset}
\end{figure*}

\textbf{Probability distributions.}
Figure~\ref{fig:prob-by-model} shows the distribution of correctness
probabilities per model. The distributions are strongly bimodal: the vast
majority of model--query pairs have probability near 0 (always incorrect) or
1 (always correct), with only 10--22\% falling in the intermediate range
(denoted ``mid'' in each subplot). This bimodality explains why soft labels do
not dramatically change the learning problem---most labels are effectively
binary. Figure~\ref{fig:prob-by-dataset} shows the same analysis grouped by
dataset; multi-turn datasets (ConFETTI, multi-turn base/long-context) exhibit
higher mid-range fractions (29--31\%), indicating greater stochasticity in
model responses for complex conversational tasks.

\subsection{Results}

Table~\ref{tab:prob-results} and Figure~\ref{fig:prob-pareto} present the
results. The model basket and evaluation protocol are the same as in
Appendix~\ref{app:old-model-basket} (the earlier basket of eight OpenAI
models).

\begin{table*}[t]
\centering
\caption{Accuracy and average inference cost per query on the test set
(1{,}224 examples) with probabilistic correctness labels. Entities are grouped
by type and sorted by accuracy.}
\label{tab:prob-results}
\small
\begin{tabular}{llcc}
\toprule
\textbf{Type} & \textbf{Entity} & \textbf{Accuracy (\%)}
  & \textbf{Avg Cost ($10^{-4}$\,\$)} \\
\midrule
\multirow{8}{*}{Single LLM}
  & GPT-5-chat     & 82.11 & 240 \\
  & GPT-4.1        & 80.31 & 506 \\
  & GPT-5          & 79.98 & 329 \\
  & GPT-5-nano     & 78.84 & 11 \\
  & GPT-4.1-mini   & 78.10 & 105 \\
  & GPT-5-mini     & 77.61 & 65 \\
  & GPT-4.1-nano   & 74.67 & 25 \\
  & o4-mini        & 73.69 & 195 \\
\midrule
\multirow{3}{*}{Heuristic Router}
  & Num.\ Tools     & 80.96 & 430 \\
  & Length           & 77.94 & 84 \\
  & Num.\ Turns     & 74.35 & 338 \\
\midrule
Chat Router
  & DistilBERT (chat-oriented) & 79.17 & 118 \\
\midrule
\textbf{Agent Router}
  & \textbf{DistilBERT (ours)} & \textbf{82.28} & \textbf{94} \\
\midrule
Upper Bound
  & Oracle         & 91.01 & 205 \\
\bottomrule
\end{tabular}
\end{table*}

\begin{figure*}[t]
\centering
\includegraphics[width=\linewidth]{graphs/probabilistic-scoring-tradeoffs.png}
\caption{Accuracy--cost Pareto plot with probabilistic correctness labels
(1{,}224 test examples). The agent router fine-tuned on soft probability labels
(\textcolor{red!70!black}{$\blacksquare$}, with seed-range error bars) achieves
82.28\% accuracy.}
\label{fig:prob-pareto}
\end{figure*}

\subsection{Discussion}

\textbf{Soft labels match binary-label performance.}
The probabilistic router achieves \textbf{82.28\%} accuracy at $94 \times 10^{-4}$\,\$ per
query. For comparison, the binary-label router on the same model basket
(Appendix~\ref{app:old-model-basket}), evaluated on all 14 dataset splits
(12{,}282 test examples), achieved 82.73\% at $8.7 \times 10^{-4}$\,\$.
However, this is not a direct comparison: the probabilistic experiment covers
only 12 dataset splits (1{,}224 test examples) due to the prohibitive cost of
20$\times$ evaluation. The accuracy difference (0.45~pp) is within noise for
the smaller test set, and the higher per-query cost ($94 \times 10^{-4}$\,\$ vs.\ $8.7 \times 10^{-4}$\,\$)
reflects the different dataset composition rather than the labeling strategy
itself---the two omitted datasets (Glaive, xLAM-60K) contain many easy queries
that drive down average cost in the full evaluation.

\textbf{The prediction task is not substantially harder.}
Our initial hypothesis was that probabilistic labels would make the prediction
task \emph{harder} for the router---since a model might be correct only 60\%
of the time, the router must learn a finer-grained distinction than
correct/incorrect. In practice, however, the accuracy distributions are
strongly bimodal: most model--query pairs have probability near 0 or near 1,
with relatively few in the intermediate range. The mean correctness probability
across all 203{,}279 model--query evaluations is 0.80, and the median is 1.0,
indicating that the majority of entries are deterministic.

\textbf{Cost implications.}
The 20$\times$ evaluation cost is prohibitive at scale: evaluating 8 models
$\times$ 12{,}282 queries $\times$ 20 iterations would require nearly
2 million API calls. We were only able to complete 12 of the 14 dataset splits within
our budget, resulting in a smaller test set (1{,}224 examples). The reduced
test set makes it difficult to draw definitive conclusions about whether soft
labels improve routing quality.

\textbf{Implications and future work.}
While this preliminary investigation does not demonstrate a clear benefit of
probabilistic labels over binary labels, the approach remains promising for
settings where:
\begin{itemize}
  \item Model responses are highly stochastic (e.g., complex multi-step
        reasoning tasks where correctness varies significantly across runs).
  \item The cost of incorrect routing is high, and a confidence-calibrated
        router that can say ``this model answers correctly only 60\% of the
        time'' would enable risk-aware fallback strategies.
  \item New evaluation data is expensive to acquire, and soft labels extract
        more information per evaluation run.
\end{itemize}
A larger-scale investigation with the full dataset and a proper cost-controlled
comparison is an important direction for future work.

\section{MIRT-DistilBERT Baseline}
\label{app:mirt}

To assess whether an alternative routing architecture that explicitly models
per-LLM capabilities could outperform our multi-label classification approach,
we implement and evaluate a router based on \textbf{Multidimensional Item
Response Theory} (MIRT)~\cite{irt_router}: the MIRT baseline of
Table~\ref{tab:main-results} and Figure~\ref{fig:pareto}. MIRT originates in
psychometrics, where it models the probability that a test-taker with latent
ability $\boldsymbol{\theta}$ answers an item with discrimination
$\boldsymbol{a}$ and difficulty $b$ correctly. Applied to LLM routing, each
candidate model plays the role of a test-taker and each query plays the role
of a test item. The MIRT router uses the same 14 in-distribution dataset
splits, seed-dependent partitions, 20 seeds, eight candidate models, and
profiled costs as Switchcraft (Appendix~\ref{app:training-config}), and we
evaluate it on the same test set and on held-out Glaive.

\subsection{Architecture}

The MIRT router uses a two-stage architecture:

\textbf{Stage 1: embedding extraction (frozen).}
Both query texts and LLM profile descriptions are encoded by a frozen
DistilBERT model into 768-dimensional mean-pooled, L2-normalized embeddings.
Each LLM is represented by a short natural-language profile of two or three
sentences, adapted from its provider's documentation, that describes its
capabilities and intended workloads (Table~\ref{tab:mirt-profiles}). A query is
embedded from its conversation text followed by its tool signatures (function
and parameter names), truncated to DistilBERT's 512-token limit.

\textbf{Stage 2: MIRT head (trained from scratch).}
A lightweight MIRT head projects query and LLM embeddings into a shared
$K$-dimensional latent space (we use $K{=}25$) via three bias-free linear
projections:
\begin{align}
  \boldsymbol{\theta} &= W_\theta \, \mathbf{e}_{\mathrm{llm}} \\
  \boldsymbol{a}      &= \mathrm{softplus}\!\bigl(W_a \, \mathbf{e}_{\mathrm{query}}\bigr) \\
  b                    &= W_b \, \mathbf{e}_{\mathrm{query}}
\end{align}
Here, $\boldsymbol{\theta}$ is LLM ability, $\boldsymbol{a}$ is query
discrimination, and $b$ is query difficulty.
The predicted probability that LLM $m$ answers query $q$ correctly follows
the 2PL (two-parameter logistic) IRT response function:
\[
  P(m, q) = \sigma\!\Bigl(\sum_{k=1}^{K} a_k \, \theta_k - b\Bigr)
\]
At inference time, the router scores all eight LLMs for a given query and
selects the one with the lowest profiled cost among those whose predicted
probability is at least 0.5, falling back to the highest-scoring LLM if none
qualifies. This is the cost-aware rule of our earlier evaluation protocol
(Appendix~\ref{app:training-config}). Unlike Switchcraft's policy
(Section~\ref{sec:routing-logic}), it calibrates nothing, tunes no threshold
on validation data, and never abstains, so MIRT's routing accuracy equals
its answer accuracy.

\begin{table*}[t]
\centering
\caption{Profiles that MIRT embeds to represent the candidate LLMs (exact
text). Each is adapted from the provider's model documentation: OpenAI's API
model pages for the GPT models and the Hugging Face model cards for Kimi-K2.5
and Qwen-3.5-9B.}
\label{tab:mirt-profiles}
\small
\begin{tabular}{lp{10.8cm}}
\toprule
\textbf{Model} & \textbf{Profile} \\
\midrule
GPT-5.3-chat & GPT-5.3 Chat is OpenAI's conversational GPT-5.3 Instant model
  used in ChatGPT. It accepts text and image inputs and produces text responses.
  It supports streaming, structured outputs, and function calling for
  interactive conversational applications. \\
\addlinespace
GPT-5.4 & GPT-5.4 is an OpenAI model for demanding professional and coding
  tasks. It supports adjustable reasoning effort, long-context text and image
  inputs, structured outputs, and function calling. Its tool capabilities
  include search, code execution, and computer use for multi-step agent
  workflows. \\
\addlinespace
GPT-5.4-mini & GPT-5.4 mini is an efficient OpenAI model for coding, computer
  use, and sub-agent workloads. It provides a lower-cost alternative to GPT-5.4
  for high-volume applications, with adjustable reasoning effort, text and image
  inputs, structured outputs, and function calling. \\
\addlinespace
GPT-5-mini & GPT-5 mini is a smaller OpenAI reasoning model suited to clearly
  specified tasks and precise prompts. It balances response latency and cost for
  high-volume workloads. It accepts text and image inputs and supports
  structured outputs and function calling. \\
\addlinespace
GPT-5.4-nano & GPT-5.4 nano is a small OpenAI model focused on inexpensive,
  fast processing of high-volume tasks. Its intended workloads include
  classification, extraction, ranking, and sub-agent work. It supports
  adjustable reasoning effort, text and image inputs, structured outputs, and
  function calling. \\
\addlinespace
GPT-5-nano & GPT-5 nano is the smallest, fastest, and least expensive model in
  OpenAI's GPT-5 family. It is suited to summarization and classification
  workloads where throughput and cost are important. It supports reasoning, text
  and image inputs, structured outputs, and function calling. \\
\addlinespace
Qwen-3.5-9B & Qwen3.5-9B is the Qwen team's post-trained model with a
  9-billion-parameter language backbone and a vision encoder. It supports
  multilingual text and visual understanding, reasoning, coding, and agent
  tasks. Its hybrid linear and full attention architecture targets efficient
  inference with long-context inputs. \\
\addlinespace
Kimi-K2.5 & Kimi K2.5 is Moonshot AI's multimodal mixture-of-experts model
  with one trillion total parameters and 32 billion active parameters. It
  combines language and visual understanding with coding and agentic tool use.
  It provides instant and thinking modes and supports complex, multi-step agent
  workflows. \\
\bottomrule
\end{tabular}
\end{table*}

\subsection{Training Differences from Our Multi-Label Router}

Each training query becomes eight (query, LLM, correct) pairs, 623{,}128 in
total, and only the MIRT head is trained, with binary cross-entropy, for 10
epochs. After each epoch we measure pairwise accuracy (whether the predicted
probability clears 0.5 exactly when the LLM is correct) on the validation
split and restore the best epoch; every seed peaks in one of the last two
epochs. As for Switchcraft, we report the seed with the highest validation
answer accuracy: seed~6, also DistilBERT's, so both routers are tested on the
same 9{,}746 queries. With the encoder frozen, a seed takes about 19 minutes
on one GPU, including embedding the queries. Table~\ref{tab:mirt-diff}
summarizes the key differences between the MIRT router and our DistilBERT
multi-label classifier.

\begin{table*}[t]
\centering
\caption{Architectural and training differences between our multi-label
DistilBERT router and the MIRT-DistilBERT baseline.}
\label{tab:mirt-diff}
\small
\begin{tabular}{lll}
\toprule
\textbf{Aspect} & \textbf{Our router} & \textbf{MIRT} \\
\midrule
Encoder & Fine-tuned end-to-end & Frozen (embeddings only) \\
Head & Single 8-logit classifier & Shared IRT 2PL head \\
Training data format & Multi-label per query & Pairwise (query, LLM, 0/1) \\
Trainable parameters & 67M (full router) & 39K (head only) \\
LLM representation & Fixed output classes & Embedded profile text \\
Input representation & Compressed token packing & Vanilla concatenation \\
Optimizer & AdamW & Adam \\
Learning rate & $5 \times 10^{-5}$ & $10^{-3}$ \\
Weight decay & 0.01 & 0.0 \\
Batch size & 64 (16 per GPU) & 16 pairs \\
Epochs & Up to 30, early stopping & 10 \\
Checkpoint metric & Decisive accuracy at budget & Pairwise accuracy \\
Decision rule & Calibrated, tuned, abstains & Fixed 0.5 threshold \\
\bottomrule
\end{tabular}
\end{table*}

The MIRT architecture has two potential advantages over fixed-head
classification: (i)~it can theoretically generalize to \emph{new} LLMs at
test time by computing an embedding from their profile text, without
retraining; and (ii)~its \emph{trainable} parameter count is orders of
magnitude smaller (39{,}168 vs.\ 67M), since only the three projection
matrices are learned while the encoder remains frozen. Note, however, that
the frozen encoder is still required at inference time, so the total model
size is comparable; the advantage is reduced training cost, not a
smaller deployment footprint.

\subsection{Results}

Table~\ref{tab:mirt-results} compares MIRT with Switchcraft and with kNN
($k{=}1$), the baseline closest to it in accuracy; Tables~\ref{tab:main-results}
and~\ref{tab:glaive-full} list every method.

\begin{table*}[t]
\centering
\caption{MIRT-DistilBERT versus Switchcraft (DistilBERT) and kNN ($k{=}1$)
on the in-distribution test set (seed-6 partition, 9{,}746 queries) and
held-out Glaive (83{,}814 queries): answer and decisive accuracy (\%) and
cost ($10^{-4}$\,\$). Best per column in \textbf{bold}.}
\label{tab:mirt-results}
\small
\setlength{\tabcolsep}{4pt}
\begin{tabular}{l ccc ccc}
\toprule
& \multicolumn{3}{c}{\textbf{In-distribution test}}
  & \multicolumn{3}{c}{\textbf{Held-out Glaive}} \\
\cmidrule(lr){2-4} \cmidrule(lr){5-7}
\textbf{Router} & \textbf{Answer} & \textbf{Decisive} & \textbf{Cost}
  & \textbf{Answer} & \textbf{Decisive} & \textbf{Cost} \\
\midrule
Switchcraft (DistilBERT) & \textbf{78.27} & \textbf{87.39} & 27.5
  & \textbf{89.20} & \textbf{79.32} & 1.46 \\
kNN ($k{=}1$)            & 70.89 & 74.82 & \textbf{4.7}
  & 87.24 & 68.90 & \textbf{0.46} \\
MIRT-DistilBERT          & 70.68 & 74.45 & 11.6
  & 87.05 & 67.87 & 0.48 \\
\bottomrule
\end{tabular}
\end{table*}

In distribution, MIRT is 7.59\,pp less accurate than Switchcraft
(12.94\,pp on decisive queries) at 58\% lower cost, and two points dominate
it: kNN ($k{=}1$) is slightly more accurate at 2.5$\times$ lower cost, and
Switchcraft's budget-5 operating point reaches 72.06\% at
$7.9 \times 10^{-4}$\,\$ (Table~\ref{tab:budget-sweep}). The gap does not
depend on the seed: MIRT's test answer accuracy spans 70.07--71.88\% across
its 20 seeds, below the 77.33--78.99\% of Switchcraft's
(Appendix~\ref{app:seed-stability}). On held-out Glaive, MIRT trails
Switchcraft by 2.15\,pp in answer accuracy and 11.45\,pp on decisive queries
at a third of its cost, again just below kNN ($k{=}1$). Its deficit is
largest on FunReason-MT (52.1\% vs.\ 67.8\%) and BFCL multi-turn (58.7\% vs.\
72.7\%; Table~\ref{tab:per-dataset}).

\subsection{Analysis}

\textbf{Why does MIRT underperform?}
We consider three factors:

\begin{enumerate}
  \item \textbf{Untuned decision rule.} The fixed 0.5 threshold sends a
        query to the cheapest LLM that MIRT rates more likely right than
        wrong, and nothing tunes it for accuracy or cost; unable to abstain,
        MIRT also pays for a wrong answer on every impossible query. Its
        scores hold more signal than this rule extracts: routing each query
        to its top-scoring LLM, ignoring cost, raises seed~6's test answer
        accuracy from 70.68\% to 77.49\%, above every single model and kNN
        ($k{=}10$). Switchcraft's most accurate operating point, which
        likewise ignores cost and sends every query it does not abstain on
        to its most-likely-correct model, reaches 79.75\%
        (Table~\ref{tab:budget-sweep}), 2.26\,pp higher.

  \item \textbf{Frozen encoder.} Our multi-label router fine-tunes the full
        67M-parameter DistilBERT-based router end-to-end, allowing the encoder to learn
        task-specific representations for agentic function-calling queries.
        The MIRT router uses frozen embeddings from a general-purpose
        pre-trained model, which may not capture the fine-grained
        distinctions (e.g., JSON structure validity, tool-schema compliance)
        that matter for routing.

  \item \textbf{Vanilla input (likely minor).} MIRT embeds the conversation
        followed by the tool signatures and truncates at 512 tokens, so long
        inputs lose their tool signatures and then their latest
        turns---the content our token packing keeps first
        (Section~\ref{sec:input-representation}). Most inputs fit, however,
        and fine-tuning our own router on this input without packing lowers
        its answer accuracy by only 0.50\,pp on average over 20 seeds
        (Appendix~\ref{app:ablation-packing}), so the input likely explains
        little of MIRT's 7.59\,pp gap.
\end{enumerate}

\textbf{MIRT's advantage: extensibility.}
The MIRT architecture represents LLMs via their profile embeddings rather
than fixed output classes, which in principle allows zero-shot routing to
new models by simply providing a profile description. Our multi-label
classifier requires updating the head and fine-tuning again when the model
pool changes
(Section~\ref{sec:discussion}). However, MIRT already trails Switchcraft by
7.59\,pp in answer accuracy on the \emph{known} model pool (2.26\,pp even
when both ignore cost), and its ability projection is fit to only eight
profiles, so its zero-shot performance on unseen models is unlikely to be
competitive without further improvements (e.g., unfreezing the encoder or
tuning its decision rule).

\textbf{Implications for router design.}
This comparison supports our design choice of end-to-end fine-tuning with a
simple multi-label head over a more structured IRT formulation, and it shows
that the decision policy matters as much as the scorer: MIRT's top-score rule
recovers most of its accuracy gap at unmeasured cost, while Switchcraft's
calibrated policy, tuned on validation data under a cost budget, yields
operating points that dominate MIRT's (Table~\ref{tab:budget-sweep}).
End-to-end fine-tuning remains feasible in our setting, where the model pool
is fixed (about 55 minutes per seed for DistilBERT on four GPUs;
Appendix~\ref{app:training-config}).

\subsection{Deviations from the Original IRT-Router}

\begin{table*}[t]
\centering
\caption{Deviations from the original IRT-Router~\cite{irt_router} and
justification. Each change is made to enable a controlled comparison
with our DistilBERT multi-label router on the same data and model pool.}
\label{tab:mirt-deviations}
\small
\setlength{\tabcolsep}{4pt}
\begin{tabular}{p{2.1cm}p{3.1cm}p{3.1cm}p{4.4cm}}
\toprule
\textbf{Aspect} & \textbf{Original IRT-Router} & \textbf{Our MIRT impl.}
  & \textbf{Justification} \\
\midrule
Embedding model
  & BERT-base-uncased (110M)
  & DistilBERT-base-uncased (66M), frozen
  & Same encoder as our multi-label router and kNN baseline. \\
\addlinespace
Task domain
  & QA, reasoning, math, and code (12 datasets; accuracy, EM, F1, or
    pass@1)
  & Agentic function calling (14 in-distribution splits plus held-out
    Glaive; AST scoring)
  & Our evaluation setting for direct comparison. \\
\addlinespace
Candidate LLMs
  & 20 LLMs (GPT-4o, Llama-3.1, etc.)
  & 8 LLMs (GPT-5.x family + Qwen + Kimi)
  & Same model pool as our main experiments for direct comparison. \\
\addlinespace
Routing decision
  & $\arg\max_j [\alpha \cdot \hat{P}(q,M_j) - \beta \cdot C(M_j)]$ with
    fixed per-model cost $C \in [0,1]$
  & Cheapest LLM with $\hat{P} \geq 0.5$; fallback to argmax
  & Cost-aware rule of our earlier protocol; uses profiled per-query cost
    (Section~\ref{sec:chattiness}) rather than fixed per-model pricing. \\
\addlinespace
Cost model
  & Fixed output-token price normalized to $[0,1]$
  & Profiled realized cost per query (\$)
  & Accounts for model chattiness; same cost model as Switchcraft. \\
\addlinespace
Warm-up mechanism
  & kNN-based embedding interpolation ($k{=}5$) for cold-start queries
  & Not used
  & In-distribution test queries share the training distribution
    (stratified split); omitting warm-up isolates the IRT head's
    contribution. Held-out Glaive is evaluated without it, as for every
    router. \\
\addlinespace
Optimization
  & Adam, learning rate 0.002, batch 512
  & Adam, learning rate 0.001, batch 16, 10 epochs, linear warm-up and decay
  & Batch size and schedule of our DistilBERT router (per GPU); best epoch
    restored on validation. \\
\addlinespace
Latent dims ($K$)
  & 25
  & 25
  & Same as original. \\
\addlinespace
NIRT variant
  & Also evaluated (uses predefined ability categories + neural network)
  & Not implemented
  & MIRT and NIRT perform similarly in the original paper; MIRT is
    simpler and sufficient to test the IRT hypothesis. \\
\bottomrule
\end{tabular}
\end{table*}

Table~\ref{tab:mirt-deviations} lists the differences between our
MIRT-DistilBERT implementation and the original IRT-Router~\cite{irt_router},
along with the rationale for each deviation. Most deviations align MIRT with
our experimental setup: it shares Switchcraft's base encoder, data,
partitions, seeds, candidate models, cost model, and test sets.

In summary, our implementation reproduces the core of IRT-Router's MIRT
variant (embedding $\to$ linear projections $\to$ 2PL response function
$\to$ BCE loss) while substituting the encoder, data domain, model pool, and
routing rule to match our experimental setup. The comparison therefore holds
data, partitions, seeds, candidate models, cost model, and test sets
constant, but not the decision policy or the input representation. Our
token-packing ablation suggests the input matters little, so MIRT's
7.59\,pp answer-accuracy gap reflects mainly the frozen-encoder IRT
formulation and its fixed, non-abstaining rule, and the top-score analysis
above shows that a different rule recovers most of it at unmeasured cost.

\section{Additional Limitations and Design Considerations}
\label{app:additional-limitations}

\textbf{Cost-model assumptions.}
We use list-price API rates for all models. Self-hosted deployments would
substitute amortized GPU-hour costs for open-weight models, which can shift
absolute numbers; the relative cost ordering is robust to such substitutions.
The policy's cost budget is expressed in per-model profiled averages, so
realized cost can exceed it when expensive models serve long multi-turn
queries (Appendix~\ref{app:cost-model}).

\textbf{Abstention in deployment.}
Like the oracle, we charge abstentions zero cost, which models a deployment
that returns the query to the agent to ask for clarification, escalate to a
human, or fail fast. A deployment that must always answer can instead fall
back to a default model on abstention.

\textbf{Reasoning effort and cascading.}
Many recent LLMs expose a configurable reasoning effort (e.g.,
low/medium/high). A natural extension is for the router to select not only
the model but also the reasoning level, and its associated cost. Similarly,
multi-model cascading (trying a cheap model first and falling back to an
expensive one on failure) could further reduce average cost.

\end{document}